\documentclass[11pt]{article}

\usepackage{fullpage}
\usepackage{amsmath}
\usepackage{amsfonts} 
\usepackage{amssymb}
\usepackage{amsthm}
\usepackage{float}
\usepackage{url}
\usepackage{graphicx}
\usepackage{color}
\usepackage{bbm}
\usepackage{thmtools}
\usepackage{tikz}
\usetikzlibrary{bayesnet}

\DeclareMathOperator*{\uniformRect}{UniformRect}
\DeclareMathOperator*{\discrete}{Categorical}

\DeclareMathOperator*{\parents}{parents}

\DeclareMathOperator*{\round}{round}
\DeclareMathOperator*{\face}{\mathsf{FACE}}
\DeclareMathOperator*{\eye}{\mathsf{EYE}}
\DeclareMathOperator*{\lefteye}{\mathsf{LEYE}}
\DeclareMathOperator*{\righteye}{\mathsf{REYE}}
\DeclareMathOperator*{\mouth}{\mathsf{MOUTH}}
\DeclareMathOperator*{\nose}{\mathsf{NOSE}}
\DeclareMathOperator*{\like}{\mathsf{T-}}
\DeclareMathOperator*{\facelike}{\mathsf{T-FACE}}
\DeclareMathOperator*{\lefteyelike}{\mathsf{T-LEYE}}
\DeclareMathOperator*{\righteyelike}{\mathsf{T-REYE}}
\DeclareMathOperator*{\mouthlike}{\mathsf{T-MOUTH}}
\DeclareMathOperator*{\noselike}{\mathsf{T-NOSE}}

\DeclareMathOperator*{\curve}{\mathsf{C}}
\DeclareMathOperator*{\ink}{\mathsf{P}}

\DeclareMathOperator*{\argmax}{arg\,max}

\newcommand{\FACE}{\mathrm{FACE}}
\newcommand{\EYE}{\mathrm{EYE}}

\newcommand{\LOWER}{\mathrm{LOWER}}
\newcommand{\EYES}{\mathrm{EYES}}
\newcommand{\SUNGLASSES}{\mathrm{SUNGLASSES}}
\newcommand{\NOSE}{\mathrm{NOSE}}
\newcommand{\MOUTH}{\mathrm{MOUTH}}
\newcommand{\PERSON}{\mathrm{PERSON}}

\newcommand{\LEGS}{\mathrm{LEGS}}
\newcommand{\ARMS}{\mathrm{ARMS}}
\newcommand{\SKIRT}{\mathrm{SKIRT}}
\newcommand{\PANTS}{\mathrm{PANTS}}
\newcommand{\SHOE}{\mathrm{SHOE}}
\newcommand{\HAT}{\mathrm{HAT}}
\newcommand{\BASEBALL}{\mathrm{BASEBALL}}
\newcommand{\SOMBRERO}{\mathrm{SOMBRERO}}

\newcommand{\R}{\mathcal{R}}
\newcommand{\N}{\mathcal{N}}

\newfloat{grammar}{tbp}{grm}
\floatname{grammar}{Grammar}
\newtheorem{theorem}{Theorem}[section]

\newtheorem{proposition}[theorem]{Proposition}
\newtheorem{remark}[theorem]{Remark}
\newtheorem{definition}[theorem]{Definition}

\begin{document}
\title{Scene Grammars, Factor Graphs, and Belief Propagation}

\author{Jeroen Chua \\
  Brown University \\
  Providence, RI, USA \\
  {\tt jeroen\_chua@alumni.brown.edu}
  \and Pedro F. Felzenszwalb \\
  Brown University \\
  Providence, RI, USA  \\
  {\tt pff@brown.edu}}
\maketitle

\begin{abstract}
We describe a general framework for probabilistic modeling of complex
scenes and inference from ambiguous observations.  The approach is
motivated by applications in image analysis and is based on the use of
priors defined by stochastic grammars.  We define a class of grammars
that capture relationships between the objects in a scene and provide
important contextual cues for statistical inference.  The distribution
over scenes defined by a probabilistic scene grammar can be
represented by a graphical model and this construction can be used for
efficient inference with loopy belief propagation.

We show experimental results with two different applications.  One
application involves the reconstruction of binary contour maps.
Another application involves detecting and localizing faces in
images. In both applications the same framework leads to robust
inference algorithms that can effectively combine local information to
reason about a scene.
\end{abstract}

\section{Introduction}
\label{sec:intro}

We consider the problem of knowledge representation for scene
understanding.  The problem has a significant history in pattern
recognition, artificial intelligence, and statistics
(e.g., \cite{GPT,Fu74,Pearl88,B86,WJ08}).  Our primary goal is to
develop a general purpose framework for modeling complex scenes and
for efficient inference and learning with such models.

The philosophy behind our approach follows Grenander's \emph{Pattern
  Theory} \cite{GPT} program.  A key idea in Pattern Theory is to
model patterns in a variety of different settings using algebraic and
probabilistic systems that define regular structures and probability
distributions over these structures.  The approach emphasizes the
relationship between pattern analysis and pattern synthesis, and is
based on the use of Bayesian statistics for inference of hidden
structures.

In the Bayesian formulation of scene understanding a prior model
captures the statistical regularities of scenes in the world.  The
prior model makes it possible to reason about typical scenes and to
estimate a scene from partial or ambiguous observations.  A
significant challenge within this framework is to define a class of
models that is both expressive \emph{and} computationally tractable.
That is, one would like to define models that can capture regular
structures in a variety of applications and design efficient
algorithms for inference with such models.

The general framework we describe in this paper is based on three key
components: (1) A class of stochastic grammars for modeling complex
scenes; (2) The construction of graphical models that represent scene
distributions; and (3) An efficient method for approximate inference
with loopy belief propagation.

The types of scenes we consider often lead to complex high-dimensional
structures.  For example, in Section~\ref{sec:exfaces} we consider
scenes with faces and parts of faces.  Even in this simple setting a
scene can contain a variable number of objects and each object can be
in many different locations.  The number of possible scenes is very
large (or infinite).  Nonetheless scenes have regularities
(e.g.\ faces have eyes).  This is similar to the situation in formal
language models where languages defined by grammars and automata often
include an infinite set of well-formed sentences.

A \emph{probabilistic scene grammar} defines a set of (regular) scenes
and a probability distribution over scenes.  Scenes are defined using
a set of building blocks and a set of production rules.  The
productions define typical co-occurances of small groups of objects.
Productions are chained together to form larger compositions, and this
process leads to a large set of possible structures.  The scenes
generated by a scene grammar capture both the presence of different
objects and relationships among them (such as part-of, or
aligned-with).  The set of scenes generated by a scene grammar defines
a language of regular scenes.  The set of scenes can also be seen as a
state space and is similar to the configuration space in Pattern
Theory.

A scene generated by a probabilistic scene grammar can be
represented using a finite set of random variables.  Importantly, we
derive a product formula for the probability of a scene using this
representation.  This leads to a graphical model that can be used for
inference via local message passing methods such as belief propagation
(\cite{Pearl88}).

Inference is a challenging computational problem.  We describe an
efficient method for inference using \emph{loopy} belief propagation.
The approach involves efficient computation of messages in the
sum-product algorithm.  This is possible due to the special structure
of the graphical models that arise within our framework.  Inference
with loopy belief propagation aggregates information using the
production rules in a grammar.  For example, if a production rule
generates the eyes, nose and mouth of a face, any evidence for the
face or one of its parts provides contextual evidence for the whole
composition.



Throughout the paper we describe examples of scene grammars that
generate different kinds of scenes motivated by problems in computer
vision and image analysis.  We illustrate the results of numerical
experiments with these models, both for reasoning about abstract
scenes and in practical applications.  All of the numerical
experiments were performed using a common implementation of a
general computational engine.  

We include experimental results with two applications.  One
application involves detecting curves in noisy images and
reconstructing binary contour maps.  We address this problem using a
grammar that generates scenes with a collection of regular curves.
Another application involves detecting faces and parts of faces in
images.  In this case we use a grammar that captures spatial
relationships between the parts that make up a face.  In both
applications the framework and computational methods described
here lead to good experimental results.

\subsection{Motivation and Related Work}

Formal grammars, and in particular context-free grammars, have been
widely used for modeling the structure of sentences in natural
language (\cite{C56,Manning99}).  Similar models have been used to
represent patterns in biological structures such as DNA and RNA
(\cite{D98}).  Grammars are also commonly used to define the syntax
of programming languages (\cite{ASU86}).

In computer graphics the recursive description of objects using
grammars and rewriting systems has been used to generate complex geometric
objects, biological structures, and landscapes (\cite{PL91,D06}).  These
methods illustrate the potential that (stochastic) grammars have for
modeling natural and man-made structures.

Since the early days of image analysis attempts have been made to
develop formal language models for two-dimensional pictures (see,
e.g., \cite{R79}).  There are significant challenges in designing
effective models and algorithms for parsing images using 
grammars, including the fact that the pixels in an image are not
linearly ordered.  One important difference between parsing images and
parsing sentences is the number of possible constituents.  The number
of continuous subsentences of a sentence of length $n$ is quadratic in
$n$.  On the other hand, the number of connected regions in an image
with $n$ pixels is exponential in $n$.  This means that classical
parsing algorithms based on dynamic programming over constituents
cannot be easily applied to images.

The interpretation of scenes via repeated application of grouping
rules is one of the key ideas behind the Gestalt theory of visual
perception (\cite{Palmer99}).  The structures generated by
a scene grammar are similar to the hierarchical description of a scene
using grouping rules.  The approach is also related to the description
of scenes using compositional models and the MDL principle
\cite{BGP97}.

Statistical models and Bayesian inference methods have been previously
used to solve a variety of problems in computer vision and image
analysis (e.g., \cite{B86, GG84, CJ93, M94b, GJ96, A02, FH05}).
Probabilistic scene grammars provide a unified framework to address
many of these problems.

Probabilistic scene grammars generalize part-based models for object
detection and recognition such as pictorial structures
\cite{FE73,FH05} and constellations of features \cite{BWP98}.  In a
grammar model objects can be defined using a hierarchy of reusable parts.
Probabilistic scene grammars can also represent objects with variable
structure (when the parts that make up an object can change) and
scenes with multiple objects.  Furthermore, the use of recursion 
allows for modeling complex objects, such as curves of varying lengths and
shape, using a small finite set of production rules.

Besides allowing for the definition of very general models, grammars
also provide a useful conceptual abstraction to reason about models and
shared parameters.  

\begin{figure}
  \centering
  \includegraphics[width=3.5in]{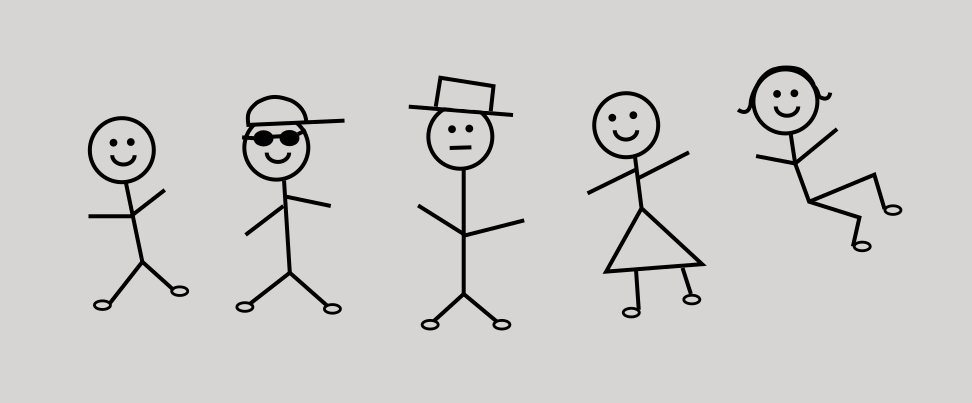}
  \vspace{.2cm}
  \begin{align*}
    \PERSON & \rightarrow \FACE, \ARMS, \LOWER \\
    \FACE & \rightarrow \EYES, \NOSE, \MOUTH \\
    \FACE & \rightarrow \HAT, \EYES, \NOSE, \MOUTH \\
    \EYES & \rightarrow \EYE, \EYE \\
    \EYES & \rightarrow \SUNGLASSES \\
    \HAT & \rightarrow \BASEBALL \\
    \HAT & \rightarrow \SOMBRERO \\
    \LOWER & \rightarrow \SHOE, \SHOE, \LEGS \\
    \LEGS & \rightarrow \PANTS \\
    \LEGS & \rightarrow \SKIRT
  \end{align*}
  \vspace{-.5cm}
  \caption{Production rules in a grammar that generates scenes with
    people.  The parts that make up a person can vary and a small set
    of rules can generate a large number of structures.}
  \label{fig:people}
\end{figure}

Figure~\ref{fig:people} shows an example of a set of production rules
in a grammar that can generate scenes with multiple people that are
composed of different parts.  In this case a person always has parts
corresponding to the face, arms, and lower-body.  Some faces have hats
while other faces do not, and hats can be of different types.
Similarly some people wear skirts while other people wear pants.
Since we have multiple independent choices for the rules used to
expand different symbols we see a combinatorial explosion in the
number of possible structures that can be generated with a small
number of rules.

Probabilistic scene grammars are closely related to the Markov
backbone in \cite{JG06}.  Other related work includes
\cite{GPZ02,TCYZ05,ZM07,ZCY09,ZZ11,FM10}.
These methods have generally relied on MCMC or heuristic
methods for inference, or dynamic programming for restricted models.
The approach we develop here using loopy belief propagation
is related to the dynamic programming approach but can be
applied in a more general setting.

Several probabilistic programming languages have been designed with
the goal of providing a general framework for probabilistic modeling
and inference.  For example, the Picture and Edward frameworks
(\cite{Picture,Edward}) share the high-level goal of having a
general-purpose modeling system for scene understanding.  These
methods seek high-probability scenes encoded as probabilistic program
traces via MCMC and variational inference.  The scene grammars we
define are closely related to a probabilistic program where all
functions are ``memoized'' (in the sense of dynamic programming).
This allows for the representation of all possible program traces with
a finite graphical model and for the implicit reasoning about program
execution via loopy belief propagation.

\subsection{Overview}

In Section~\ref{sec:model} we define a class of probabilistic scene
grammars and the generative process used to generate scenes.
In Section~\ref{sec:examples} we give examples of scene grammars that
generate different types of scenes.  In Section~\ref{sec:factorgraph}
we derive a product formula for the probability of a scene and define
a general method for constructing graphical models that represent
scene distributions.  In Section~\ref{sec:infbp} we describe an
efficient inference method based on loopy belief propagation and in
Section~\ref{sec:infexamples} we illustrate the result of inference
using the example grammars from Section~\ref{sec:examples}.
Section~\ref{sec:learning} considers the problem of learning model
parameter using maximum-likelihood estimation.  In
Section~\ref{sec:exp} we illustrate the results of our methods in
two different applications.  

\section{Probabilistic Scene Grammars}
\label{sec:model}

Our goal is to reason about scenes and to estimate a description of a
scene from a set of observations.  A key component of the approach is
a prior model over scenes, $p(S)$, that captures the statistical
regularities of scenes in the world.

A \emph{probabilistic scene grammar} defines a set of possible scenes
and a probability distribution over scenes.  Scenes are defined using
a set of building blocks, or \emph{bricks}, that are
connected together using a set of basic compositions, or
\emph{production rules}.

To generate scenes we define a random process that grows a scene
starting from an initial set of bricks.  Initially every brick is
considered to be included in the scene independently.  We then
repeatedly use production rules to expand (or grow) bricks that
are in the scene but have not been expanded before.  The result of
this process defines a distribution, $p(S)$, that can capture
regularities in natural scenes.  For example, the process can capture
which objects tend to co-occur in a scene and the typical relative
positions between such objects.

Each brick is defined by a pair consisting of a
\emph{symbol} and a \emph{pose}.  For example, one brick 
might be the pair $(\face, (30,40))$ representing a face at location
$(30,40)$ in an image.  

In a probabilistic scene grammar the initial generation of bricks is
governed by self-rooting probabilities.  The possible expansions of a
brick are determined by a set of production rules, rule selection
probabilities and conditional pose distributions.


Let $\Sigma$ be a finite set.  We use $\Sigma^*$ to denote the set of
finite strings of elements from $\Sigma$, including the empty string.

\begin{definition}
  \label{def:psg}
  A probabilistic scene grammar (PSG) is defined by a 6-tuple
  $\mathcal{G}= (\Sigma, \Omega, \R, q, \rho, \epsilon)$, where
  \begin{enumerate}
  \item $\Sigma$ is a finite set (the symbols).
    
  \item $\Omega = \{\,\Omega_A \mid A \in \Sigma\,\}$ where $\Omega_A$
    is a finite set (the pose spaces).

  \item $\R$ is a finite subset of $\Sigma \times \Sigma^*$ (the production rules).
    
    A production rule is specified in the form $A_0 \rightarrow
    A_1,\ldots,A_n$ where $n \ge 0$ and $A_i \in \Sigma$.
    
    For $r \in \R$ we use $n_r$ to denote the number of symbols in the
    right-hand-side of $r$ and $A_{(r,i)}$ to denote the $i$-th symbol
    in $r$.  

    Let $\R_A$ be the set of rules with $A \in \Sigma$ in the
    left-hand-side.  We assume $R_A \neq \emptyset$.

  \item $q = \{\,q_A \mid A \in \Sigma\,\}$ where $q_A$ is a
    distribution over $\R_A$ (the rule selection probabilities).
    
  \item $\rho = \{\,\rho_{(r,i)} \mid r \in \R,\, 1 \le i \le n_r\,\}$
    is a set of conditional pose distributions.
    
    For a rule $r = A_0 \rightarrow A_1,\ldots,A_n$ we have
    $\rho_{(r,i)} : \Omega_{A_i} \times \Omega_{A_0} \rightarrow \mathbb{R}_{\ge 0}$ with
    $$\sum_{\omega_i \in \Omega_{A_i}} \rho_{(r,i)}(\omega_i,\omega_0) = 1,\;\;\ \forall \omega_0 \in \Omega_{A_0}.$$
    We use $\rho_{(r,i)}(\omega_i | \omega_0)$ to denote $\rho_{(r,i)}(\omega_i,\omega_0)$.
    
  \item $\epsilon = \{\,\epsilon_A \mid A \in \Sigma\,\}$ where $0 \le \epsilon_A \le 1$ (the self-rooting probabilities).
  \end{enumerate}
\end{definition}

The definition of a probabilistic scene grammar is closely related to
the definition of a probabilistic context-free grammar (PCFG) used in
formal language models and in natural language processing (see, e.g.,
\cite{Manning99}).  However, the structures that are generated by a
scene grammar are different from the structures generated by a PCFG
(see Remark~\ref{rem:PCFG}).


Let $\mathcal{G} = (\Sigma, \Omega, \R, q, \rho, \epsilon)$ be a scene
grammar.  The structure of a scene is defined by $(\Sigma, \Omega, \R)$
 and the parameters $(q,\rho,\epsilon)$ define a
(non-uniform) probability distribution over ``well-formed'' scenes.
In practice we assume $\rho_{(r,i)}(\omega_i | \omega)$ is
invariant to a set of transformations (such as translations) of the
pose spaces and this significantly reduces the number of parameters in
the model.

\begin{definition}
The bricks defined by a scene grammar $\mathcal{G}$ are pairs of symbols and poses,
$$\mathcal{B} = \{\,(A,\omega) \mid A \in \Sigma, \omega \in \Omega_A\,\}.$$
\end{definition}

\begin{definition}
A scene $S$ generated by $\mathcal{G}$ is defined by:
\begin{enumerate}
\item A subset $O \subseteq \mathcal{B}$ of bricks that are present in $S$.
\item For each $(A,\omega) \in O$ a selection of rule $A \rightarrow
  A_1,\ldots,A_n \in \R$ and poses $\omega_i$ such
  that $(A_i,\omega_i) \in O$ for $1 \le i \le n$.
  We say $(A,\omega)$ expands to, or is a parent of,
  $(A_i,\omega_i)$ in $S$.
\end{enumerate}
\end{definition}

Let $\mathcal{S}$ be the set of scenes generated by a grammar
$\mathcal{G}$.  The set $\mathcal{S}$ is the ``Language'' generated by
$\mathcal{G}$.  Note that each $S \in \mathcal{S}$ specifies both a
set of objects that are present in the scene (the bricks) and
relationships between these objects.  

To define the scene generation process we use a set $O$ to keep track
of bricks in the scene and a set $Q$ to keep track of a queue of
bricks that are in the scene but have not been expanded yet.
Initially bricks are included in the scene independently, according to
self-rooting probabilities.  All of these bricks are queued for
expansion.  The process terminates when all bricks in the scene
have been expanded ($Q$ is empty).  If an expansion generates a brick
that is not already in the scene we add the brick to the scene and
queue it for expansion.

\begin{definition}
  Random scene generation with a grammar $\mathcal{G} = (\Sigma, \Omega, \R, q, \rho, \epsilon)$.
  \begin{enumerate}
  \item Initially $O = \emptyset$ and $Q = \emptyset$.
  \item For each brick $(A,\omega) \in \mathcal{B}$ add $(A,\omega)$ to both $O$ and $Q$ with probability $\epsilon_A$.
  \item While $Q \neq \emptyset$
    remove a brick $(A,\omega)$ from $Q$ and expand it.
  \item Expanding $(A,\omega)$ involves
  \begin{enumerate}
  	\item Sampling a rule $r = A \rightarrow A_1,\ldots,A_n \in \R$ according to $q_A$.
  	\item Sampling poses $\omega_i$ according to $\rho_{(r,i)}(\omega_i | \omega)$.
  	\item If $(A_i,\omega_i) \not\in O$ add it to both $O$ and $Q$.
  \end{enumerate}
  \end{enumerate}

  The output of the process is a scene $S$ defined by $O$ and
  the rules and poses selected when expanding each brick in $O$.
\end{definition}

The scene generation process always terminates after at most
$|\mathcal{B}|$ expansions.  When the process terminates every brick
in $O$ will have been expanded exactly once, and therefore the output
is a valid scene $S \in \mathcal{S}$.  We note that the order in which
the bricks in $Q$ are selected for expansion does not influence the
probability, $p(S)$, of generating a scene.

The self-rooting probabilities, $\epsilon_A$, control the number of
bricks that are included in the scene initially, before any
bricks are expanded.  For $A \in \Sigma$ the expected number of bricks
of type $A$ that self-root is $\epsilon_A |\Omega_A|$.  Since the pose
spaces are often very large, the self-rooting probabilities are
usually very small.  Note however that even if a brick $(A,\omega)$
self-roots, another brick $(B,z)$ may expand and become a parent of
$(A,\omega)$.  The number of self-rooted bricks is an upper bound on
the number of bricks with no parents.

\begin{remark}
  \label{rem:PCFG}
  Scene grammars are related to PCFGs used in language modeling but they
  generate different types of structures.
  
  Recall that a PCFG generates rooted derivation trees, where the
  vertices are labeled with symbols from a finite alphabet.  In a
  derivation tree there is a single vertex (the root) with no parents
  and every other vertex has a unique parent.
  
  A scene generated by a PSG defines a scene graph $G$ over the bricks
  that are present in the scene.  The edges of $G$ capture which bricks
  expand to each other.  The scene graph $G$ resembles a derivation tree
  generated by a PCFG, but it has more general connectivity structure.
  In particular there can be multiple vertices with no parents (roots)
  in $G$, and the graph can have multiple disjoint components.  There
  can also be vertices with multiple parents in $G$.  The scene graph
  $G$ can also have directed cycles, when a brick leads to a sequence of
  expansions that eventually generates the same brick again.
  
  Finally we note that every scene graph is a subgraph of the complete
  directed graph over $\mathcal{B}$, and the number of possible scene
  graphs is finite (although it can be very large).  This is in contrast
  to the fact that a context-free grammar can generate trees of
  unbounded size.
\end{remark}

\section{Example Grammars}
\label{sec:examples}

In this section we give examples of PSGs and illustrate the
random scenes they generate.

As described in Section~\ref{sec:model} a scene grammar $\mathcal{G}$
is defined by a 6-tuple $(\Sigma, \Omega, \R, q, \rho, \epsilon)$.  In
the examples below we combine the description of $\R$, $q$ and $\rho$
to simplify the notation.  Let $r = A_0 \rightarrow A_1,\ldots,A_n$ be
a rule in $\R$.  To specify the rule, $r$, the rule selection
probability, $q_r$, and the conditional pose distributions,
$\rho_{(r,i)}$, we write,
\begin{equation*}
\begin{array}{llll}
  q_r, & (A_0,\omega_0) & \rightarrow & (A_1,\rho_{(r,i)}(\cdot|\omega_0)),\ldots,(A_n,\rho_{(r,n)}(\cdot|\omega_0)).
\end{array}
\label{eqn:rnotation}
\end{equation*}

In the examples in this section the pose spaces are grids of integer
points $[N_1] \times \cdots \times [N_D]$, where $[N] =
\{0,\ldots,N-1\}$.  Let $a$ and $b$ be two points in a grid.  We use
$\uniformRect(a,b)$ to denote a uniform distribution over grid
points in the rectangular region with corners $a$ and $b$.  We use
$\delta(a)$ to denote the distribution concentrated at $a$.

\subsection{Scenes with Faces}
\label{sec:exfaces}

\begin{figure}
  \centering
  \setlength\fboxsep{0pt}
  \begin{tabular}{cc}
    \fbox{\includegraphics[width=6.5cm]{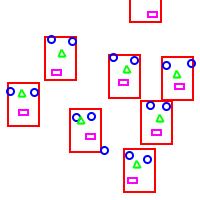}} &
    \fbox{\includegraphics[width=6.5cm]{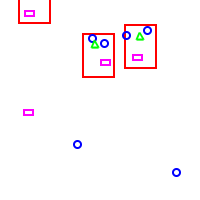}} \\
    \fbox{\includegraphics[width=6.5cm]{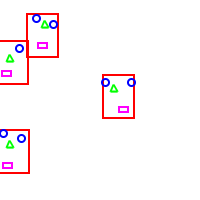}} &
    \fbox{\includegraphics[width=6.5cm]{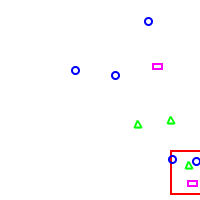}}
  \end{tabular}
  \caption{Random scenes generated using
    Grammar \ref{grammar:face_simple}.  Faces are represented by red
    rectangles, eyes by blue circles, noses by green triangles, and
    mouths by magenta rectangles.  Scenes have multiple objects and
    parts of faces can appear both in the context of a face and on
    their own.  The location of each face part can vary
    within a range of possible locations relative to the face.}
  \label{fig:face_samples}
\end{figure}

Grammar \ref{grammar:face_simple} generates scenes with faces and
parts of faces.  Figure~\ref{fig:face_samples} illustrates four random
scenes generated by this grammar.  The model captures the notion that
faces have certain parts at appropriate locations.
This is similar to other part-based representations
such as pictorial structures (e.g.\ \cite{FE73} and \cite{FH05}) and
deformable part models (\cite{FGMR10}).
However, the grammar generates scenes with multiple
faces.  It also allows for parts of faces to appear on their own,
capturing the notion that a scene is made up of faces and other
components that look like parts of faces.  

\begin{grammar}
  
  \noindent{\parbox{\textwidth}{%

      \hrule
      \vspace{.2cm}

  $\Sigma = \{ \face, \eye, \nose, \mouth \}$.
      
  $\forall A \in \Sigma,\; \Omega_A = [N] \times [M]$.
  
  Rules: 
  
  $\begin{array}{lllll}
    (1) & 1.0, & (\face,(x,y)) & \rightarrow & (\eye,\uniformRect((x,y)+a_1,(x,y)+b_1)), \\
    & & & & (\eye,\uniformRect((x,y)+a_2,(x,y)+b_2)), \\
    & & & & (\nose,\uniformRect((x,y)+a_3,(x,y)+b_3)), \\
    & & & & (\mouth,\uniformRect((x,y)+a_4,(x,y)+b_4)). \\
    (2) & 1.0, & (\eye,(x,y)) & \rightarrow & \emptyset. \\
    (3) & 1.0, & (\nose,(x,y)) & \rightarrow & \emptyset. \\
    (4) & 1.0, & (\mouth,(x,y)) & \rightarrow & \emptyset.
  \end{array}$
  
  $\epsilon_{\face} =  10^{-4}.$
  
  $\epsilon_{\eye} = \epsilon_{\nose} = \epsilon_{\mouth} = 10^{-5}.$

  \vspace{.2cm}
  \hrule

  }}
  
  \caption{A grammar for scenes with faces and parts of faces.}
  \label{grammar:face_simple}     
\end{grammar}

\begin{figure}
  \centering
  \includegraphics[height=7cm]{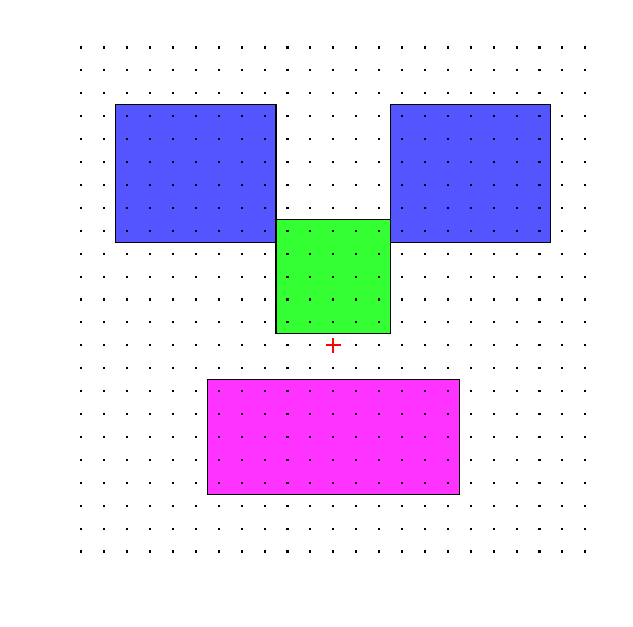}  
  \caption{Possible locations for each face part when
    expanding a face at the red-cross location.  The
    location of each part is selected uniformly at random from
    a rectangular region.  The regions for the eyes, nose, and mouth
    are shown in blue, green and magenta.}
  \label{fig:face_parts}
\end{figure}

In this grammar the pose of a brick specifies a 2D discrete location
for an object (face, eye, nose or mouth) of fixed size and
orientation.  Rule (1) defines a part-based model for faces.  The
location of each face part is selected at random from a rectangular
region defined \emph{relative} to the location of the face.
Figure~\ref{fig:face_parts} illustrates the possible locations for the
parts that make up a face at one location.
The self-rooting probabilities of the part symbols
are smaller than the self-rooting probability of the face symbol, so
that most parts appear in the context of a face.

In Section~\ref{sec:facereason} we show how this model can be used for
reasoning about scenes with faces.  In Section~\ref{sec:face} we show how a
similar model can be used for face detection.

The grammar defined here can be extended to represent objects of
different sizes (and orientations) if we augment the pose spaces with
scale (and orientation) information.  The parts of the face can also
be defined in terms of smaller parts (such as eyebrow and pupil for
an eye).  The grammar can also be extended to define scenes with many
types of objects, and where objects of different types are defined
using a set of reusable parts.  By including additional productions
rules in the grammar we can also define objects with variable
structure.

\subsection{Scenes with Curves}
\label{sec:excurves}

Grammar~\ref{grammar:curve_simple} generates scenes with discrete
curves in a finite two-dimensional grid.  The grammar generates scenes
with a random number of curves and where each curve has a random
length and shape, giving preference to curves with low curvature
everywhere.  As discussed in \cite{SU88} these are the types of curves
that are often perceptually salient to humans.  

\begin{figure}
  \centering
  \setlength\fboxsep{0pt}
  \begin{tabular}{cc}
    \fbox{\includegraphics[width=6.5cm]{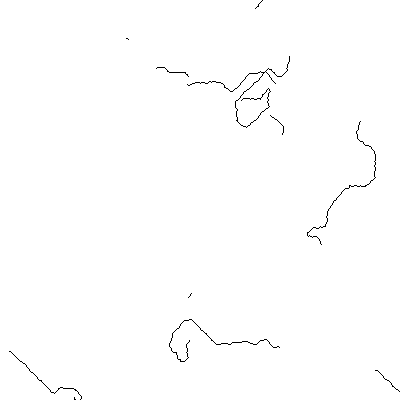}} & 
    \fbox{\includegraphics[width=6.5cm]{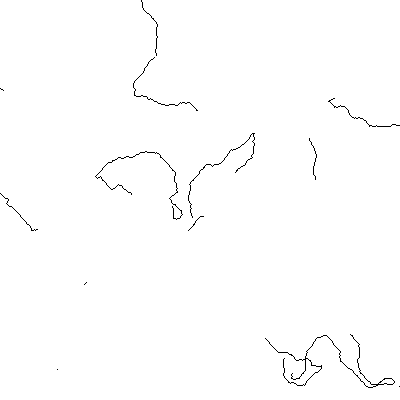}} \\
    \fbox{\includegraphics[width=6.5cm]{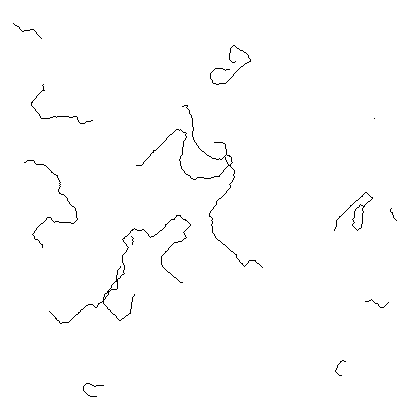}} & 
    \fbox{\includegraphics[width=6.5cm]{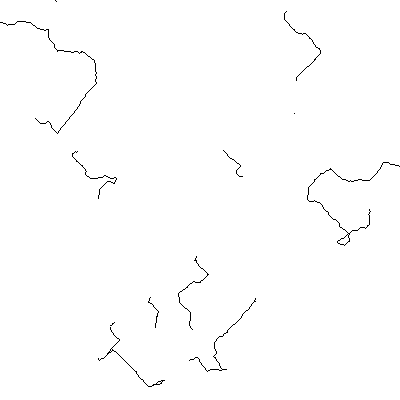}}
  \end{tabular}
  \caption{Random images generated using Grammar
    \ref{grammar:curve_simple}. The grammar generates scenes with
    multiple curves of varying lengths and shapes, giving a preference
    to curves with low curvature.  In each image the black pixels
    represent the $\ink$ bricks that are part of a scene.}
  \label{fig:curve_samples}
\end{figure}

The process that generates an individual curve with
Grammar~\ref{grammar:curve_simple} traces the path of a particle undergoing
stochastic motion.  Similar models were
introduced in \cite{M94} and \cite{WJ97} in the context of contour
completion.  Figure~\ref{fig:curve_samples} shows four random
images generated by the grammar.
In Section~\ref{sec:completion} we
show how this model can be used for contour completion.  In
Section~\ref{sec:curve} we show how the model can be used to detect
curves in noisy images and to reconstruct binary contour maps.

\begin{grammar} 

  \noindent{\parbox{\textwidth}{%

      \hrule
      \vspace{.2cm}

  $\Sigma = \{ \curve, \ink \}$.
  
  $\Omega_{\curve} = [N] \times [M] \times [8]$.
  
  $\Omega_{\ink} = [N] \times [M]$.
      
  Rules:
  
  $\begin{array}{lllll}
    (1) & 0.65, & \hspace{-0.20cm} (\curve,(x,y,\theta)) & \rightarrow  & (\ink, \delta((x,y))),
    (\curve,\delta(((x,y)+\round(T_{\theta}(1,0)),\theta))). \\
    (2) & 0.124, & \hspace{-0.20cm} (\curve,(x,y,\theta))  & \rightarrow & (\ink, \delta((x,y))),
    (\curve,\delta(((x,y)+\round(T_{\theta-1}(1,0)),\theta))). \\
    (3) & 0.124, & \hspace{-0.20cm} (\curve,(x,y,\theta)) & \rightarrow & (\ink, \delta((x,y))),
    (\curve,\delta(((x,y)+\round(T_{\theta+1}(1,0)),\theta))). \\
    (4) & 0.05, & \hspace{-0.20cm} (\curve,(x,y,\theta)) & \rightarrow & (\curve, \delta((x,y,\theta-1))). \\
    (5) & 0.05, & \hspace{-0.20cm} (\curve,(x,y,\theta)) & \rightarrow & (\curve, \delta((x,y,\theta+1))). \\
    (6) & 0.002, & \hspace{-0.20cm} (\curve,(x,y,\theta)) & \rightarrow & (\ink, \delta((x,y))). \\
    (7) & 1.00, & \hspace{-0.20cm} (\ink,(x,y)) & \rightarrow & \emptyset.
  \end{array}$
  
  $\epsilon_{\curve} = \epsilon_{\ink} =  10^{-5}$.

  \vspace{.2cm}
  \hrule
  
  }}

  \vspace{.1cm}

  The function $T_{\theta}$ denotes a rotation in the plane by a
  discrete angle $\theta$ and $\round$ maps a point in the plane to
  the nearest grid point. 

  \caption{A grammar for scenes with discrete curves.}
  \label{grammar:curve_simple}
\end{grammar}

The grammar has two symbols, $\Sigma = \{ \curve, \ink \}$.  The
$\curve$ bricks represent oriented elements (or particle states) that are
generated in a sequence using a recursive rule to form discrete curves.  The
pose of a $\curve$ brick specifies a location in an $[N] \times [M]$
pixel grid and one of 8 possible discrete orientations.  The $\ink$
bricks represent the trace of the curves in the pixel grid.  The
pose of a $\ink$ brick specifies only a grid location and has no
orientation information.  

\begin{figure}
  \centering
  \includegraphics[height=4cm]{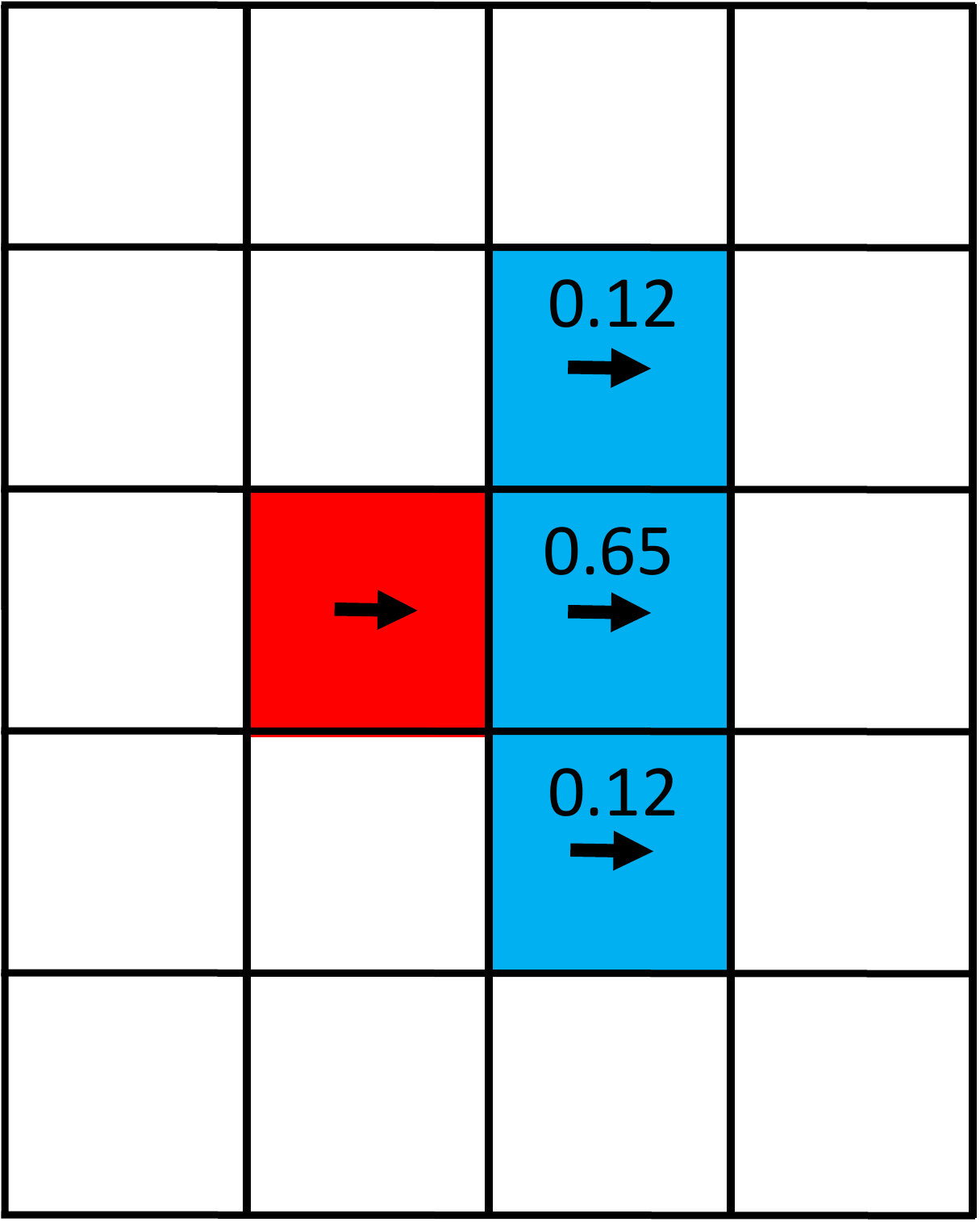}  
  \caption{Possible extensions of a curve by one pixel.  A horizontal
    $\curve$ brick (in red) expands to one of three possible
    horizontal $\curve$ bricks (in blue) with certain probabilities.
    The remaining probability mass is reserved for the choice to end
    the curve or change its orientation.}
  \label{fig:curve_to}
\end{figure}

Rules (1)-(3) capture the possible extensions of a curve along a
direction close to the current orientation.  Figure~\ref{fig:curve_to}
illustrates the possible extensions defined by rules (1)-(3) for a
horizontal brick.  When we extend a curve at pixel $(x,y)$ with
orientation $\theta$, we move to one of 3 neighbors of $(x,y)$ that
are approximately in the direction $\theta$.  The curves generated are
therefore contiguous.  As we generate a curve we also generate
$\ink$ bricks tracing the path of the curve.

Rules (4)-(5) capture changes in orientation and rule (6) captures the
termination of a curve.  The probability of changing orientation is
small, therefore curves tend to take multiple steps with a single
discrete orientation before turning.  This leads to curves with ``low
curvature'' almost everywhere.  The probability of selecting rule (6)
controls the expected length of a curve.

\section{Graphical Model}
\label{sec:factorgraph}

A PSG defines a probability distribution over a set
of scenes.  In this section we consider a representation of scenes
using a finite set of random variables, and derive a closed form
expression for the probability of a scene using a product of local
functions.  This leads to an undirected graphical model that can be used for
inference (Section~\ref{sec:infbp}) and for learning model parameters
(Section~\ref{sec:learning}).

We start by defining a representation of scenes using binary random variables.

Let $\rho = \{\, \rho_{(r,i)} \,\}$ be the conditional pose
distributions associated with a scene grammar $\mathcal{G}$.  Let
$\Gamma_{(r,i,\omega)}$ denote the support of
$\rho_{(r,i)}(\cdot\,|\,\omega)$,
$$\Gamma_{(r,i,\omega)} = \{\, z \in \Omega_{A_{(r,i)}} \mid \rho_{(r,i)}(z\,|\,\omega) > 0 \,\}.$$

\begin{definition}
  Let $\mathcal{G}$ be a scene grammar and $S$ be a random scene generated by
  the grammar.  Define the following random variables associated with
  each brick $(A,\omega) \in \mathcal{B}$,
  \begin{eqnarray*}
    & & X(A,\omega)  \in \{0,1\}, \\
    & & R(A,\omega) = \{\, R(A,\omega,r) \in \{0,1\} \mid r \in \R_A \,\}, \\
    & & C(A,\omega) = \{\, C(A,\omega,r,i,z) \in \{0,1\} \mid r \in \R_A, 1 \le i \le
  n_r, z \in \Gamma_{(r,i,\omega)} \,\},
  \end{eqnarray*}
  where

  $X(A,\omega) = 1$ iff $(A,\omega)$ is in $S$,

  $R(A,\omega,r) = 1$ iff rule $r$ is used to expand $(A,\omega)$ in $S$,
  
  $C(A,\omega,r,i,z) = 1$ iff $(A,\omega)$ is expanded with rule $r$ in $S$,
  and $z$ is the pose selected for $A_{(r,i)}$.
\label{def:rv}
\end{definition}

\begin{remark}
  A scene $S$ is uniquely defined by the value of the random variables $\{ X, R, C \}$.
\end{remark}
  
\subsection{Product Formula}

We now consider a class of \emph{acyclic} PSGs and derive
an expression for the probability of a scene, $p(S)$, using a product
of local functions.  This expression is analogous to the expression of
the joint distribution in a Bayesian network.  For the case of
\emph{cyclic} grammars the product expression can be used as a
tractable approximation to $p(S)$.

Let $\mathcal{G}$ be a PSG with a set of bricks $\mathcal{B}$.  We say
$\mathcal{G}$ is \emph{acyclic} if there is no sequence of expansions
that generates a brick starting from itself.  Equivalently, let $H$ be
the directed graph over $\mathcal{B}$, with an edge from $(A,\omega)$
to $(B,z)$ if $(A,\omega)$ can generate $(B,z)$ in one expansion.  The
grammar $\mathcal{G}$ is acyclic if $H$ is an acyclic directed graph.
For example, the grammar for scenes with faces in
Section~\ref{sec:exfaces} is acyclic.  On other other hand, the
grammar for scenes with curves in Section~\ref{sec:excurves} is
cyclic.

A \emph{topological ordering} of $\mathcal{B}$ is a linear ordering of
$\mathcal{B}$ such that $(A,\omega)$ appears before $(B,z)$ whenever
$(A,\omega)$ can generate $(B,z)$ after one or more expansions.
When $\mathcal{G}$ is acyclic there is always a topological
ordering of $\mathcal{B}$ and such an ordering can be computed by
topologically sorting the vertices of $H$ (see, e.g., \cite{CLRS}).

There are two types of potential functions in the factorization of
$p(X,R,C)$.

\begin{definition}
  \label{def:leaky-or}
  A leaky-OR potential $\Psi^L_\epsilon(y,z)$ is a function of a
  $d$-dimensional binary vector $y$ (the input) and a binary value $z$
  (the output).  It encodes the conditional probability of each output
  value in a probabilistic OR gate.  For a binary vector $y$ let
  $c(y)$ be the number of ones in $y$.  If $c(y) > 0$ we have $z=1$
  with probability $1$.  If $c(y) = 0$ we have $z=1$ with probability
  $\epsilon$.
$$\Psi^L_\epsilon(y,z) =
\begin{cases}
  1 & c(y) > 0,\, z = 1, \\
  0 & c(y) > 0,\, z = 0, \\
  \epsilon & c(y) = 0,\, z = 1, \\
  1-\epsilon & c(y) = 0,\, z = 0.
\end{cases}$$
\end{definition}

\begin{definition}
  \label{def:selection}
  A selection potential $\Psi^S_\theta(y,z)$ is a function of a binary
  value $y$ (the input) and a $d$-dimensional binary vector $z$ (the
  output).  It encodes the conditional probability of each output
  vector in a switch with $d$ binary outputs.  If $y=0$ no output
  is selected with probability $1$.  If $y=1$ a single output is
  selected according to probabilities defined by $\theta \in
  \mathbb{R}^d$.
$$\Psi^S_\theta(y,z) =
\begin{cases}
  1 & y = 0,\, c(z) = 0, \\
  0 & y = 0,\, c(z) > 0, \\
  0 & y = 1,\, c(z) \neq 1, \\
  \theta_i & y = 1,\, c(z) = 1,\, z_i = 1.
\end{cases}$$
\end{definition}

To formulate the expression for $p(X,R,C)$ we also
need to define the following collections of random variables,
\begin{eqnarray*}  
  C(A,\omega,r,i) & = & \{\, C(A,\omega,r,i,z) \mid z \in \Gamma_{(r,i,\omega)} \,\}, \\
  \parents(X(A,\omega)) & = & \{\, C(B,z,r,i,\omega) \mid (B,z) \in \mathcal{B}, r \in \R_B,
1 \le i \le n_r, A_{(r,i)} = A, \omega \in \Gamma_{(r,i,z)}\,\}.    
\end{eqnarray*}
Let $(A,\omega)$ be a brick.  The set $C(A,\omega,r,i)$ includes all
the random variables that specify the $i$-th child of $(A,\omega)$ if
rule $r$ is used to expand $(A,\omega)$.  The set
$\parents(X(A,\omega))$ includes all the random variables that specify
the parents of $(A,\omega)$.

\begin{theorem}
  \label{thm:factorization}
  The distribution $p(X,R,C)$ defined by an acyclic
  grammar $\mathcal{G}$ can be expressed as,
  \begin{multline}
  \label{eqn:factored}
  p(X,R,C) = \\
  \prod_{(A,\omega) \in \mathcal{B}} \left( p(X(A,\omega) \,|\, \parents(X(A,\omega))) \, p(R(A,\omega) \,|\, X(A,\omega)) \hspace{-4pt} \prod_{\substack{r \in \mathcal{R}_A, \\ 1 \le i \le n_r}} \hspace{-4pt} p(C(A,\omega,r,i) \,|\, R(A,\omega,r)) \right).
  \end{multline}
  Moreover if $\mathcal{G}$ is acyclic,
  \begin{enumerate}
  \item $p(X(A,\omega)=z \,|\, \parents(X(A,\omega))=y) = \Psi^L_{\epsilon_A}(y,z),$
  \item $p(R(A,\omega)=z \,|\, X(A,\omega)=y) = \Psi^S_{q_A}(y,z),$
  \item $p(C(A,\omega,r,i)=z \,|\, R(A,\omega,r)=y) = \Psi^S_{\rho_{(r,i)}}(y,z).$
  \end{enumerate}
\end{theorem}

\begin{proof}  
  Let $V_j = \{X_j,R_j,C_j\}$ denote the
  random variables associated with the $j$-th brick in a 
  topological ordering of $\mathcal{B}$.  We can write
  \begin{align*}
    p(X,R,C) & = \prod_j p(V_j \,|\, V_{k<j}) \\
    & = \prod_j p(X_j \,|\, V_{k<j}) p(R_j \,|\, X_j, V_{k<j}) p(C_j \,|\, R_j, X_j, V_{k<j}).
  \end{align*}
  Based on the definition of the scene generation algorithm, and using the topological ordering constraint we see that
  \begin{align*}
    p(X_j \,|\, V_{k<j}) & = p(X_j \,|\, \parents(X_j)), \\
    p(R_j \,|\, X_j, V_{k<j}) & = p(R_j \,|\, X_j), \\
    p(C_j \,|\, R_j, X_j, V_{k<j}) &= p(C_j \,|\, R_j).
  \end{align*}
  This leads to the expression for $p(X, R, C)$ above.  The expression
  of each term in the factorization using the leaky-OR and selection
  potentials also follows directly from the definition of the scene
  generation algorithm and the topological ordering constraint.
\end{proof}  

\subsection{Factor Graph Construction} 

We now give a general construction for defining a
graphical model that represents a probability distribution over
scenes.

A factor graph (see, e.g., \cite{KFL01}) is an undirected graphical
model that defines a probability distribution using a product of local functions.
Let $\mathcal{F} = (V \cup F, E)$ be a bipartite graph where $V$ is a
set of (discrete) random variables, $F$ is a set of factors and $E$ is a set of
edges connecting variables to factors.
Let $\N(u)$ denote the neighbors of a
node $u$.
Let $x$ denote an outcome for
the variables in $V$.  For $U \subseteq V$ we use $x_U$ to denote the
values of the variables in $U$.
Let $\Psi_f$ be a non-negative potential function associated
with a factor $f \in F$.  The factor graph $\mathcal{F}$ defines a
joint distribution,
$$q(x) = \frac{1}{Z} \prod_{f \in F} \Psi_f(x_{\N(f)}),$$ where the
normalizing constant $Z$ is selected so that $q$ sums to one.

\begin{definition}
  \label{def:factorgraph}
  Let $\mathcal{G}$ be a scene grammar.  Define the factor graph
  $\mathcal{F}(\mathcal{G}) = (V \cup F, E)$ as follows:
\begin{enumerate}
\item[(1)] The variables in $V$ correspond to the variables associated with a scene
  in Definition~\ref{def:rv}.
\vspace{.1cm}  
\item[(2a)] For each $(A,\omega) \in \mathcal{B}$ there is a factor 
  in $F$ with potential $\Psi^L_{\epsilon_A}$ connected to a set of input
  variables $\parents(X(A,\omega))$ and an output variable
  $X(A,\omega)$.
  
\item[(2b)] For each $(A,\omega) \in \mathcal{B}$ there is a factor 
  in $F$ with potential $\Psi^S_{q_A}$ connected to an input variable
  $X(A,\omega)$ and a set of output variables $R(A,\omega)$.

\item[(2c)] For each $(A,\omega) \in \mathcal{B}$, rule $r \in \R_A$, and
  $1 \le i \le n_r$ there is a factor in $F$ with potential
  $\Psi^S_{\rho_{(r,i)}}$ connected to an input variable
  $R(A,\omega,r)$ and a set of output variables $C(A,\omega,r,i)$.
  
\end{enumerate}
\end{definition}

\begin{figure}
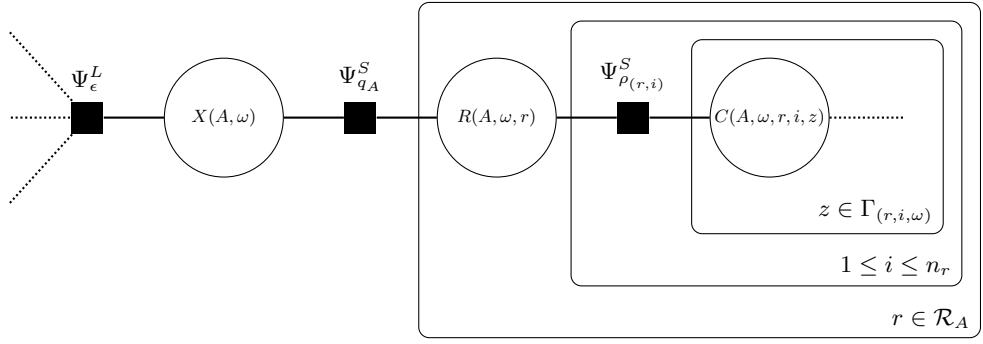

  \centering
      \tikz{ %
        \node (in0) {};
        \node[above=of in0] (in1) {};
        \node[below=of in0] (in2) {};
        \factor[right=of in0] {f1} {$\Psi^L_\epsilon$} {} {} ;
        \node[latent, right=of f1, scale=0.75] (x) {$X(A,\omega)$} ;
        \factor[right=of x] {f2} {$\Psi^S_{q_A}$} {} {} ;
        \node[latent, right=of f2, scale=0.75] (r) {$R(A,\omega,r)$} ;
        \factor[right=of r] {f3} {$\Psi^S_{\rho_{(r,i)}}$} {} {} ;
        \node[latent, right=of f3, scale=0.75] (c) {$C(A,\omega,r,i,z)$} ;
        \node[right=of c] (out) {};
        \plate {pc} {(c)(out)} {$z \in \Gamma_{(r,i,\omega)}$} ;
        \plate {pcf3} {(c)(out)(f3)(f3-caption)(pc.north west)(pc.south east)} {$1 \le i \le n_r$} ;
        \plate {} {(c)(r)(out)(pcf3.north west)(pcf3.south east)} {$r \in \mathcal{R}_A$} ;
        \edge[-] {x} {f1} ;
        \edge[-] {x,r} {f2} ;
        \edge[-] {r,c} {f3} ;
        \dotedge[-] {in0,in1,in2} {f1} ;
        \dotedge[-] {c} {out} ;
      }
      \caption{The block of $\mathcal{F}(\mathcal{G})$ involving
        variables and factors associated with a brick $(A,\omega)$ in
        plate notation.  Variable nodes are represented by circles while
        factor nodes are represented by filled squares.  Each plate
        indicates a subgraph is repeated multiple times, with one copy
        for each index in the plate caption.  The dotted edges
        represent connections between blocks.}
      \label{fig:factorbrick}
\end{figure}

The factor graph $\mathcal{F}(\mathcal{G})$ has a collection of
variables and factors associated with each brick in $\mathcal{B}$.
Figure~\ref{fig:factorbrick} illustrates the part, or \emph{block}, of
$\mathcal{F}(\mathcal{G})$ associated with a single brick.  An example
of a complete factor graph $\mathcal{F}(\mathcal{G})$ for a small grammar is shown
in Figure~\ref{fig:factorgraph}.

Let $p(X,R,C)$ denote the distribution over scenes defined by
$\mathcal{G}$ using the scene generation process in
Section~\ref{sec:model}.  Let $q(X,R,C)$ denote the distribution
defined by $\mathcal{F}(\mathcal{G})$.
Theorem~\ref{thm:factorization} implies the equivalence between the
$p$ and $q$ when $\mathcal{G}$ is acyclic.

\begin{remark}
When $\mathcal{G}$ is acyclic, $q(X,R,C) = p(X,R,C)$ and $Z=1$.  For
cyclic grammars the two distributions are not the same.  In this case
the model defined by $\mathcal{F}(\mathcal{G})$ can be used as an
approximation, or as an alternative model, to the model defined by
$\mathcal{G}$.
\end{remark}

It is important to note that even when $\mathcal{G}$ is acyclic the
corresponding factor graph $\mathcal{F}(\mathcal{G})$ can have
cycles.  This in turn makes inference with $\mathcal{F}(\mathcal{G})$ 
a challenging problem.  Figure~\ref{fig:factorgraph}
illustrates a simple example of an acyclic grammar where the
corresponding factor graph has a cycle.  In this grammar two different
bricks can both generate two other bricks.  This leads to a cycle in
the factor graph even though the grammar is acyclic.  A similar
example is the grammar for scenes with faces in
Section~\ref{sec:exfaces}.  The face grammar is acyclic but multiple
faces can generate multiple common parts and this leads to cycles in
$\mathcal{F}(\mathcal{G})$.

\begin{figure}  
  \centering      

  \tikz{ %
        \begin{scope}
        \factorsmall {f1} {$\Psi^L$} {} {} ;
        \node[latentsmall, right=of f1] (x) {$X$} ;
        \factorsmall[right=of x] {f2} {$\Psi^S$} {} {} ;
        \node[latentsmall, right=of f2] (r) {$R$} ;
        \factorsmall[right=of r] {f3} {$\Psi^S$} {} {} ;
        \node[latentsmall, right=of f3, yshift=+0.5cm] (a1c1) {$C$} ;
        \node[latentsmall, right=of f3, yshift=-0.5cm] (a1c2) {$C$} ;
        \edge[-] {x} {f1} ;
        \edge[-] {x,r} {f2} ;
        \edge[-] {r,a1c1,a1c2} {f3} ;
        \block {p} {(f1)(a1c1)(a1c2)} {$(A,1)$} ;        
        \end{scope}
        
        \begin{scope}[yshift=-3cm]
        \factorsmall {f1} {$\Psi^L$} {} {} ;
        \node[latentsmall, right=of f1] (x) {$X$} ;
        \factorsmall[right=of x] {f2} {$\Psi^S$} {} {} ;
        \node[latentsmall, right=of f2] (r) {$R$} ;
        \factorsmall[right=of r] {f3} {$\Psi^S$} {} {} ;
        \node[latentsmall, right=of f3, yshift=+0.5cm] (a2c1) {$C$} ;
        \node[latentsmall, right=of f3, yshift=-0.5cm] (a2c2) {$C$} ;
        \edge[-] {x} {f1} ;
        \edge[-] {x,r} {f2} ;
        \edge[-] {r,a2c1,a2c2} {f3} ;
        \block {p} {(f1)(a2c1)(a2c2)} {$(A,2)$} ;
        \end{scope}

        \begin{scope}[xshift=6cm]
        \factorsmall {b1f1} {$\Psi^L$} {} {} ;
        \node[latentsmall, right=of b1f1] (x) {$X$} ;
        \factorsmall[right=of x] {f2} {$\Psi^S$} {} {} ;
        \node[latentsmall, right=of f2] (r) {$R$} ;
        \factorsmall[right=of r] {f3} {$\Psi^S$} {} {} ;
        \node[circle,fill=white,draw=white,inner sep=1pt,minimum size=15pt, node distance=0.3, right=of f3, yshift=+0.5cm] (c1) {} ;
        \node[circle,fill=white,draw=white,inner sep=1pt,minimum size=15pt, node distance=0.3, right=of f3, yshift=-0.5cm] (c2) {} ;
        \edge[-] {x} {b1f1} ;
        \edge[-] {x,r} {f2} ;
        \edge[-] {r} {f3} ;
        \block {p} {(b1f1)(c1)(c2)} {$(B,1)$} ;
        \end{scope}

        \begin{scope}[xshift=6cm,yshift=-3cm]
        \factorsmall {b2f1} {$\Psi^L$} {} {} ;
        \node[latentsmall, right=of b2f1] (x) {$X$} ;
        \factorsmall[right=of x] {f2} {$\Psi^S$} {} {} ;
        \node[latentsmall, right=of f2] (r) {$R$} ;
        \factorsmall[right=of r] {f3} {$\Psi^S$} {} {} ;
        \node[circle,fill=white,draw=white,inner sep=1pt,minimum size=15pt, node distance=0.3, right=of f3, yshift=+0.5cm] (c1) {} ;
        \node[circle,fill=white,draw=white,inner sep=1pt,minimum size=15pt, node distance=0.3, right=of f3, yshift=-0.5cm] (c2) {} ;
        \edge[-] {x} {b2f1} ;
        \edge[-] {x,r} {f2} ;
        \edge[-] {r} {f3} ;
        \block {p} {(b2f1)(c1)(c2)} {$(B,2)$} ;
        \end{scope}

        \edge[-] {a1c1} {b1f1} ;
        \edge[-] {a1c2} {b2f1} ;
        \edge[-] {a2c1} {b1f1} ;
        \edge[-] {a2c2} {b2f1} ;
      }

  \vspace{0.5cm}

  \noindent{\parbox{\textwidth}{%
      \hrule
      \vspace{.2cm}
      
      $\Sigma = \{ A, B \}$
      
      $\Omega_A = \Omega_B = \{1, 2\}$.
      
      Rules:
      
      $\begin{array}{lllll}
        (1) & 1.00, & \hspace{-0.20cm} (A,\omega) & \rightarrow  & (B,\uniformRect(1,2)). \\
        (2) & 1.00, & \hspace{-0.20cm} (B,\omega) & \rightarrow  & \emptyset. 
      \end{array}$
      
      \vspace{.2cm}
      \hrule
  }}

  \caption{The factor graph corresponding to a small acyclic grammar.
    The dashed boxes indicate blocks of the factor graph that are
    associated with each brick.  Note that the factor graph has a
    cycle even though the grammar is acyclic.}
  \label{fig:factorgraph}
\end{figure}

For an acyclic grammar the cycles in $\mathcal{F}(\mathcal{G})$
reflect the structure of a data association problem.  As part of
inference we determine the parents and children of each brick in a
scene.  In $\mathcal{F}(\mathcal{G})$ the association is made
explicit, with a random variable that indicates if a brick generates
another one, and factors that require every brick to have the proper
number of children.  


Consider a grammar $\mathcal{G}$ where all the
pose spaces have size at most $K$, the supports
$\Gamma_{(r,i,\omega)}$ have size at most $L$ and the arity of every
rule is bounded by $N$.  Then the number of nodes and edges in the
factor graph $\mathcal{F}(\mathcal{G})$ is $O(|\R|KNL)$.
In practical applications this is a big graph and the explicit
construction of $\mathcal{F}(\mathcal{G})$ requires substantial memory.

In practice (see Section~\ref{sec:exp}) we will also augment
$\mathcal{F}(\mathcal{G})$ by attaching unary factors to each variable
$X(A,\omega)$.  These unary factors can be used as an ``external
field'' and capture local evidence for
bricks.  For example, we can attach a unary factor to the variable
$X(\face,(3,4))$ to encode the image evidence for a face being present
at location $(3,4)$ in the scene.

\section{Efficient Belief Propagation}
\label{sec:infbp}

Thus far we have described a general framework for defining scene
distributions and an algorithm for sampling from these distributions.
Now we turn to the computational problem of inference.  Our goal is to develop a
general purpose computational engine for inference with scene grammars.

Markov Chain Monte Carlo (MCMC) techniques are often used for
inference with graphical models.  However, standard MCMC methods like
Gibbs sampling and Metropolis-Hastings appear to be impractical in our
setting.  The random variables in a scene model are so tightly coupled
that it is very difficult to design MCMC methods that mix in a
reasonable amount of time.

The approach we develop for inference is based on an efficient
implementation of loopy belief propagation, building on the factor
graph representation described in the last section.

The methods we describe here are applicable for inference with a broad
class of graphical models.  One significant challenge with belief
propagation is dealing with high-order factors.  Our work shows it is
possible to use belief propagation on extremelly large factor graphs,
including graphs with high-order factors.  Our techniques can be used
for inference (as an alternative to MCMC) with any factor graph with
binary variables and where each factor is defined using either a
leaky-OR or a selection potential.

Loopy belief propagation (LBP) involves the application of belief
propagation (BP) to graphical models with cycles (see, e.g.,
\cite{W00,YFW01,KFL01}).  LBP aggregates information in a graphical
model by passing messages between neighboring nodes in the graph.  In
this section we show how to implement LBP efficiently for the factor
graphs that represent scene distributions.  We concentrate on the
problem of computing conditional marginals using the
sum-product variant of LBP.


Let $\mathcal{F} = (V \cup F, E)$ be a factor graph.  Let $v \in V$
and $f \in F$ be two neighboring nodes.  The messages sent from $v$ to
$f$ and from $f$ to $v$ are denoted by $\mu_{v \rightarrow f}$ and
$\mu_{f \rightarrow v}$ respectively.  Both messages are non-negative
vectors of dimension given by the number of possible values for $v$.
BP computes a fixed point of the message update equations below.  The
algorithm starts from an arbitrary initialization and repeatedly
updates the messages, either sequentially or in parallel, until
convergence.  After convergence the messages sent to each variable $v
\in V$ are aggregated to obtain a local belief vector $b_v$.

Here we assume all messages are normalized to sum to $1$.
Although this is not strictly necessary, the assumption simplifies the
derivation of the efficient message update equations.  Also, in
practice, one typically normalizes messages to avoid numerical
underflow.

\begin{definition}
  Let $v \in V$ and $f \in \N(v)$.  The message update equation for $\mu_{v \rightarrow f}$ is given by,
  \begin{equation*}
    \mu_{v \rightarrow f}(x_v) = \kappa \prod_{g \in \N(v) \setminus f}\mu_{g  \rightarrow v }(x_v),
  \end{equation*}
  where $\kappa$ is chosen so that $\sum_{x_v}\mu_{v \rightarrow f}(x_v)=1$.
\end{definition}

\begin{definition}
  Let $f \in F$ and $v \in \N(f)$.  The message update equation for $\mu_{f \rightarrow v}$ is given by,
  \begin{equation*}
    \mu_{f \rightarrow v}(x_v) = \kappa \sum_{x_{\N(f) \setminus v}} \Psi_f(x_{\N(f)}) \prod_{u \in \N(f) \setminus v} \mu_{u \rightarrow f}(x_u),
  \end{equation*}
  where $\kappa$ is chosen so that $\sum_{x_v}\mu_{f \rightarrow
    v}(x_v)=1$.
  \label{def:msg_fac_to_var}
\end{definition}

\begin{definition}
The belief of a variable node $v$ is given by,
\begin{equation*}
  b_v(x_v) = \kappa \prod_{f \in \N(v)} \mu_{f \rightarrow v}(x_v),  
\end{equation*}
where $\kappa$ is chosen so that $\sum_{x_v} b_v(x_v)=1$.
\end{definition}

If the factor graph is acyclic, BP converges to a unique fix point
independent of the initialization of the messages.  In this case BP is
equivalent to a dynamic programming algorithm for inference, and the
final beliefs equal the marginal distributions for each variable.
Conditional marginals can also be obtained by ``clamping'' the value
of some variables.  For example, to condition on a value $x_u$ for $u
\in V$ we can set all the messages $\mu_{u \rightarrow f}$ to be an
indicator vector for $x_u$.

The key idea of \emph{loopy} belief propagation is to apply the BP
algorithm to cyclic graphs, such as $\mathcal{F}(\mathcal{G})$, and
treat the final beliefs as approximations to the marginal
distributions.  In the case of cyclic graphs there is no guarantee
that LBP will converge, and the algorithm may converge to different
fixed points depending on the initialization of messages.  Nonetheless
the LBP algorithm has been succesfully used in practice for inference
with a variety of cyclic models (see, e.g., \cite{MWJ99}).

\subsection{Efficient Message Computation}
\label{sec:efficientbp}

The computational complexity of LBP depends crucially on the
complexity of computing messages.  Naively, the run time for computing
a message $\mu_{f \rightarrow v}$ appears to be exponential in the
degree of $f$.  The update involves summing over all joint
configuration of values for the variables in $\N(f) \setminus v$.
However, by exploiting special structure of certain potential
functions, it is sometimes possible to update
messages more efficiently (see, e.g., \cite{Tarlow12}).

The main result in this section is an efficient method for updating
LBP messages for a large class of factor graphs.  The class includes the
factor graphs $\mathcal{F}(\mathcal{G})$ that represent scene
distributions.  The approach we describe updates \emph{all} messages
in a factor graph with $k$ edges in $O(k)$ time total.  Since there
are $2k$ messages to be updated the method is essentially optimal.  The
main result is summarized in Theorem~\ref{thm:bp}.

To derive the efficient message update algorithm we first show how to
update messages from variable nodes, and then show how to update
messages from factor nodes for the two types of factors that appear in
the factor graphs under consideration.  In each case we show how to
compute messages from a node to all of its neighbors in time linear in
the degree of the node.

We start with two simple facts that will be used repeatedly in the
efficient computation of messages.  We delineate them as propositions for
ease of referencing.

\begin{proposition}
  \label{prop:sumTrick}
  For $m \in \mathbb{R}^n$ let $d \in \mathbb{R}^n$ with $d_i =
  \sum_{j \neq i} m_j$.  We can compute $d$ in $O(n)$ time.
\end{proposition}

\begin{proof}
First compute $M = \sum_i m_i$ and then compute $d_i = M-m_i$ for $1
\le i \le n$.
\end{proof}

\begin{proposition}
  \label{prop:prodTrick}
  For $m \in \mathbb{R}^n$ let $d \in \mathbb{R}^n$ with $d_i =
  \prod_{j \neq i} m_j$.  We can compute $d$ in $O(n)$ time.
\end{proposition}

\begin{proof}
If $m_i \neq 0$ for all $i$ compute $M = \prod_i m_i$ and then compute
$d_i = M/m_i$ for $1 \le i \le n$.  If $m_i = 0$ for some $i$ then
$d_j = 0$ for all $j \neq i$ and $d_i$ can be computed explicitly.
\end{proof}

\begin{proposition}
  \label{prop:bp_var}
  Let $v$ be a binary variable in a factor graph.  The messages from
  $v$ to its neighbors can be updated in time linear in the degree of
  $v$.
\end{proposition}

\begin{proof}
  To update $\mu_{v \rightarrow f}(x_v)$ for all $f \in \N(v)$ we use
  Proposition~\ref{prop:prodTrick} twice, once for $x_v=0$ and once
  for $x_v=1$.  Let $m$ be a vector indexed by $f
  \in \N(v)$ with $m_f = \mu_{f \rightarrow v}(x_v)$.  Let $d$ be the
  vector defined by Proposition~\ref{prop:prodTrick}.  Then $\mu_{v
    \rightarrow f}(x_v) \propto d_f$.  Each message can be
  normalized in $O(1)$ time after computation of the un-normalized messages.
\end{proof}

To update the messages from a leaky-OR factor efficiently we use the
structure of a leaky-OR potential in Definition~\ref{def:leaky-or} to
re-write the message update equations.

\begin{restatable}{lemma}{leakyortoz}
  \label{lemma:leakyortoz}
  Let $f$ be a leaky-OR factor with inputs $Y=(Y_1,\ldots,Y_n)$ and
  output $Z$.  The message update equation for $\mu_{f \rightarrow Z}$ can be re-expressed as
  \begin{eqnarray*}
    \mu_{f \rightarrow Z}(0)  & = & (1-\epsilon) \prod_i \mu_{Y_i \rightarrow f}(0), \\
    \mu_{f \rightarrow Z}(1)  & = & 1-\mu_{f \rightarrow Z}(0).
  \end{eqnarray*}
\end{restatable}

\begin{restatable}{lemma}{leakyortoy}
  \label{lemma:leakyortoy}
  Let $f$ be a leaky-OR factor with inputs $Y=(Y_1,\ldots,Y_n)$ and
  output $Z$.  The message update equation for $\mu_{f \rightarrow Y_i}$ can be re-expressed as
  \begin{eqnarray*}
    \mu_{f \rightarrow Y_i}(0) & = & \kappa (\mu_{Z \rightarrow f}(1) + (\prod_{j \neq i} \mu_{Y_i \rightarrow f}(0))((1-\epsilon)(\mu_{Z \rightarrow f}(0)-\mu_{Z \rightarrow f}(1)))), \\
    \mu_{f \rightarrow Y_i}(1) & = & \kappa ( \mu_{Z \rightarrow f}(1) ),
  \end{eqnarray*}
  where $\kappa$ is chosen so that $\mu_{f \rightarrow Y_i}(0) + \mu_{f \rightarrow Y_i}(1) = 1$.
\end{restatable}

The proofs of Lemma \ref{lemma:leakyortoz} and Lemma
\ref{lemma:leakyortoy} can be found in Appendix~\ref{app:msg}.  Using
Lemmas \ref{lemma:leakyortoz} and \ref{lemma:leakyortoy} together with
Proposition~\ref{prop:prodTrick} we obtain the following theorem.

\begin{theorem} \label{thm:leaky-or}
The messages from a leaky-OR factor to its neighbors can be updated in
time linear in the degree of the factor.
\end{theorem}
\begin{proof}
  Let $k = |\N(f)|$.  The message $\mu_{f \rightarrow z}$ can be
  computed in $O(k)$ time using Lemma~\ref{lemma:leakyortoz}.
  Ignoring the normalizing constants the messages $\mu_{f \rightarrow
    y_i}$ can be computed in $O(k)$ time using
  Lemma~\ref{lemma:leakyortoy} and Proposition~\ref{prop:prodTrick}.
  The messages can be normalized in $O(k)$ time total.
\end{proof}

Similarly, to update the messages from a selection factor efficiently we use the
structure of a selection potential in Definition~\ref{def:selection} to
re-write the message update equations.

Here we assume all messages are non-zero and note that in our setting
if this is true for the initial messages it remains true after each
message update.

\begin{restatable}{lemma}{selectiontoy}
  \label{lemma:selectiontoy}
  Let $f$ be a selection factor with input $Y$ and outputs $z=(Z_1,\ldots,Z_n)$.
  The message update equation for $\mu_{f \rightarrow Y}$ can be re-expressed as
  \begin{eqnarray*}
    \mu_{f \rightarrow Y}(0) & = & \kappa (\prod_i \mu_{Z_i \rightarrow f}(0)), \\
    \mu_{f \rightarrow Y}(1) & = & \kappa (\prod_i \mu_{Z_i \rightarrow f}(0)) (\sum_i \theta_{i} \frac{\mu_{Z_i \rightarrow f}(1)}{\mu_{Z_i \rightarrow f}(0)}),
  \end{eqnarray*}
  where $\kappa$ is chosen so that $\mu_{f \rightarrow Y}(0) + \mu_{f \rightarrow Y}(1) = 1$.
\end{restatable}

\begin{restatable}{lemma}{selectiontoz}
  \label{lemma:selectiontoz}
  Let $f$ be a selection factor with input $Y$ and outputs $Z=(Z_1,\ldots,Z_n)$.
  The message update equation for $\mu_{f \rightarrow Z_i}$ can be re-expressed as
  \begin{eqnarray*}
    \mu_{f \rightarrow Z_i}(0) & = & \kappa (\prod_{j \neq i} \mu_{Z_j \rightarrow f}(0)) (\mu_{Y \rightarrow f}(0) + \mu_{Y \rightarrow f}(1) (\sum_{j \neq i} \theta_j \frac{\mu_{Z_j \rightarrow f}(1)}{\mu_{Z_j \rightarrow f}(0)})), \\
    \mu_{f \rightarrow Z_i}(1) & = & \kappa \theta_i \mu_{Y \rightarrow f}(1) (\prod_{j \neq i}\mu_{Z_j \rightarrow f}(0)),
  \end{eqnarray*}
  where $\kappa$ is chosen so that $\mu_{f \rightarrow Z_i}(0) + \mu_{f \rightarrow Z_i}(1) = 1$.
\end{restatable}

The proofs of Lemma \ref{lemma:selectiontoy} and Lemma
\ref{lemma:selectiontoz} can be found in Appendix \ref{app:msg}.
Using Lemmas \ref{lemma:selectiontoy} and \ref{lemma:selectiontoz}
together with Propositions~\ref{prop:sumTrick}
and~\ref{prop:prodTrick} we obtain the following theorem.

\begin{theorem} \label{thm:selection}
The messages from a selection factor to its
neighbors can be updated in time linear in the degree of the factor.
\end{theorem}
\begin{proof}
  Let $k = |\N(f)|$.  The message $\mu_{f \rightarrow y}$ can be
  computed in $O(k)$ time using Lemma~\ref{lemma:selectiontoy}.
  Ignoring the normalization the messages $\mu_{f \rightarrow
    z_i}$ can be computed in $O(k)$ time using
  Lemma~\ref{lemma:selectiontoz}, Proposition~\ref{prop:sumTrick}
  and Proposition~\ref{prop:prodTrick}.  The messages can be
  normalized in $O(k)$ time total.
\end{proof}

Finally we obtain the main result of this section.

\begin{theorem} 
\label{thm:bp}
Consider a factor graph with binary variables and where each factor is
defined by a leaky-OR or selection potential.  The total time required to
update \emph{all} LBP messages is $O(k)$, where $k$ is the number
of edges in the graph.
\end{theorem}
\begin{proof}
  The result follows directly from Proposition~\ref{prop:bp_var} and
  Theorems~\ref{thm:leaky-or} and~\ref{thm:selection}.
\end{proof}

\section{Inference Examples}
\label{sec:infexamples}

In this section we illustrate the results of numerical experiments with
the LBP algorithm described in the last section using the scene
grammars from Section~\ref{sec:examples}.  In these experiments we
condition on the presence of a set of bricks and run LBP on the factor
graph $\mathcal{F}(\mathcal{G})$ to compute the conditional
probability that the remaining bricks are present in the scene.  The
results illustrate how LBP can use the prior knowledge defined by a
scene grammar to ``reason'' about a scene.  

\subsection{Inference with a Face Grammar}
\label{sec:facereason}

\begin{figure}
  \hspace{0.05cm} $\face$  \hspace{2.3cm} $\eye$ \hspace{2.5cm} $\nose$ \hspace{2.0cm} $\mouth$
    \centering
    \includegraphics[width=\textwidth]{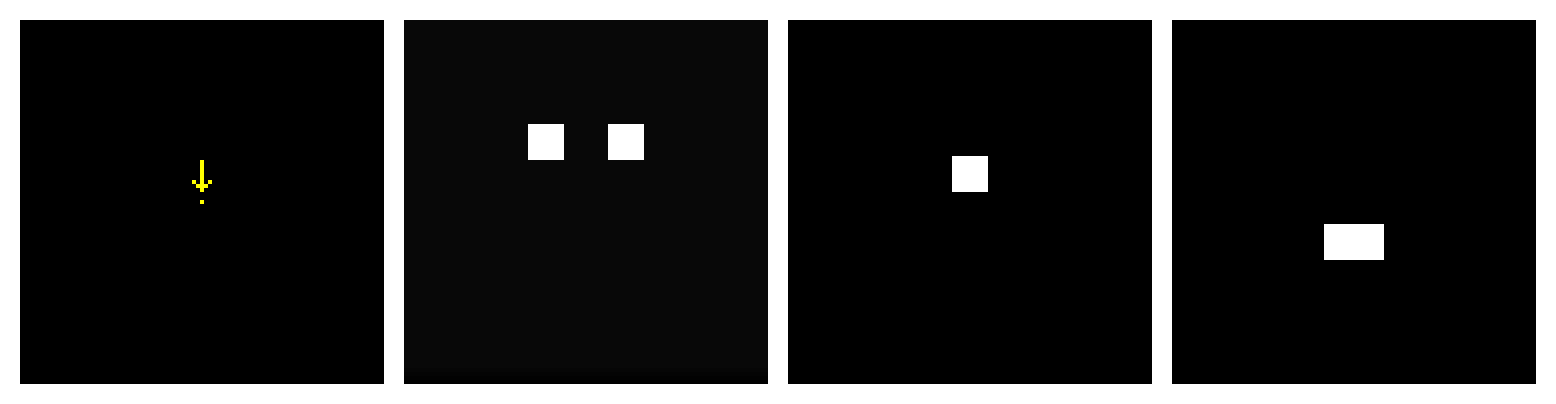} \\
    \includegraphics[width=\textwidth]{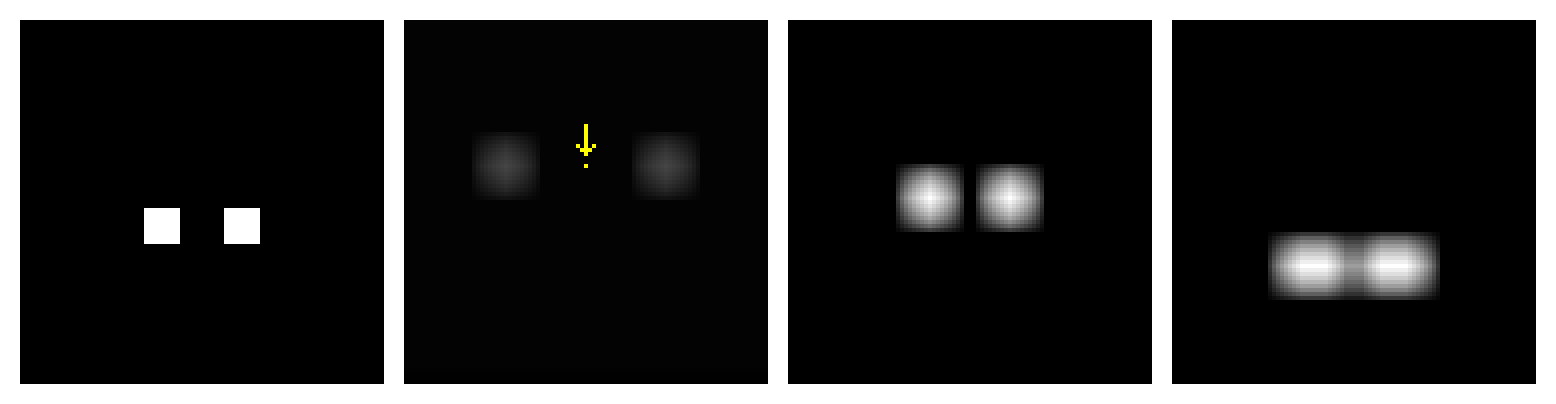} \\
    \includegraphics[width=\textwidth]{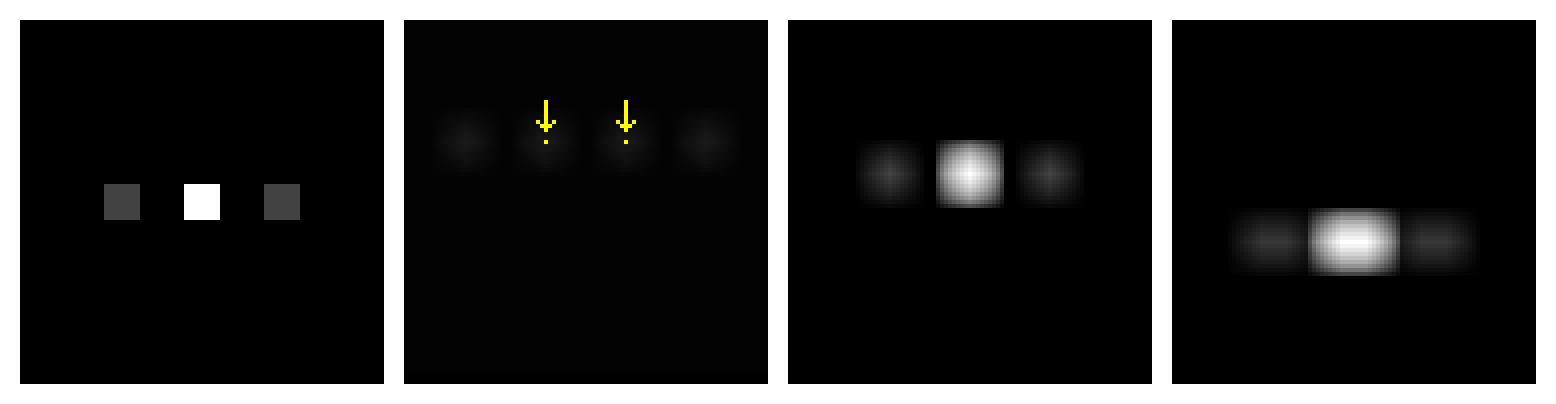} \\
    \caption{Inference with a face grammar.  Each row is a different
      experiment where we condition on the presence of a set of
      bricks.  Each column represents a symbol, with one pixel for
      each pose.  The yellow pixels and yellow arrows indicate the
      bricks that we condition on.  The grayscale values illustrate
      the \emph{estimated} conditional probabilities that each of the
      remaining bricks is in the scene, computed by LBP.  Brighter
      pixels indicate higher probabilities. }
\label{fig:simple_face_bel}
\end{figure}

Here we illustrate the use of LBP for inference with
Grammar~\ref{grammar:face_simple} in Section~\ref{sec:exfaces} to
reason about scenes with faces and part of faces.
Figure~\ref{fig:simple_face_bel} shows the results of inference with
LBP when conditioning on the presence of three different sets of
bricks in the scene.  The resulting algorithm combines \emph{top-down}
(from object to parts) and \emph{bottom-up} (from parts to object)
contextual information to predict unobserved parts of a scene.

The first row of Figure \ref{fig:simple_face_bel} shows the results of
LBP when we condition on the presence of a $\face$ brick in the center
of the scene.  In this case the algorithm uses top-down information
to infer that some face parts must be present as well.  Since the
grammar allows for variability in the location of the parts, there is
a region of plausible locations for each part.

The middle row of Figure \ref{fig:simple_face_bel} shows the results
of LBP when conditioning on the presence of a single $\eye$ brick in
the center of the scene.  In this case the algorithm uses both
bottom-up and top-down information.  Since an $\eye$ seldom appears on
its own, the algorithm infers there is a high probability that a
$\face$ is present in the scene.  Moreover, the distribution over
possible $\face$ poses is bimodal because an $\eye$ brick can either
be the left eye or the right eye of a face.  For each $\face$ brick
that is likely to be present in the scene the algorithm also infers
possible locations for the face parts.  Note that to deduce that there
is a $\nose$ or $\mouth$ in the scene based on observing an $\eye$
requires a chain of reasoning including both bottom-up and top-down
information.

The last row of Figure \ref{fig:simple_face_bel} shows the results of
LBP when two $\eye$ bricks that can both be generated by a single
$\face$ brick are conditioned to be present in the scene.  We see
three regions that are somewhat likely to contain a face.  Since faces
are rare (the self-rooting probabilities are small) the prior model
places higher probability on the event that there is a single $\face$
in the middle of the scene generating both $\eye$ bricks.

\subsection{Inference with a Curve Grammar}
\label{sec:completion}

Here we illustrate the
use of LBP for contour completion using Grammar~\ref{grammar:curve_simple} in
Section~\ref{sec:excurves}.
The \emph{contour completion} problem involves the estimation of a set
of contours, or curves, from a set of observed fragments (see, e.g.\ \cite{WJ97}).  

\begin{figure}
  \centering
  \begin{tabular}{cc}
  \includegraphics[width=5cm]{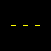} &
  \includegraphics[width=5cm]{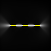} \\ \\
  \includegraphics[width=5cm]{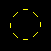} &
  \includegraphics[width=5cm]{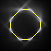}
  \end{tabular}
  \caption{Two contour completion examples.  We condition on the
    presence of a set of $\ink$ bricks (first column) and compute the
    conditional probability that each of the remaining $\ink$ bricks are
    present (second column).  The grayscale values in the second
    column illustrate the conditional probabilities computed by LBP,
    where brighter pixels indicate higher probabilities.}
  \label{fig:completion}
\end{figure}

Figure~\ref{fig:completion} shows the results of two completion
experiments.  In the curve grammar the $\ink$ bricks indicate the grid
points, or pixels, that are part of a curve in the scene.  In each
experiment we condition on the presence of a set of $\ink$ bricks, and use LBP to
compute the conditional probability that each of the remaining $\ink$
bricks are in the scene.

The results in Figure~\ref{fig:completion} show how the LBP algorithm
is able to complete a scene using the prior model defined by the curve
grammar.  In the curve grammar the self-rooting probabilities are
small.  Therefore a scene with a few long curves has higher prior
probability when compared to a scene with many short curves.  In both
examples in Figure~\ref{fig:completion} the LBP algorithm completes
the gaps between observed bricks.  In both cases we can also see the
uncertainty in the path of the completion as it crosses the gaps
between observed bricks.

\section{Learning Model Parameters}
\label{sec:learning}

Recall from Definition \ref{def:psg} that a PSG is defined by a
6-tuple $\mathcal{G} = (\Sigma, \Omega, \R, q, \rho, \epsilon)$.  In
this section we consider the problem of estimating the 
parameters $(q, \rho, \epsilon)$ from a set of observations.

Parameter estimation from fully observed or partially observed scenes
can be addressed using the maximum-likelihood principle and an
approximation to the Expectation-Maximization (EM) algorithm
(\cite{DLR77}).  The approximate EM algorithm described here builds on
the LBP approach for inference described in Section~\ref{sec:infbp}.
The idea of using LBP to approximate the EM algorithm was previously
discussed in \cite{HZW04}.

The problem of learning the structure of a
scene grammar, including the set of symbols $\Sigma$ and productions $\R$, is
significantly different from parameter estimation, and is not
addressed here (see, e.g., \cite{S94,H05,K05} for relevant
approaches).

\subsection{Estimation from Fully Observed Scenes}

Let $\mathcal{G}$ be a PSG with fixed structure $(\Sigma, \Omega, \R)$
and parameters $\Phi = (q,\rho,\epsilon)$.  Let $p(S\,|\,\Phi)$ be the
scene distribution defined by $\mathcal{G}$.  Let $D=\{\,
S_1,\ldots,S_n \,\}$ be $n$ independent samples from $p(S\,|\,\Phi)$.
The maximum-likelihood estimate of $\Phi$ is,
\begin{equation}
  \Phi^* = \argmax_{\Phi} p(D \,|\, \Phi) = \argmax_{\Phi} \prod_{j=1}^n p(S_j \,|\, \Phi)
  \label{eqn:max-like}.
\end{equation}

If the grammar $\mathcal{G}$ is acyclic the factorization
of $p(S)$ described in Section~\ref{sec:factorgraph} can be used to obtain a
closed form solution for $\Phi^*$.  

\begin{proposition}
  Let $\mathcal{G}$ be an acyclic grammar with parameters $\Phi$.  Let
  $D=\{\, S_1,\ldots,S_n \,\}$ be $n$ independent samples from $p(S
  \,|\, \Phi)$.  Let $\{X_j, R_j, C_j\}$ be the
  random variables in Definition~\ref{def:rv} associated with the
  $j$-th sample $S_j$.

  Below let $1 \le j \le n$, $A \in \Sigma$, $\omega \in \Omega_A$, $v \in \{0,1\}$, $r \in \R_A$,
  $1 \le i \le n_r$, $\omega_i \in \Gamma_{(\omega,r,i)}$.
  
  The maximum-likelihood estimate of $\Phi$ is given by,
  
  $$\epsilon_A^* = \frac{\sum_j \sum_{\omega} \alpha(j,A,\omega,1)}
  {\sum_j \sum_{\omega} \sum_v \alpha(j,A,\omega,v)},$$

  $$q_A^*(r) = \frac{\sum_j \sum_{\omega} \beta(j,A,\omega,r)}
  {\sum_j \sum_{\omega} \sum_{r'} \beta(j,A,\omega,r')},$$

  $$\rho^*_{(r,i)}(\omega_i|\omega) = \frac{\sum_j \gamma(j,A,\omega,r,i,\omega_i)}
  {\sum_j \sum_{\omega'_i} \gamma(j,A,\omega,r,i,\omega'_i)},$$
  where 
$$\alpha(j,A,\omega,v) = \mathbbm{1}((\parents(X_j(A,\omega)) = 0) \land (X_j(A,\omega) = v)),$$
$$\beta(j,A,\omega,r) = \mathbbm{1}((X_j(A,\omega) = 1) \land (R_j(A,\omega,r) = 1)),$$
$$\gamma(j,A,\omega,r,i,\omega_i) = \mathbbm{1}((R_j(A,\omega,r) = 1) \land (C_j(A,\omega,r,i,\omega_i) = 1)).$$
\end{proposition}

\begin{proof}
  The result follows directly from the maximization of the
  log-likelihood function, and using the expression for $p(X_j,R_j,C_j
  \,|\, \Phi)$ as a product of potentials in
  Theorem~\ref{thm:factorization}.
\end{proof}

Although the maximum-likelihood estimate for $\Phi$ above was derived for
acyclic grammars, in practice the same estimator can be used for
cyclic grammars as well.

\subsection{Estimation from Partially Observed Scenes}

Now we consider the problem of estimating grammar parameters when we
have incomplete data from a set of scenes.  For example, the available
data may specify the set of bricks that are present in a scene but not
the rules that were used to expand each brick.  Another relevant
example is when the available data reveals only some of the bricks
that are present in a scene.  This is the situation in
Section~\ref{sec:curve} where we estimate the parameters of a curve
grammar from binary images.

As in the case of parameter estimation from fully observed scenes, the
derivation of the estimation procedure here assumes the grammar under
consideration is acyclic.  Nonetheless, we have found that in practice
the same method is effective for cyclic grammars as well.

Let $Z=\{\, S_1,\ldots,S_n \,\}$ be $n$ independent samples from $p(S
\,|\, \Phi)$.  Let $D=\{\, D_1,\ldots,D_n \,\}$ where $D_j$ is a set of
observations from $S_j$.  The maximum-likelihood estimate of $\Phi$
given $D$ is,
\begin{eqnarray}
\Phi^* = \argmax_{\Phi} p(D \,|\, \Phi) = \argmax_{\Phi} \prod_{j=1}^n \sum_{S_j} p(S_j, D_j \,|\, \Phi).
\label{eqn:sum-loglike}
\end{eqnarray}

Here we consider an approximation of the EM algorithm (\cite{DLR77})
for computing $\Phi^*$.  

The EM algorithm starts with an initial set
of parameters and alternates between two steps to generate a sequence
of parameters that increases the likelihood in each iteration and
converges to a critical point of the log-likelihood function.

When $\mathcal{G}$ is an acyclic grammar the two steps of the EM
algorithm are given below.  The derivation of the two steps follows
from the standard definition of the EM algorithm and
Theorem~\ref{thm:factorization}.

Let $1 \le j \le n$, $A \in \Sigma$, $\omega \in \Omega_A$, $v \in \{0,1\}$, $r \in \R_A$,
$1 \le i \le n_r$, $\omega_i \in \Gamma_{(\omega,r,i)}$.

\paragraph{E-step} In the Expectaction step the algorithm computes conditional probabilities
of different events using the current model parameters ($\Phi^t$),

$$\alpha(j,A,\omega,v) = p((\parents(X_j(A,\omega)) = 0) \land (X_j(A,\omega) = v) \,|\, D_j, \Phi^t),$$
$$\beta(j,A,\omega,r) = p(X_j((A,\omega) = 1) \land (R_j(A,\omega,r) = 1) \,|\, D_j, \Phi^t),$$
$$\gamma(j,A,\omega,r,i,\omega_i) = p((R_j(A,\omega,r) = 1) \land (C_j(A,\omega,r,i,\omega_i) = 1) \,|\, D_j, \Phi^t).$$

\paragraph{M-step} In the Maximization step the algorithm computes new
model parameters ($\Phi^{t+1}$),

$$\epsilon_A^{t+1} = \frac{\sum_j \sum_\omega \alpha(j,A,\omega,1)}
{\sum_j \sum_\omega \sum_v \alpha(j,A,\omega,v)},$$

$$q_A^{t+1}(r) = \frac{\sum_j \sum_\omega \beta(j,A,\omega,r)}
{\sum_j \sum_\omega \sum_{r'} \beta(j,A,\omega,r')},$$

$$\rho^{t+1}_{(r,i)}(\omega_i|\omega) = \frac{\sum_j \gamma(j,A,\omega,r,i,\omega_i)}
{\sum_j \sum_{\omega'_i} \gamma(j,A,\omega,r,i,\omega'_i)}.$$

The key challenge in implementing the EM algorithm in our setting is
to compute the conditional probabilities in the E-step.  However,
instead of using the exact conditional probabilities we can use
approximate values computed using LBP.


Let $\mathcal{F}$ be a factor graph defining a distribution $q(x)$.
In addition to approximating marginal
distributions of a single variable, LBP can also be used
to approximate joint marginal distributions of a set of
variables that are all neighbors of a single factor.

\begin{definition}
  Let $f$ be a factor with potential function $\Psi$.  Let $U = \N(f)$.
  Let $\mu$ denote a fixed point of LBP.  The joint belief of $U$ is,
  \begin{equation*}
    b_f(x_U) = \kappa\, \Psi(x_U) \prod_{v \in U} \mu_{v \rightarrow f}(x_v)
    \label{eqn:jointmarg-q}
  \end{equation*}
  where $\kappa$ is chosen so that $\sum_{x_U} b_f(x_U)=1$.
\end{definition}

When the factor graph is acyclic $b_f(x_U)$ matches the
marginal distribution $q(x_U)$.  If the factor graph has cycles (as is
the case for the factor graphs we consider) $b_f(x_U)$ can be used as
an approximation to $q(x_U)$.

Let $\mathcal{F}(\mathcal{G})$ be the factor graph representing 
the distribution defined by $\mathcal{G}$ with parameters $\Phi^t$.
We can use LBP to
approximate the conditional probabilities in the E-step of EM as follows.
Conditioning on $D_j$ involves fixing the messages sent from observed
variables to be indicator vectors.  One can condition on each $D_j$
separately, and run LBP to convergence in each case.

Now consider the conditional probabilities that
appear in the E-step of EM.

\begin{itemize}
\item Let $f$ be the leaky-OR factor in $\mathcal{F}(\mathcal{G})$
  connected to $\parents(X(A,\omega))$ and $X(A,\omega)$.  
  $$\alpha(j,A,\omega,v) = p(\parents(X(A,\omega))=0 \land
  X(A,\omega)=v \,|\, D_j, \Phi^t) \approx b_f(y,v),$$ where $y = (0,\dots,0)$.  
\item Let $f$ be the selection factor in $\mathcal{F}(\mathcal{G})$
  connected to $X(A,\omega)$ and $R(A,\omega)$.  
  $$\beta(j,A,\omega,r) = p(X(A,\omega)=1 \land R(A,\omega,r)=1 \,|\,
  D_j, \Phi^t) \approx b_f(1,z),$$ where $z$ is an indicator vector
  for $r \in \R_A$.
\item Let $f$ be the selection factor in $\mathcal{F}(\mathcal{G})$
  connected to $R(A,\omega,r)$ and $C(A,\omega,r,i)$.
  $$\gamma(j,A,\omega,r,i,\omega_i) = p(R(A,\omega,r)=1 \land
  C(A,\omega,r,i,\omega_i) = 1 \,|\, D_j, \Phi^t) \approx b_f(1,z),$$
  where $z$ is an indicator vector for $\omega_i \in \Gamma_{(\omega,r,i)}$.
  
\end{itemize}

Finally we note that the normalization constant, $\kappa$, in the
definition of a factor belief, $b_f$, can be computed efficiently for
leaky-OR and selection factors using algebraic expressions similar to
the ones used for computing LBP messages in
Section~\ref{sec:efficientbp}.  This makes it possible to approximate
all of the quantities that appear in the E-step of EM efficiently.

\section{Numerical Experiments}
\label{sec:exp}

We now describe the results of computational experiments with two
applications.  The first application involves the reconstruction of
binary contour maps from noisy images.  The second application
involves the detection of faces and parts of faces in photographs.

All of the numerical experiments were performed using a common
implementation of the PSG framework.  In each case a scene grammar is
specified in a high-level language (as the examples in
Section~\ref{sec:examples}).  The implementation automatically
constructs a factor graph representation for a grammar, and can perform
parameter estimation (learning) and inference using LBP.

The PSG framework was implemented in Matlab and C, using a single
thread for computation.  Although the implementation is sequential, it
simulates a flooding schedule for updating BP messages, where all
messages are updated in parallel.  The implementation also uses a damping
factor to improve the convergence of the message update equations.

The running times reported were measured on a personal computer with
an Intel i7 2.5GHz CPU and 16 GB of RAM.  The quantitative experiments
were performed on a cluster of similar computers.  Note that LBP is
highly parallelizable and although we have used a single CPU
implementation of the algorithm, it is possible to use a GPU to
greatly reduce inference time.  It is also possible to use other
message update schedules to reduce the running time of the algorithm.

\subsection{Contour Detection}
\label{sec:curve}

For the experiments with contour detection we use the Berkeley
Segmentation Dataset (BSD500) described in \cite{AMFM11} and follow
the experimental setup in \cite{FOP}.

The BSD500 contains a set of natural images and a collection of object
boundaries marked by human annotators in each image.  We use the
standard split of the dataset with $200$ training images and $200$
test images.  For each image in the BSD500 we use the boundaries
marked by a single human annotator to define a binary image
$B$ representing a \emph{contour map}.

We use a simple imaging model, $p(I \,|\, B)$, to define a real-valued
image $I$ of ``noisy measurements''.  The pixels in $I$ are
conditionally independent given $B$, and $I(i,j)$ is a sample from one
of two possible distributions, $p_0(x)$ or $p_1(x)$, depending on the
value of $B(i,j)$,
$$p(I \,|\, B) = \prod_{(i,j)} p_{B(i,j)}(I(i,j)).$$ 

\begin{figure}
  \centering
  \begin{tabular}{cc}
    (a) & \fbox{\includegraphics[width=3.3in]{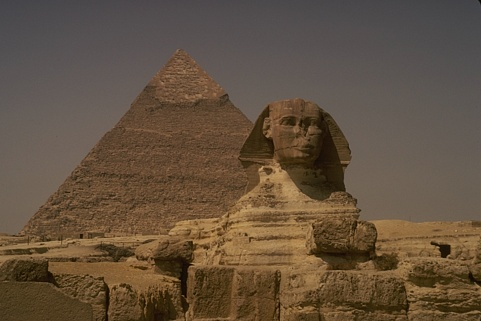}} \\ \\
    (b) & \fbox{\includegraphics[width=3.3in]{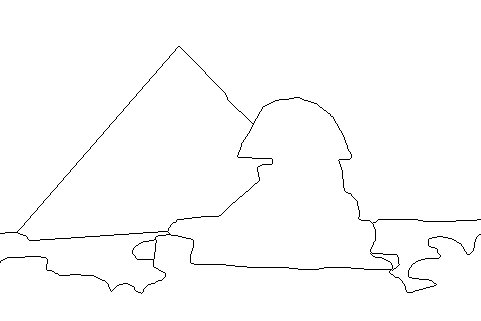}} \\ \\ 
    (c) & \fbox{\includegraphics[width=3.3in]{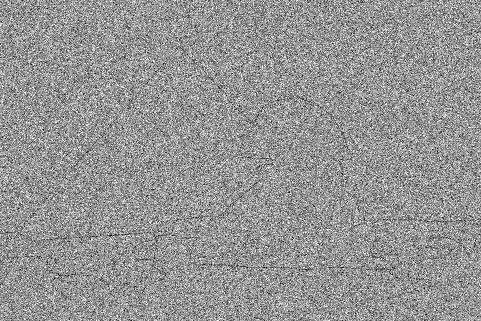}}
  \end{tabular}
  \caption{(a) An image from the BSD500. (b) Ground-truth contour map
    $B$ defined by the object boundaries traced by a human.  (c)
    Real-valued image of noisy measurements $I$.}
  \label{fig:contourdata}
\end{figure}

Figure~\ref{fig:contourdata} shows an example
of an image from the BSD500, the contour map $B$, and the
image $I$ generated using the imaging model.  The
goal of inference is to recover $B$ from $I$.

In all of the experiments we let $p_0(x)$ and $p_1(x)$ be normal
distributions with $\mu_0=150$, $\mu_1=100$, and $\sigma_0 = \sigma_1
= 40$.  In this setting it is impossible to accurately estimate
$B(i,j)$ from $I(i,j)$ alone because $p_0(x)$ and $p_1(x)$ have
significant overlap.  However, we can aggregate information and
disambiguate the problem using a prior model for $B$.

\subsubsection{Scene Grammar}

To specify a prior model for contour maps we use the PSG contour model
defined by Grammar~\ref{grammar:curve_simple_learned}.  This grammar
is similar to Grammar~\ref{grammar:curve_simple} in
Section~\ref{sec:excurves}, but with model parameters estimated from
contour maps in the BSD500 training set.

The $\ink$ bricks in a scene represent the trace of a set of curves in
the image grid, and define a contour map $B$.  That is, we set $B(i,j)
= 1$ iff $(\ink,(i,j)) \in S$.

Note that a ground-truth contour map $B$ only specifies the set of
$\ink$ bricks in a scene.  We do not have observations for the
$\curve$ bricks or the rules used to generate a scene.  Therefore we
used the approximate EM procedure described in
Section~\ref{sec:learning} to estimate model parameters.

\begin{grammar} 
  \noindent{\parbox{\textwidth}{%

      \hrule
      \vspace{.2cm}

  $\Sigma = \{ \curve, \ink \}$.
  
  $\Omega_{\curve} = [N] \times [M] \times [8]$.
  
  $\Omega_{\ink} = [N] \times [M]$.
      
  Rules:
  
  $\begin{array}{llll}
    0.647, & \hspace{-0.20cm} (\curve,(x,y,\theta)) & \rightarrow  & (\ink, \delta((x,y))),
    (\curve,\delta(((x,y)+\round(T_{\theta}(1,0)),\theta))). \\
    0.147, & \hspace{-0.20cm} (\curve,(x,y,\theta))  & \rightarrow & (\ink, \delta((x,y))),
    (\curve,\delta(((x,y)+\round(T_{\theta-1}(1,0)),\theta))). \\
    0.152, & \hspace{-0.20cm} (\curve,(x,y,\theta)) & \rightarrow & (\ink, \delta((x,y))),
    (\curve,\delta(((x,y)+\round(T_{\theta+1}(1,0)),\theta))). \\
    0.019, & \hspace{-0.20cm} (\curve,(x,y,\theta)) & \rightarrow & (\curve, \delta((x,y,\theta-1))). \\
    0.019, & \hspace{-0.20cm} (\curve,(x,y,\theta)) & \rightarrow & (\curve, \delta((x,y,\theta+1))). \\
    0.012, & \hspace{-0.20cm} (\curve,(x,y,\theta)) & \rightarrow & (\ink, \delta((x,y))). \\
    1.00, & \hspace{-0.20cm} (\ink,(x,y)) & \rightarrow & \emptyset.
  \end{array}$
  
  $\epsilon_{\curve} = 4.28 \times 10^{-5}$, \\
  $\epsilon_{\ink} =  1.87 \times 10^{-12}$.

  \vspace{.2cm}
  \hrule
  
  }}

  \vspace{.1cm}

  The function $T_{\theta}$ denotes a rotation in the plane by a
  discrete angle $\theta$ and $\round$ maps a point in the plane to
  the nearest grid point. 

  \caption{The PSG contour model (see Section~\ref{sec:excurves}).
    The model parameters were estimated using EM with contour maps from the BSD500.}
  \label{grammar:curve_simple_learned}
\end{grammar}

\subsubsection{Inference}

To estimate $B$ from $I$ we incorporate the imaging model into the
factor graph representation of the PSG contour model.  We run LBP on
this factor graph to compute conditional probabilities $p(B(i,j)=1
\,|\, I)$.

\begin{figure}
  \centering
  \begin{tabular}{ccc}
    \fbox{\includegraphics[width=0.29\textwidth]{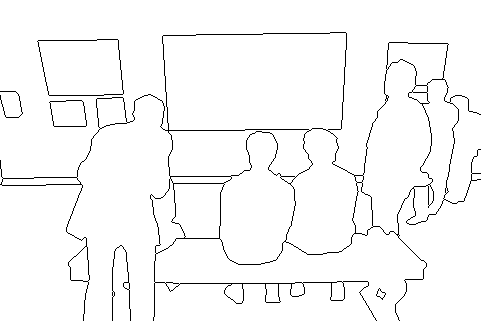}} &
    \fbox{\includegraphics[width=0.29\textwidth]{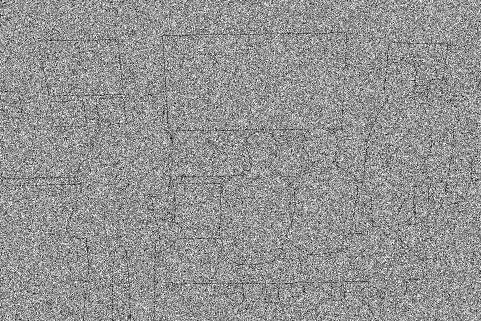}} &
    \fbox{\includegraphics[width=0.29\textwidth]{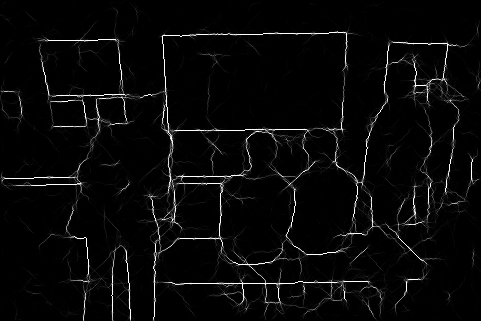}} \\ \\
    \fbox{\includegraphics[width=0.29\textwidth]{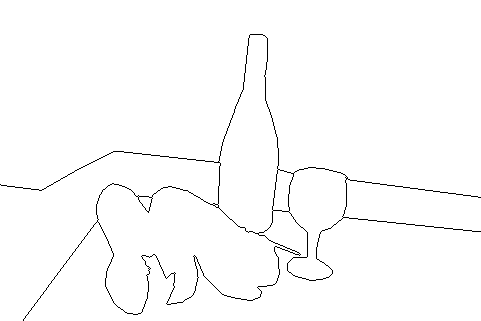}} &
    \fbox{\includegraphics[width=0.29\textwidth]{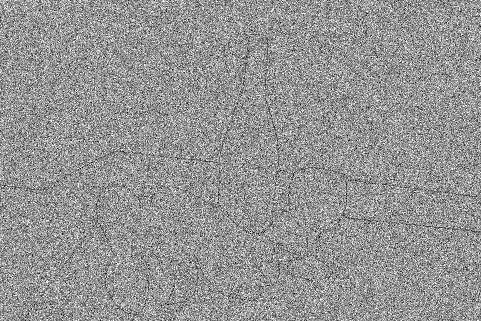}} &
    \fbox{\includegraphics[width=0.29\textwidth]{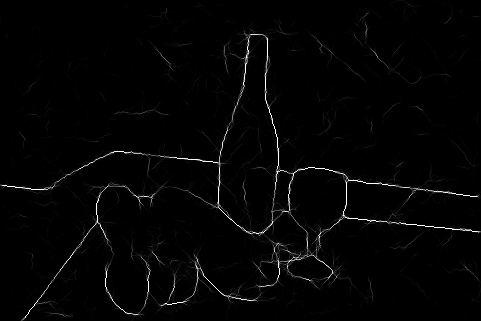}} \\ \\
    \fbox{\includegraphics[width=0.29\textwidth]{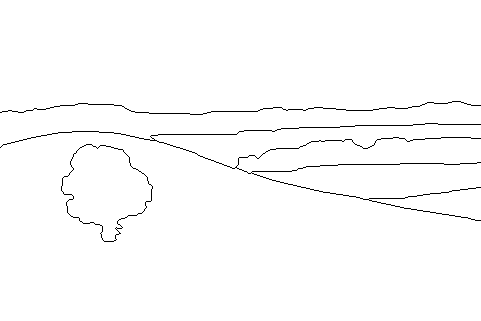}} &
    \fbox{\includegraphics[width=0.29\textwidth]{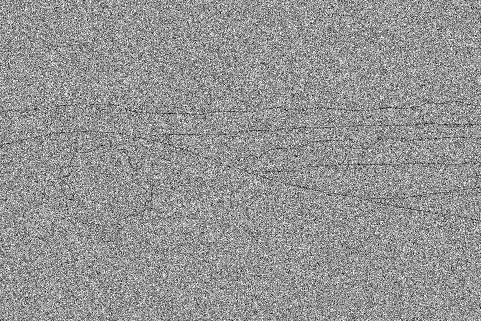}} &
    \fbox{\includegraphics[width=0.29\textwidth]{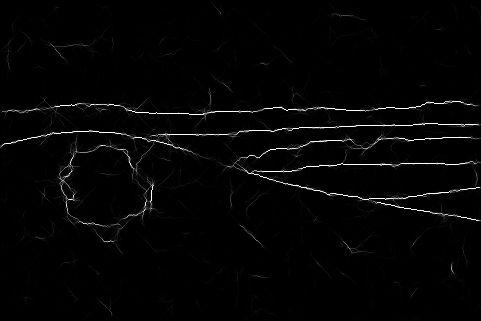}} \\ \\
    \fbox{\includegraphics[width=0.29\textwidth]{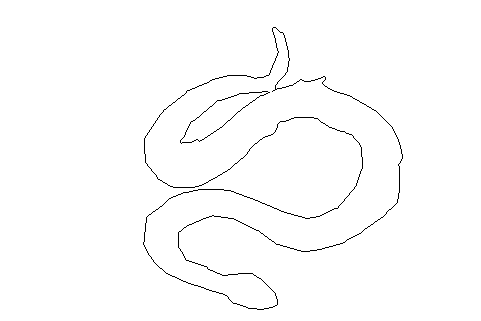}} &
    \fbox{\includegraphics[width=0.29\textwidth]{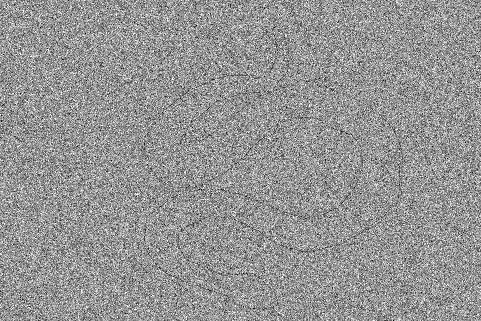}} &
    \fbox{\includegraphics[width=0.29\textwidth]{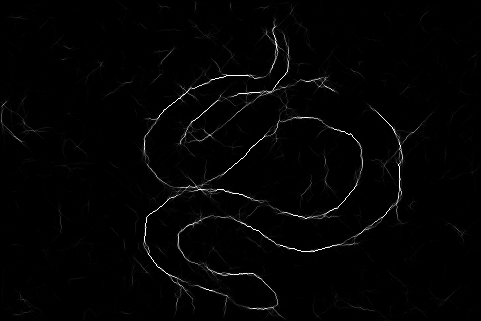}} \\ \\
    (a) & (b) & (c)
  \end{tabular}    
\caption{Contour detection results on four examples from the BSD500
  test set. (a) Ground-truth contour maps $B$.  (b) Noisy measurements $I$.
  (c) Estimated conditional probabilities $p(B(i,j)=1 \,|\, I)$
  with brighter values indicating a higher probability.}
\label{fig:orientedRes}
\end{figure}

Figure~\ref{fig:orientedRes} shows the result of inference on several
examples from the BSD500 test set.  These results illustrate how we
can recover good contour maps despite the ambiguous local
observations.

By Bayes' rule $p(S \,|\, I) \propto p(S) p(I \,|\,S)$.  Recall that
$B(i,j) = 1$ iff $(\ink,(i,j)) \in S$.  In the factor graph
$\mathcal{F}(\mathcal{G})$ there is a binary variable $X(\ink,(i,j))$
for each pixel $(i,j)$.  We attach an additional unary
factor to each of these variables, with potential $\Psi(x) =
p_x(I(i,j))$.  With these additional factors the graphical model
represents the conditional distribution $p(S \,|\, I)$.


The factor graph representing the conditional distribution $P(S \,|\,
I)$ for an $N$ by $M$ image has $O(NM)$ binary variables and $O(NM)$
edges.  Therefore, using the approach in Section~\ref{sec:infbp}, one
iteration of LBP takes $O(NM)$ time (linear in the size of the image).  With our
implementation of the PSG framework running LBP to convergence
on a $481 \times 321$ image took on average 1.5 hours.

\subsubsection{Quantitative Evaluation}

For a quantitative evaluation we compare the result of thresholding
the estimated conditional probabilities $p(B(i,j) = 1 \,|\, I)$ to the
ground-truth contour maps.  The results were evaluated using the 200
test images in the BSD500.  Each threshold leads to a total number of
correct (true positives) and incorrect (false positives) contour
pixels detected, counted over all images in the test set.

\begin{figure}    
  \centering
  \includegraphics[width=0.85\textwidth]{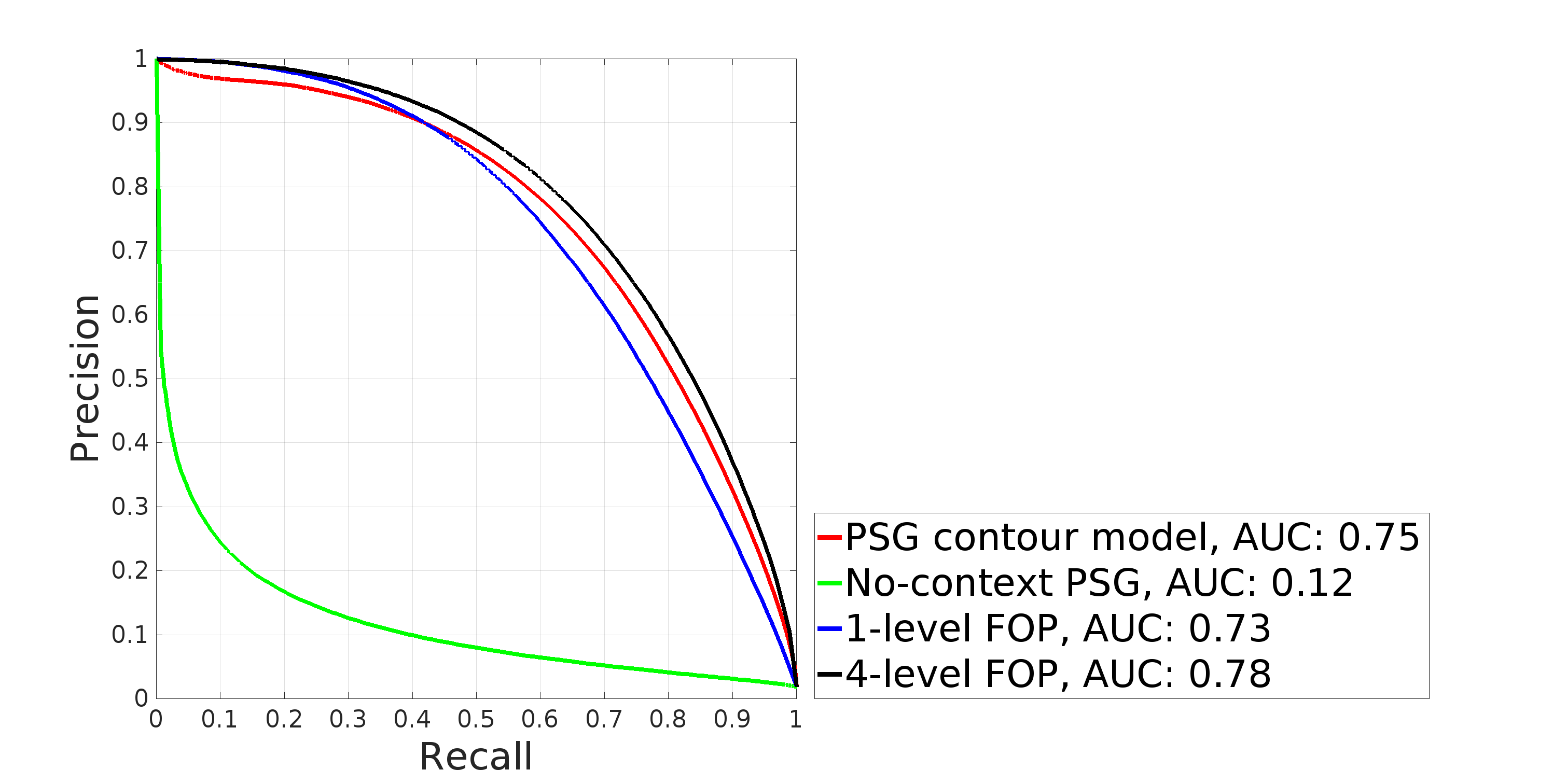}	
  \caption{Precision-recall curves of different models for contour
    detection.  The AUC numbers indicate the area under the
    curves.}
  \label{fig:contour_prec_rec}
\end{figure}

Figure~\ref{fig:contour_prec_rec} shows the precision-recall curve
obtained using different thresholds for detecting contour pixels, and
also the results obtained with other methods in the same dataset.
The figure also summarizes the area under the
precision-recall curves (AUC) for each method.

To measure the importance of context for detecting contour pixels we
evaluate a trivial PSG model with only the $\ink$ bricks and the rule
$\ink \rightarrow \emptyset$.  We refer to this model as the
No-Context PSG.

We also include the results obtained with the Field-of-Patterns (FOP)
models in \cite{FOP}.  The 1-level FOP model captures the statistics
of 3x3 patterns in a binary image.  The 4-level FOP model captures
similar statistics of a multiscale representation of the image.

The quantitative results in Figure~\ref{fig:contour_prec_rec} indicate
the importance of context for restoring binary contour maps.  The
evaluation also shows how the grammar based model and general
inference method described in this paper lead to results that are
comparable to state-of-the-art methods developed for this specific
application.

\subsection{Face Detection}
\label{sec:face}

\begin{figure}
  \centering
  \begin{tabular}{ccc}
  \includegraphics[width=0.22\textwidth]{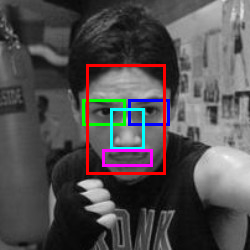} & 
  \includegraphics[width=0.22\textwidth]{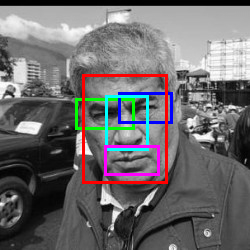} &
  \includegraphics[width=0.22\textwidth]{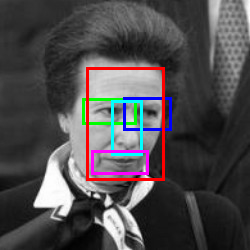}
  \end{tabular}
  \caption{Examples of images in the LFW dataset.  The images
    were annotated with bounding boxes for the face (red), left
    eye (green), right eye (blue), nose (cyan), and mouth
    (magenta).}
  \label{fig:face_examples_lfw}
\end{figure}

\begin{figure}
  \centering
    \includegraphics[width=0.6\textwidth]{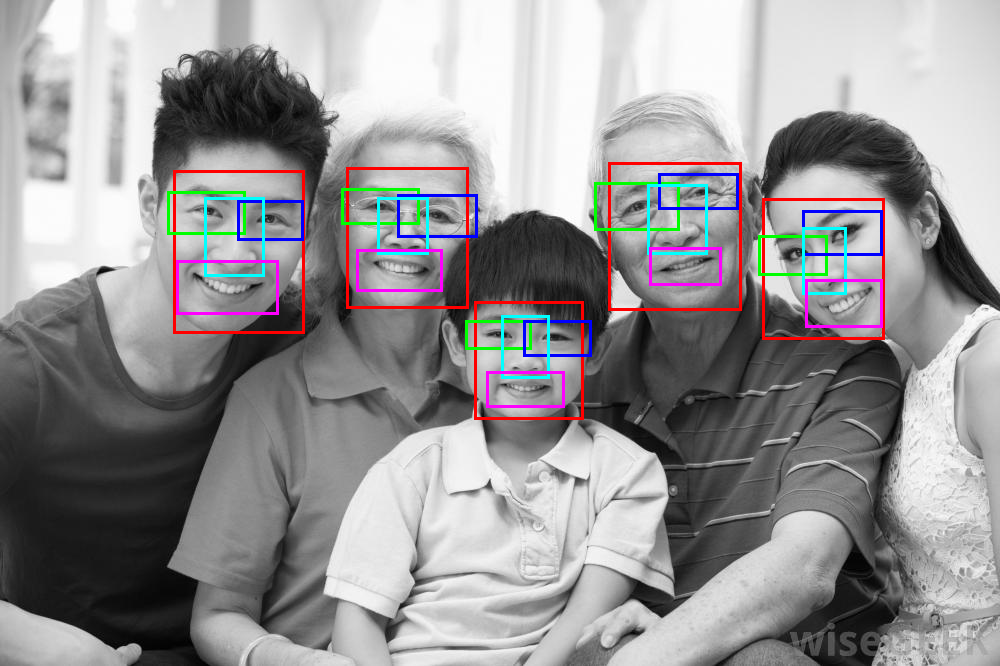}
  \caption{Example of an image in the Portraits dataset.  The images
    were annotated with bounding boxes for each face (red), left eye
    (green), right eye (blue), nose (cyan), and mouth (magenta).}
  \label{fig:face_examples_portraits}
\end{figure}

Now we consider the problem of detecting faces and parts of faces in
images.  We use two datasets for the experiments with face detection:

\begin{enumerate}
\item \emph{Labeled Faces in the Wild (LFW) dataset:}
  The LFW dataset was introduced in
  \cite{LFW}.  The dataset contains images with a single face.
  For the experiments in this
  section we randomly select $200$ images for training, and $100$ images
  for testing.  We manually annotated each image with bounding box
  information for the face, left eye, right eye, nose, and mouth.
  Examples of labeled images are shown
  Figure~\ref{fig:face_examples_lfw}.

\item \emph{Portraits dataset:} To study face detection in more
  complex scenes, we use a dataset of $40$ images of family and class
  portraits collected from the Internet.  We used the search strings
  ``family portraits'', ``class portraits'' and ``school portraits''
  on Google in November 2016 to collect images for the dataset.  The
  images were manually annotated with bounding box information for
  each face, left eye, right eye, nose, and mouth.  One example of a
  labeled image is shown in Figure~\ref{fig:face_examples_portraits}.
  The average number of faces per image in the dataset is $5.9$.  
\end{enumerate}

\subsubsection{Face Detection Grammar}

The PSG face model used in the face detection experiments is defined by
Grammar~\ref{grammar:face_scale}.  This model is similar to
Grammar~\ref{grammar:face_simple} in Section~\ref{sec:examples}.
However, the pose spaces in the model used here include scale
information to represent objects of different sizes.  The model used
here also differentiates between parts of faces and objects that
simply look like parts of faces but can appear outside the context of
a face.  

\begin{grammar}
  \noindent{\parbox{\textwidth}{%

      \hrule
      \vspace{.2cm}

      $\Sigma = \{\, \face, \lefteye, \righteye, \nose, \mouth, \facelike, \lefteyelike, \righteyelike, \noselike, \mouthlike \,\}$
      
      $\forall A \in \Sigma,\; \Omega_A = \{\, (s,i,j) \mid s \in [L],\, (i,j) \in [N_s] \times [M_s] \,\}$. 
         
      Rules: 

      $\begin{array}{llll}
        1.0, & (\face,\omega) & \rightarrow & (\facelike,\delta(\omega)), \\
        & & & (\lefteye,\discrete(\theta_1(\omega))), \\
        & & & (\righteye,\discrete(\theta_2(\omega))), \\
        & & & (\nose,\discrete(\theta_3(\omega))), \\
        & & & (\mouth,\discrete(\theta_4(\omega))), \\
        1.0, & (\lefteye,\omega) & \rightarrow & (\lefteyelike,\delta(\omega)) \\
        1.0, & (\righteye,\omega) & \rightarrow & (\righteyelike,\delta(\omega)) \\
        1.0, & (\nose,\omega) & \rightarrow & (\noselike,\delta(\omega)) \\
        1.0, & (\mouth,\omega) & \rightarrow & (\mouthlike,\delta(\omega)) \\
        1.0, & (\lefteyelike,\omega) & \rightarrow & \emptyset \\
        1.0, & (\righteyelike,\omega) & \rightarrow & \emptyset \\
        1.0, & (\noselike,\omega) & \rightarrow & \emptyset \\
        1.0, & (\mouthlike,\omega) & \rightarrow & \emptyset
      \end{array}$

      $\epsilon_{\face} = 10^{-4}$ 

      $\epsilon_{\lefteye} = \epsilon_{\righteye} = \epsilon_{\nose} = \epsilon_{\mouth} = 10^{-12}$
      
      $\epsilon_{\facelike} = \epsilon_{\lefteyelike} = \epsilon_{\righteyelike} = \epsilon_{\noselike} = \epsilon_{\mouthlike} = 10^{-4}$.

      \vspace{.2cm}
      \hrule
  }}
  \caption{The PSG face model.  The grammar defines scenes with faces
    and parts of faces of different sizes.  A pose $(s,i,j)$ specifies
    a size $s$ and a location $(i,j)$ in a grid.  The parameters
    $\theta_i(\omega)$ define categorical distributions for the pose
    of a part relative to the pose of a face.}
  \label{grammar:face_scale}
\end{grammar}

The PSG face model has a different symbol for each face part,
including separate symbols for the left and right eyes.  To model
visual appearance of objects the grammar includes additional
symbols that represent \emph{templates}.  A template symbol is denoted
with the $\like$ prefix.  The set of template symbols is denoted by
$\Sigma_T$.

Templates can appear either in the context of a face or in the
background.  This makes it possible to differentiate between an eye
(nose, etc.) and an object that simply looks like an eye (nose, etc.).
The distinction is important to supress false positive
detections that arise from background clutter (see
\cite{CJZBG11}).  The parts of the face have very low self-rooting
probability ($10^{-12}$) while the corresponding template symbols have
a relatively higher self-rooting probability ($10^{-4}$).  As a
result, when objects appear out of context they are interpreted as a
self-rooted template.

The poses in the PSG face model include position and scale information
to capture objects of different sizes.  The pose spaces are defined so
that larger objects are localized with lower spatial resolution.  This
allows for representing objects of different sizes efficiently.

Let $L$, $K$, $N$, and $M$ be positive integers.  For each $A \in
\Sigma$ we have
$$\Omega_A = \{\, (s,i,j) \mid s \in [L],\, (i,j) \in [N_s] \times [M_s]
\,\},$$
$$N_s = \left\lfloor \frac{N}{2^{s/K}} \right\rfloor,\;\; M_s =
\left\lfloor \frac{M}{2^{s/K}} \right\rfloor.$$

A pose $(s,i,j) \in \Omega_A$ specifies a scale $s \in [L]$
and a position $(i,j)$ in a $N_s \times M_s$ grid.  Objects in
scale $s$ have size proportional to $2^{s/K}$ and are localized in a
grid with spacing $2^{s/K}$.  With this construction
$|\Omega_A| = O(KNM)$, independent of $L$.

To represent object sizes accurately we use $K=8$ in all of our
experiments.  The value of $N$ and $M$ were defined by the width and
height of an image image divided by the size of the local image
features used for modeling the appearance of templates.

The conditional pose distributions for each part of a face in the PSG
face model are categorical distributions with parameters
$\theta_i(\omega)$ for $1 \le i \le 4$.
The pose distributions are defined using relative
poses to impose scale and shift (translation) invariance.
The distributions were estimated from the frequencies of
relative positions and sizes between objects in the LFW
training dataset.


The pose of an object specifies a location and scale, and we
approximate the extent of the object by a rectangular box of varying
size but with fixed aspect ratio.  It is also possible to define
objects using a collection of smaller parts, like corners and edges.
In this case the location of the smaller parts can be used to compute
a more accurate bounding box for the object.

\subsubsection{Face Data Model}
\label{sec:faceData}

To define a data model we use templates that capture the appearance of
local image features.  For the experiments in this section we define
template responses using histogram-of-gradient (HOG) features in a
feature pyramid (see \cite{DT05} and \cite{FGMR10}).

\begin{figure}
  \centering
  \begin{tabular}{ccccc}     
  \includegraphics[scale=0.2]{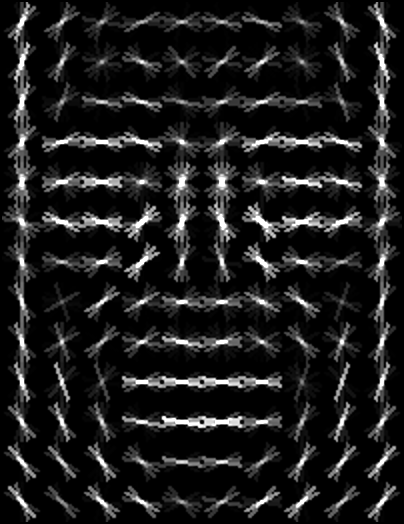} &
  \includegraphics[scale=0.2]{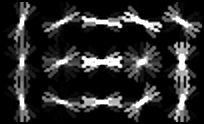} &
  \includegraphics[scale=0.2]{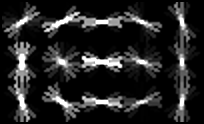} &
  \includegraphics[scale=0.2]{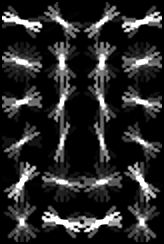} &
  \includegraphics[scale=0.2]{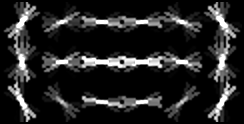} \\
  $\facelike$ & $\lefteyelike$ & $\righteyelike$ & $\noselike$ & $\mouthlike$
  \end{tabular}
  \caption{Visualization of the HOG templates used for face
    detection.  Note that the templates for the $\lefteyelike$
    and $\righteyelike$ symbols are subtly different, indicating
    there is a visual difference between the two parts. Also
    note that the $\mouthlike$ template shares some similarities
    to both the $\lefteyelike$ and $\righteyelike$ filters.}
  \label{fig:hog_filters}
\end{figure}

We trained templates for each face part and a template for the whole
face using the publicly-available code from \cite{voc-release4}.  The
templates are trained to have positive responses on a set of positive
examples and negative responses on a set of negative examples.  We
used the annotations in the LFW training set to define positive
examples for each object type.  The negative examples were taken from
images in the PASCAL VOC 2012 dataset (\cite{pascal-voc-2012}).
Figure \ref{fig:hog_filters} shows the resulting HOG templates that
were used for the experiments with face detection.

For a template symbol $A \in \Sigma_T$ let $H_A$ be a template
associated with $A$ and $H_A(I,\omega)$ be the response of $H_A$ at
the position and scale specified by $\omega$ within $I$.  The data we
use for inference of a scene $S$ is the collection of template
responses,
$$H = \{ H_A(\omega, I) \,|\, A \in \Sigma_T, \omega \in \Omega_A
\}.$$

To derive a tractable model for $p(H \,|\, S)$ we use the 
approach described in \cite{GJ96} to define a model for a family of local tests.

We assume the template responses are conditionally independent
given $S$,
$$p(H \,|\, S) = \prod_{A \in \Sigma_T} \prod_{\omega \in \Omega_A}
p(H_A(\omega, I) \,|\, S).$$
We also assume the conditional distributions $p(H_A(\omega, I)=v \,|\, S)$
depend only on whether or not $(A,\omega)$ is present in $S$,
$$p(H_A(\omega, I) = v\,|\, S) =
\begin{cases}
  p_{A,0}(v) & (A,\omega) \not\in S, \\
  p_{A,1}(v) & (A,\omega) \in S.
\end{cases}$$

\begin{figure}
  \centering
  \includegraphics[width=0.45\textwidth]{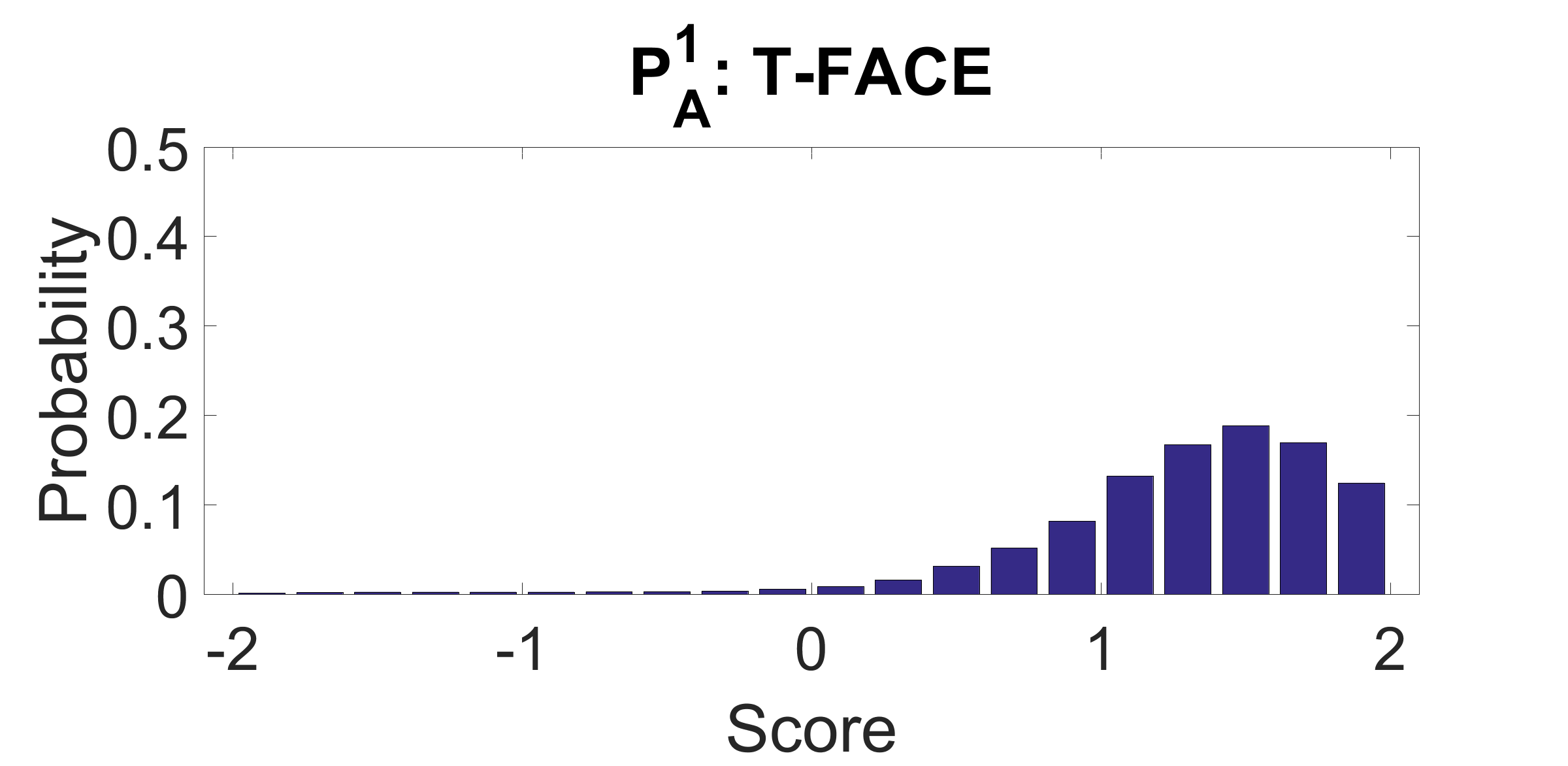}
  \includegraphics[width=0.45\textwidth]{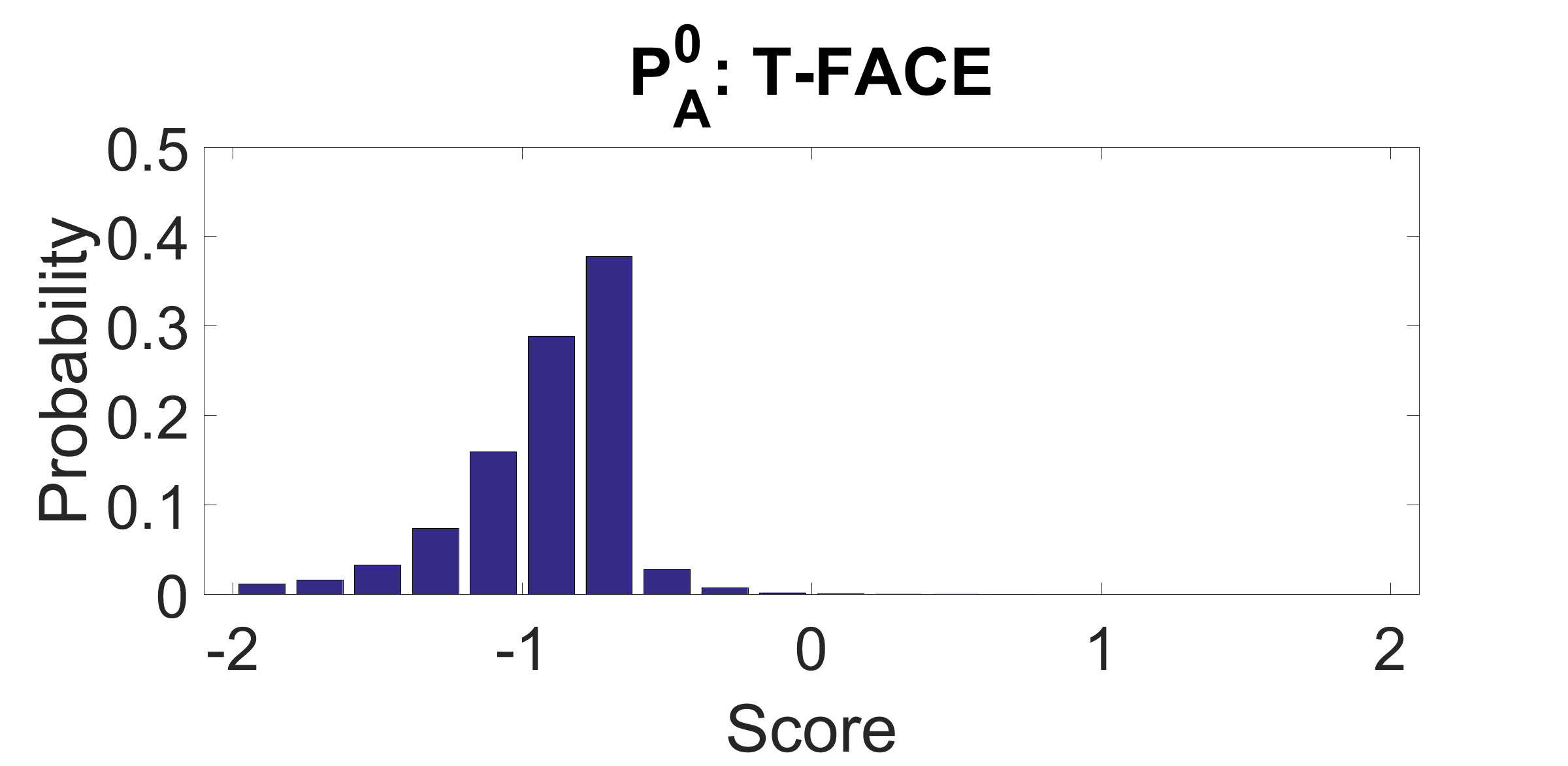}
  \includegraphics[width=0.45\textwidth]{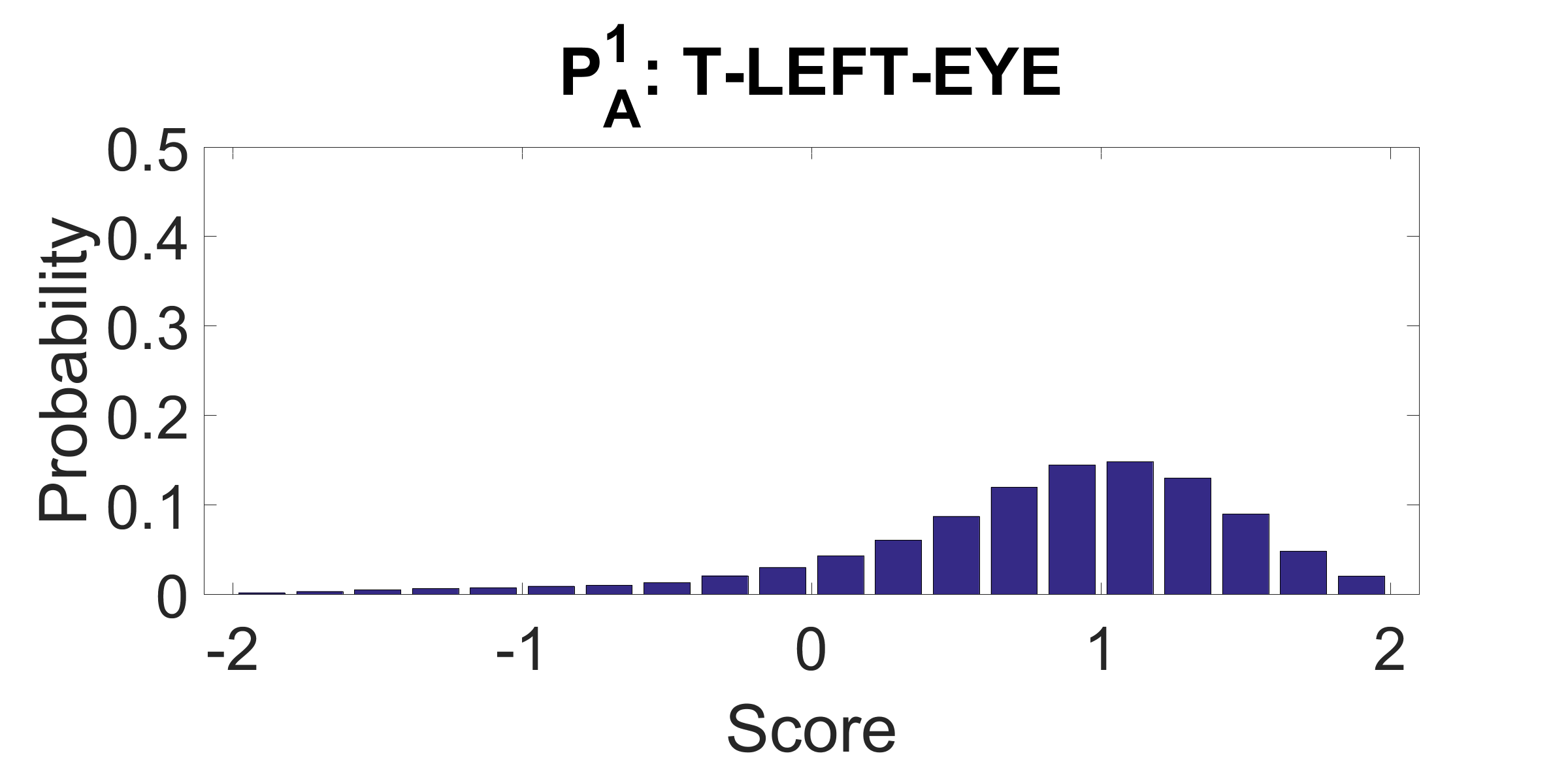}
  \includegraphics[width=0.45\textwidth]{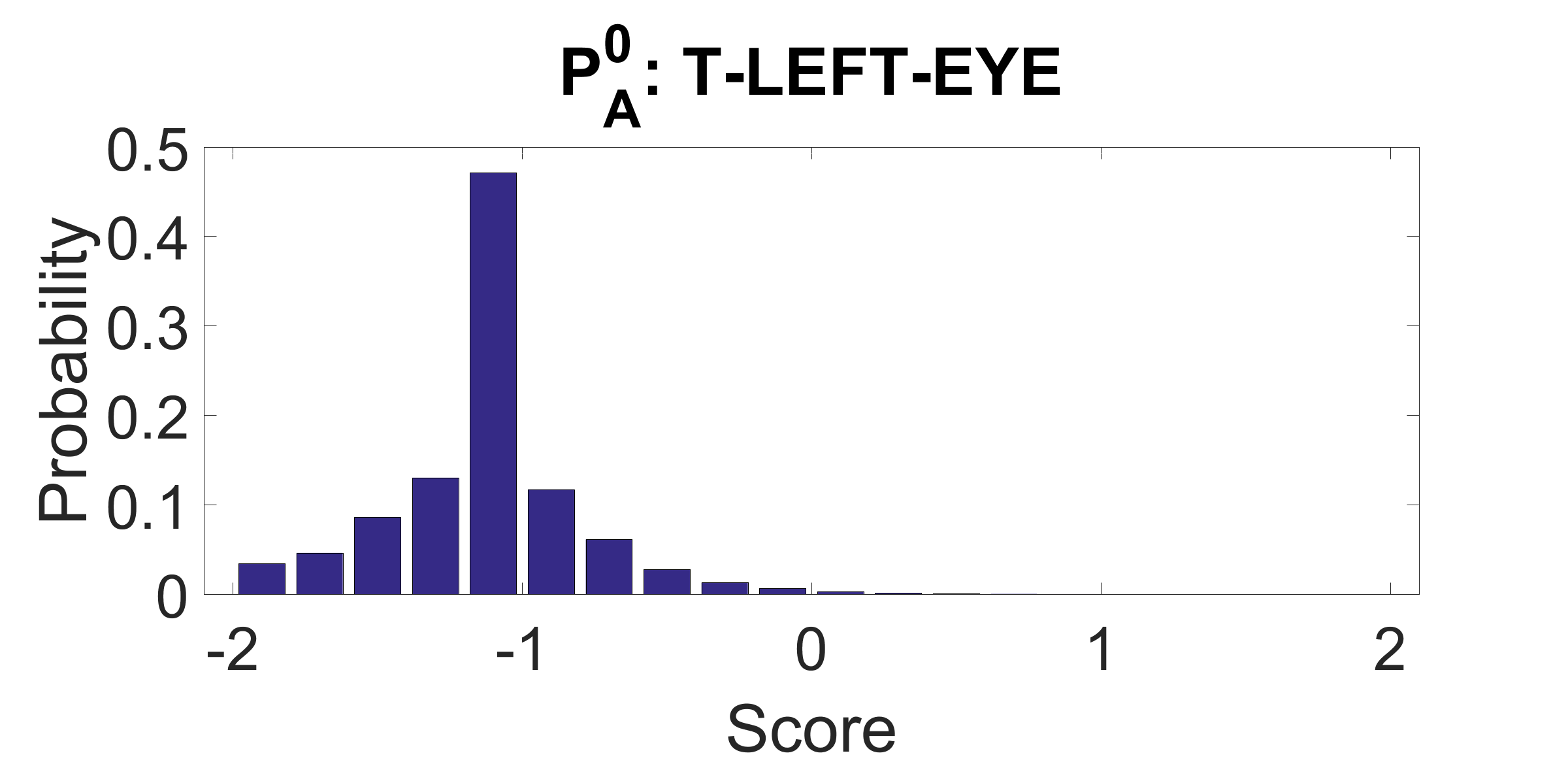}
  \includegraphics[width=0.45\textwidth]{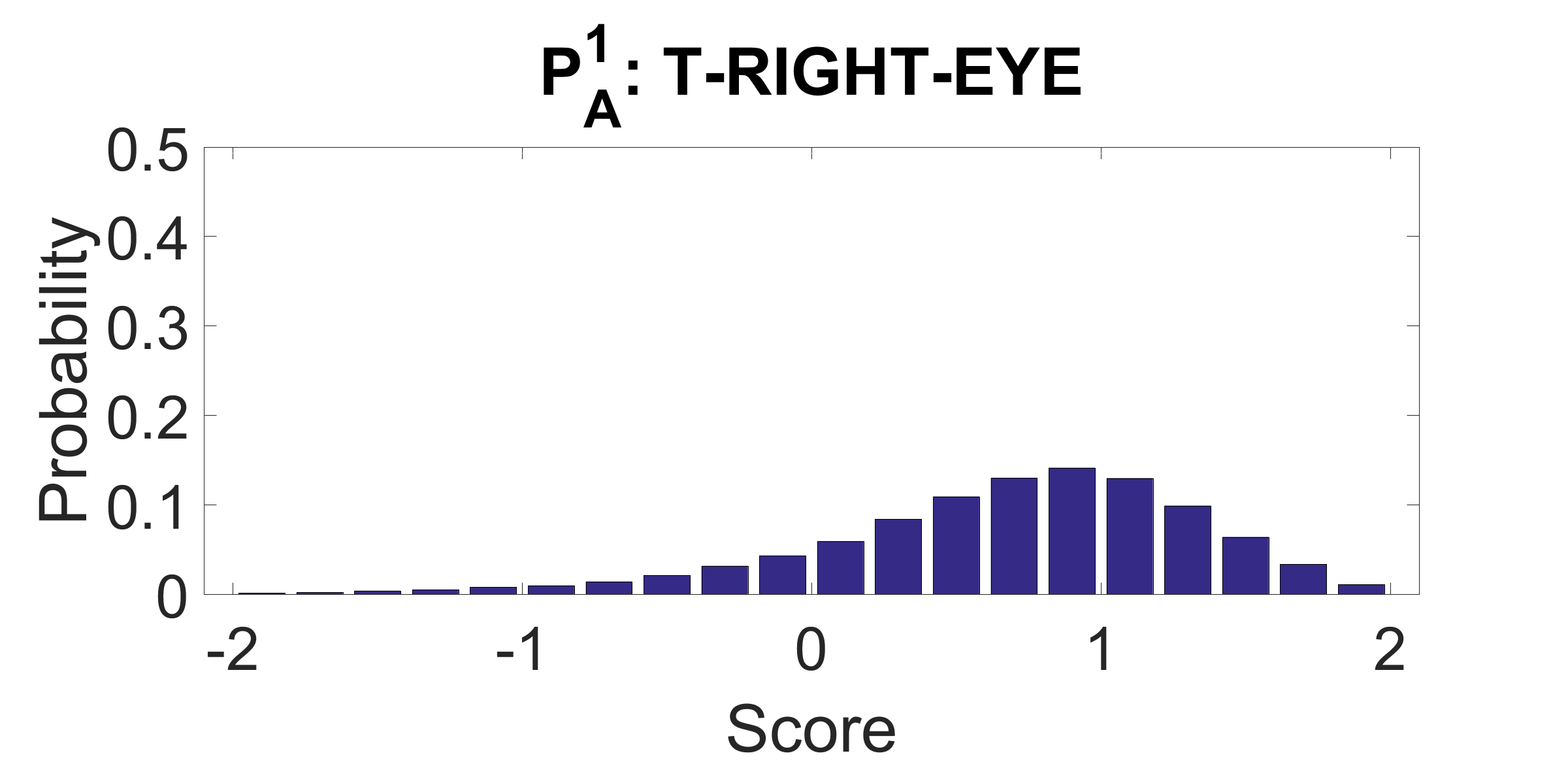}
  \includegraphics[width=0.45\textwidth]{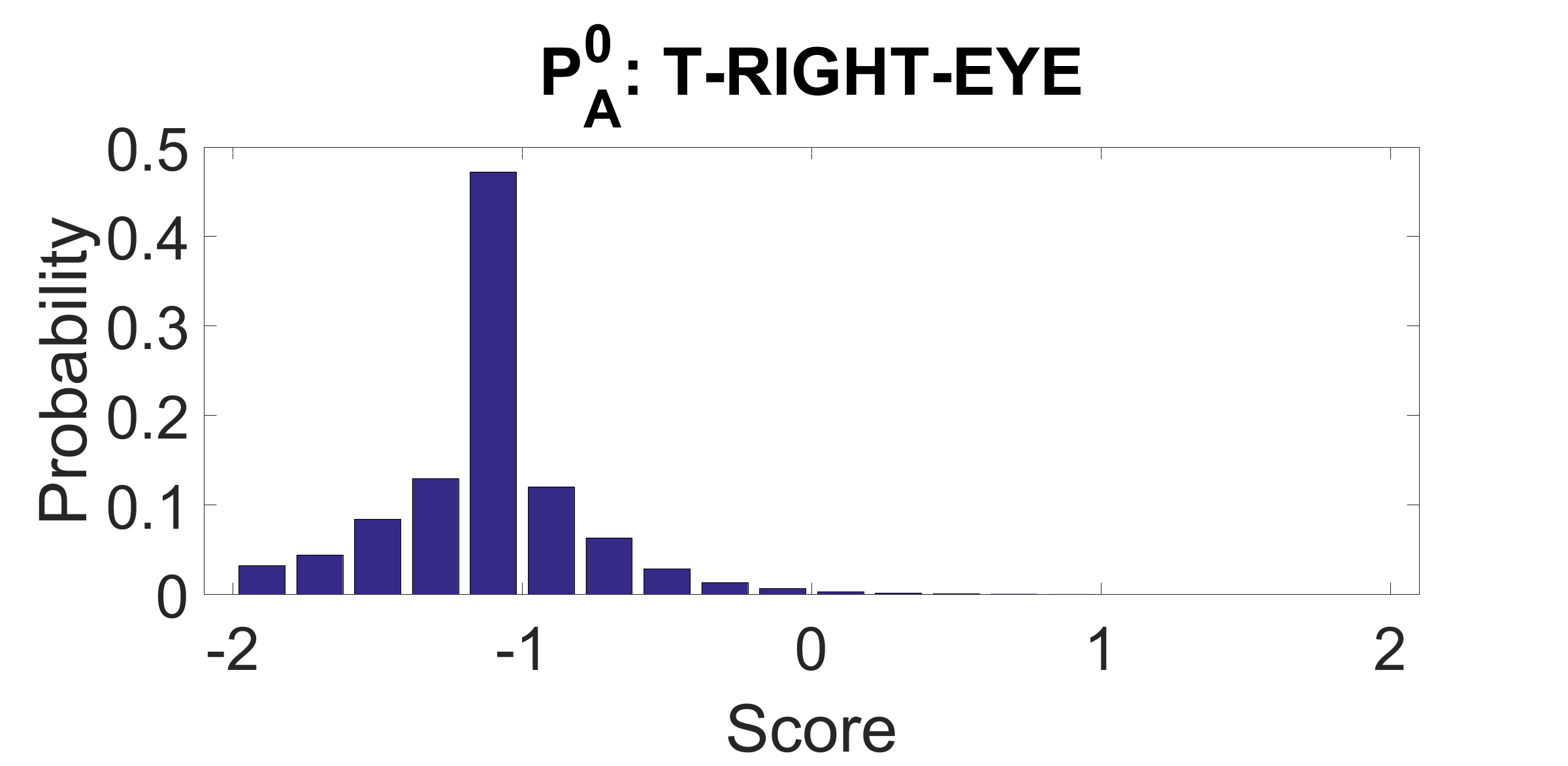}
  \includegraphics[width=0.45\textwidth]{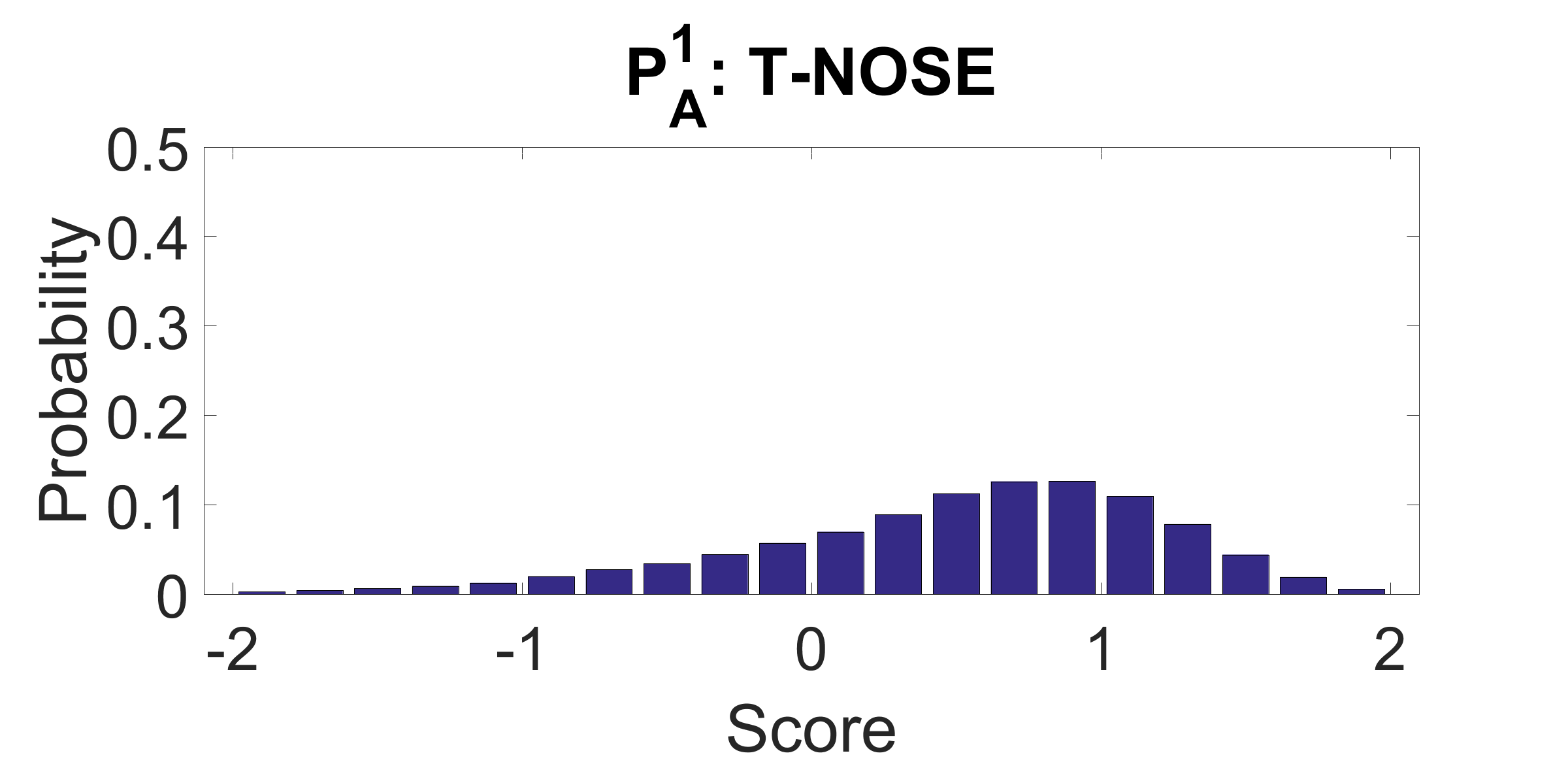}
  \includegraphics[width=0.45\textwidth]{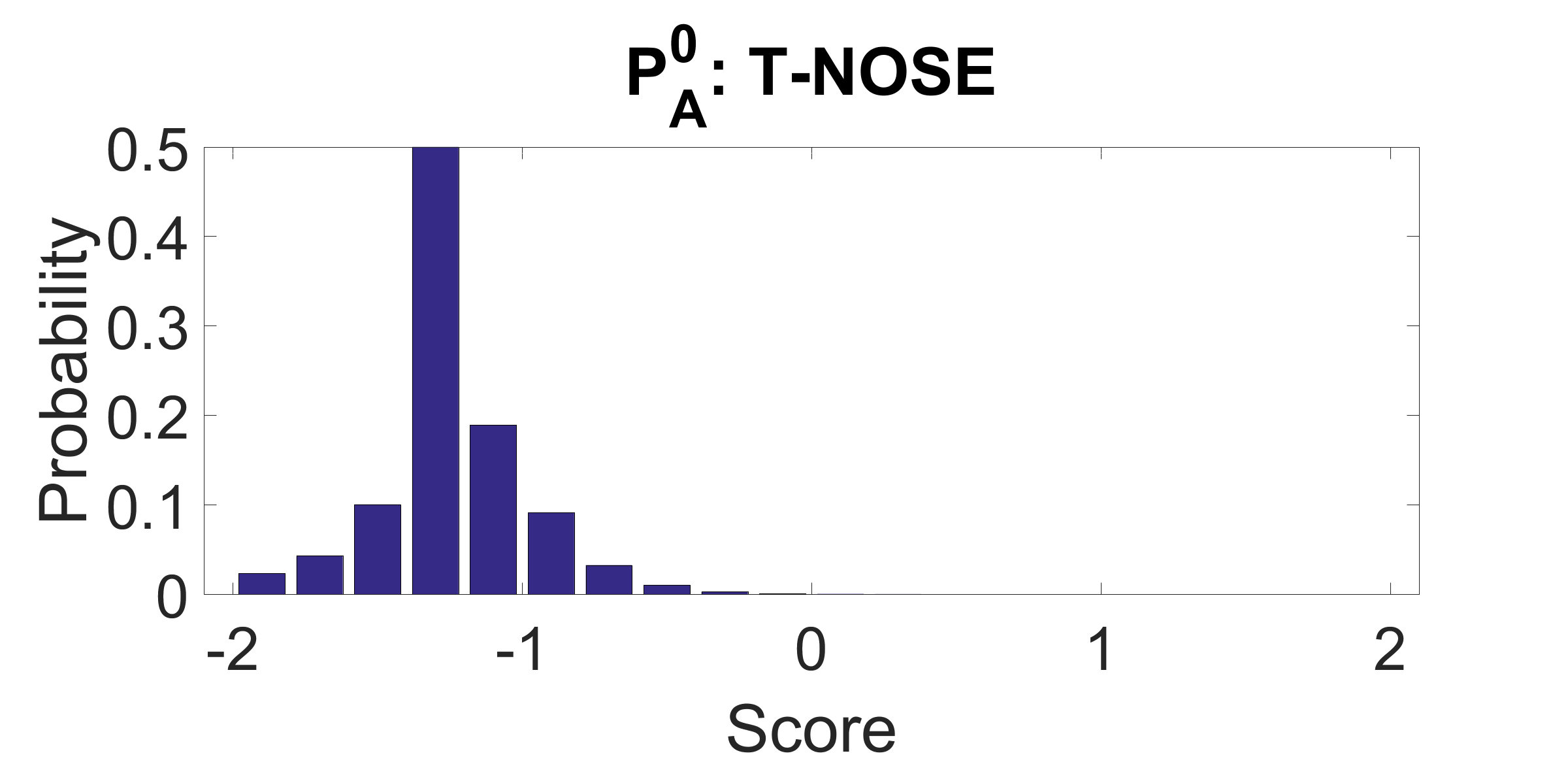}
  \includegraphics[width=0.45\textwidth]{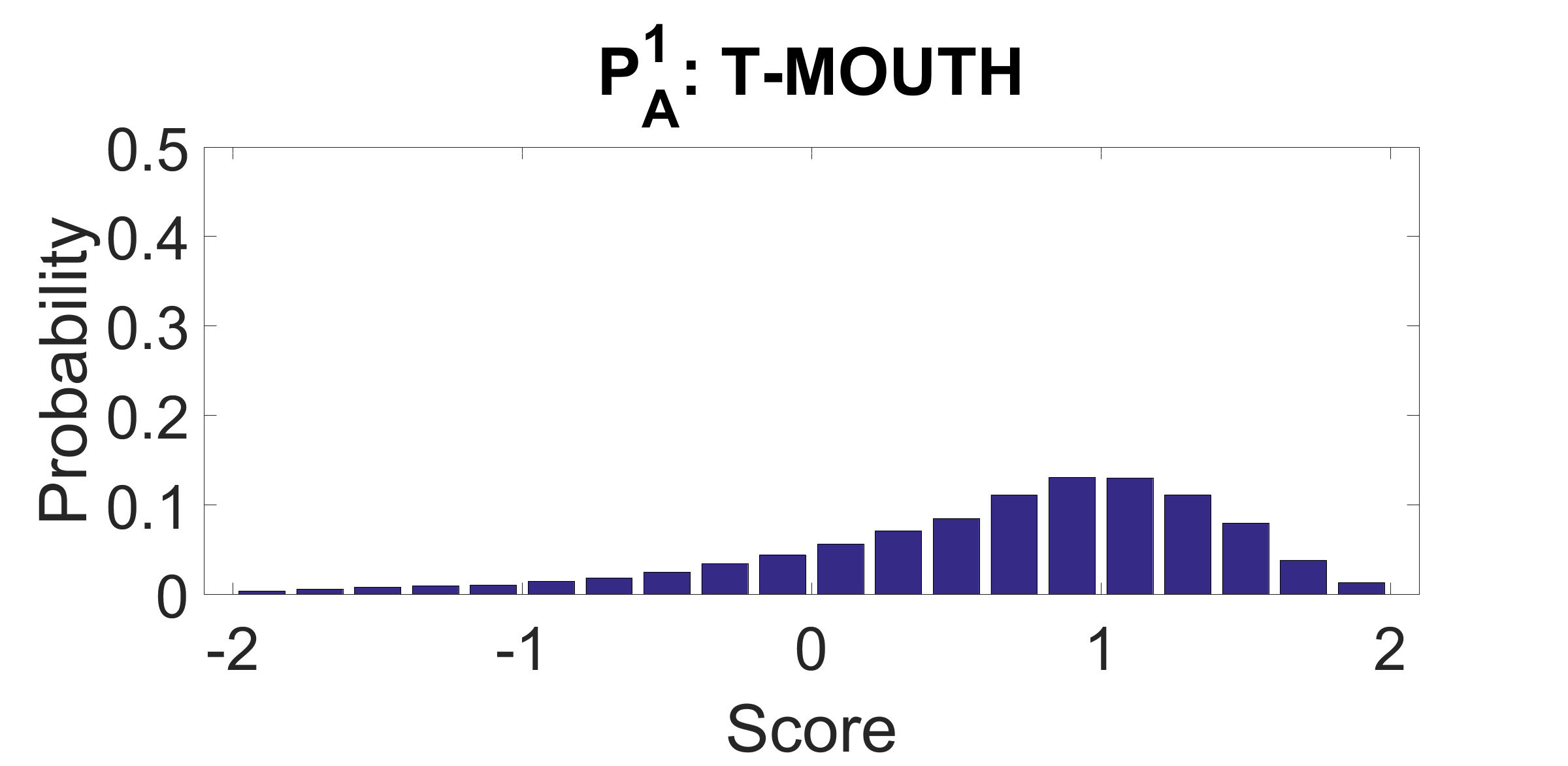}
  \includegraphics[width=0.45\textwidth]{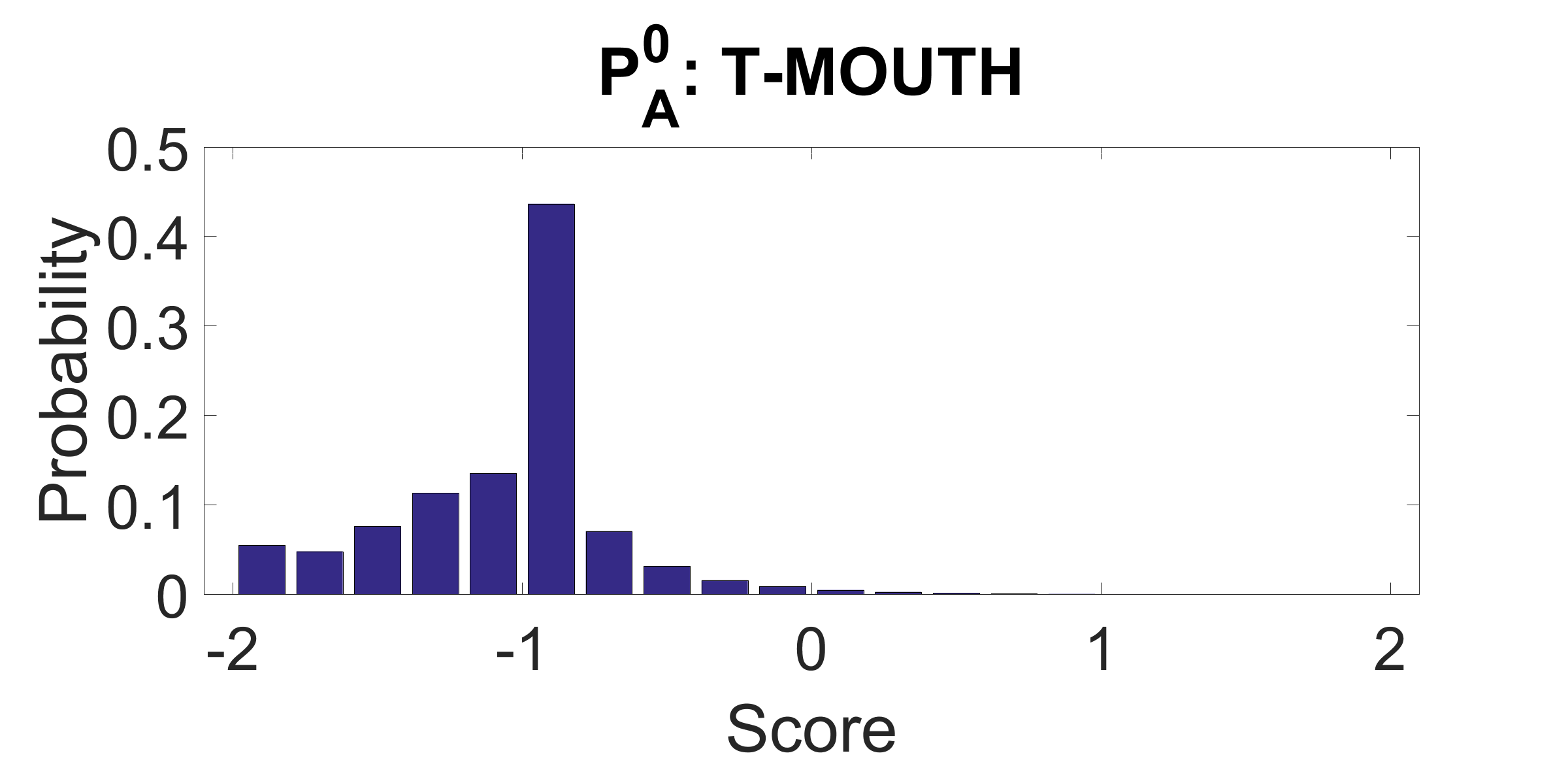}	
  \caption{Distributions over template responses in the face data
    model.  The first column shows the distribution of responses
    conditional on the presence of an object.  The second column shows
    the distribution of responses in the background.}
  \label{fig:hist_scores}
\end{figure}

To estimate $p_{A,0}$ and $p_{A,1}$ we compute histograms of
the template responses in positive and negative training
examples of each object type.  Figure \ref{fig:hist_scores} shows the
estimated distributions of template responses for the various template
symbols in the grammar.  

\subsubsection{Inference}

To detect objects in an image $I$ we consider the posterior $p(S \,|\,
H)$.  The probability that there is an object of type $A$ in pose
$\omega$ is given by $p(X(A,\omega)=1 \,|\, H)$.

By Bayes' rule, $p(S \,|\, H) \propto p(S) p(H \,|\,S)$ and we
incorporate $p(H \,|\,S)$ into the factor graph
representation of $p(S)$ for posterior inference with LBP.

The factor graph $\mathcal{F}({\mathcal G})$ has a binary variable
$X(A,\omega)$ for each $A \in \Sigma_T$ and $\omega \in \Omega_A$.  We
attach unary factors to these these variables with potential $\Psi(x)
= p_{A,x}(H_A(\omega,I))$.  With these additional factors the graphical
model represents the conditional distribution $p(S \,|\,H)$.

\subsubsection{Face Detection in the LFW Dataset}
\label{sec:face_exp_lfw}

In the LFW dataset there is a single face in each image.  To estimate
the locations and sizes of the different objects (face and face parts)
we simply select for each $A \in \{\, \face, \righteye, \lefteye,
\nose, \mouth \,\}$ the pose
\begin{equation*} 
  \omega_A^* = \argmax_{\omega \in \Omega_A} \hat{p}(X(A,\omega)=1 \,|\, H),
  \label{eqn:face-localization-lfw}
\end{equation*}
where $\hat{p}$ is the belief computed by LBP.

For comparison we also evaluate the results of using the templates
to localize objects of each type independently.  We refer to this
method as the No-Context approach.  In this case for
each $A \in \{\, \facelike, \lefteyelike, \righteyelike,
\noselike, \mouthlike \,\}$ we select the pose
\begin{equation*}
  \omega_A^* = \argmax_{\omega \in \Omega_A} \frac{p_{A,1}(H_A(\omega,I))}{p_{A,0}(H_A(\omega,I))}.
\end{equation*}

\begin{figure}
  \centering
  \begin{tabular}{cccc}
    \includegraphics[width=0.22\textwidth]{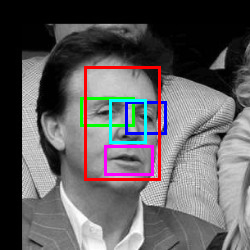} &	
    \includegraphics[width=0.22\textwidth]{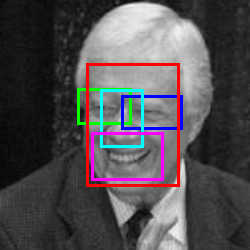} &	
    \includegraphics[width=0.22\textwidth]{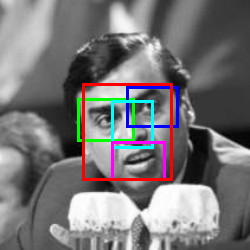} &	
    \includegraphics[width=0.22\textwidth]{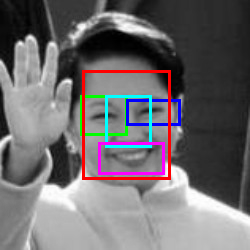} \\ 
    \multicolumn{4}{c}{Ground Truth} \\ \\
    \includegraphics[width=0.22\textwidth]{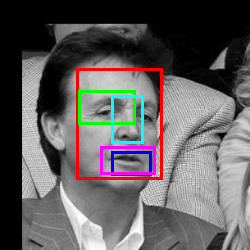} &	
    \includegraphics[width=0.22\textwidth]{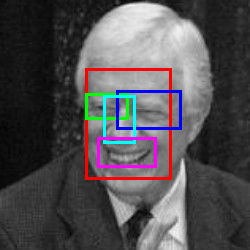} &
    \includegraphics[width=0.22\textwidth]{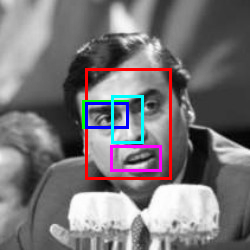} &
    \includegraphics[width=0.22\textwidth]{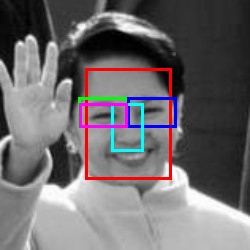} \\ 
    \multicolumn{4}{c}{No-Context} \\ \\
    \includegraphics[width=0.22\textwidth]{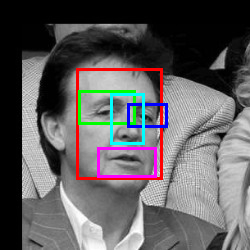} &	
    \includegraphics[width=0.22\textwidth]{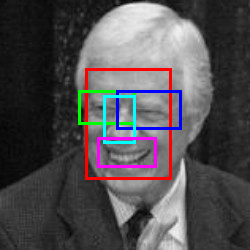} &
    \includegraphics[width=0.22\textwidth]{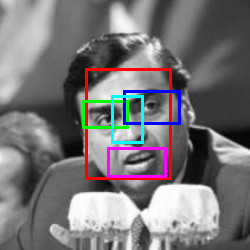} &	
    \includegraphics[width=0.22\textwidth]{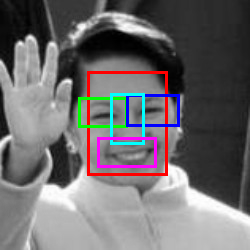} \\
    \multicolumn{4}{c}{PSG face model}
  \end{tabular}
  \caption{Localization results in the LFW dataset.  The
    objects are $\face$ (red), $\lefteye$ (green), $\righteye$ (blue),
    $\nose$ (cyan), and $\mouth$ (magenta).  The No-Context approach
    makes several mistakes due to the visual similarity between
    different parts.  The PSG face model accurately localizes all
    objects in these examples. }
    \label{fig:exp_detect_lfw}
\end{figure}

Figure~\ref{fig:exp_detect_lfw} shows several results using both
methods.  Inference with the PSG face model correctly localized all
parts in these images.  On the other hand, inference with the
No-Context approach leads to mistakes in three of the four images
shown.  These mistakes can be attributed to the visual similarity of
the face parts, at least when represented using HOG features.  We see
that the context provided by the PSG face model resolves ambiguities and 
improves object localization.

Inference with the PSG face model took 120 seconds on a $250 \times
250$ image.  Inference with the No-Context model simply involves
evaluating the template responses at every position and scale within
the image, and took approximately 5 seconds per image.

Table~\ref{table:exp_lfw_dist} summarizes the localization accuracy
obtained using the PSG face model and the No-Context approach in the
LFW dataset.  For comparison Table~\ref{table:exp_lfw_dist} also
summarizes the results obtained using a pictorial structure model
described in Section~\ref{sec:pstruct}.

\begin{table}
  \centering
\begin{tabular}{|c|c|c|c|c|c|c|}
\hline
\textbf{Model} & $\face$ & $\lefteye$ & $\righteye$ & $\nose$ & $\mouth$ & \textbf{Average}\\ \hline
No-Context model & 3.7 & 4.7 & 8.2 & 3.3 & 13.6 & 6.7 \\ \hline
PSG face model & 3.5 & 2.6 & 3.3 & 2.4 & 3.5 & 3.1 \\ \hline 
Pictorial Structure & 3.3 & 2.6 & 3.1 & 2.4 & 3.4 & 3.0 \\ \hline
\end{tabular}
\vspace{.2cm}
\caption{Mean distance between predicted and ground truth object
  location on the LFW dataset.  The locations are the centers of
  bounding boxes and distances were measured in pixels.}
\label{table:exp_lfw_dist}
\end{table}
  
\subsubsection{Face Detection on the Portraits Dataset}

\begin{figure}
  \centering
  \begin{tabular}{cc}
    \includegraphics[height=4cm]{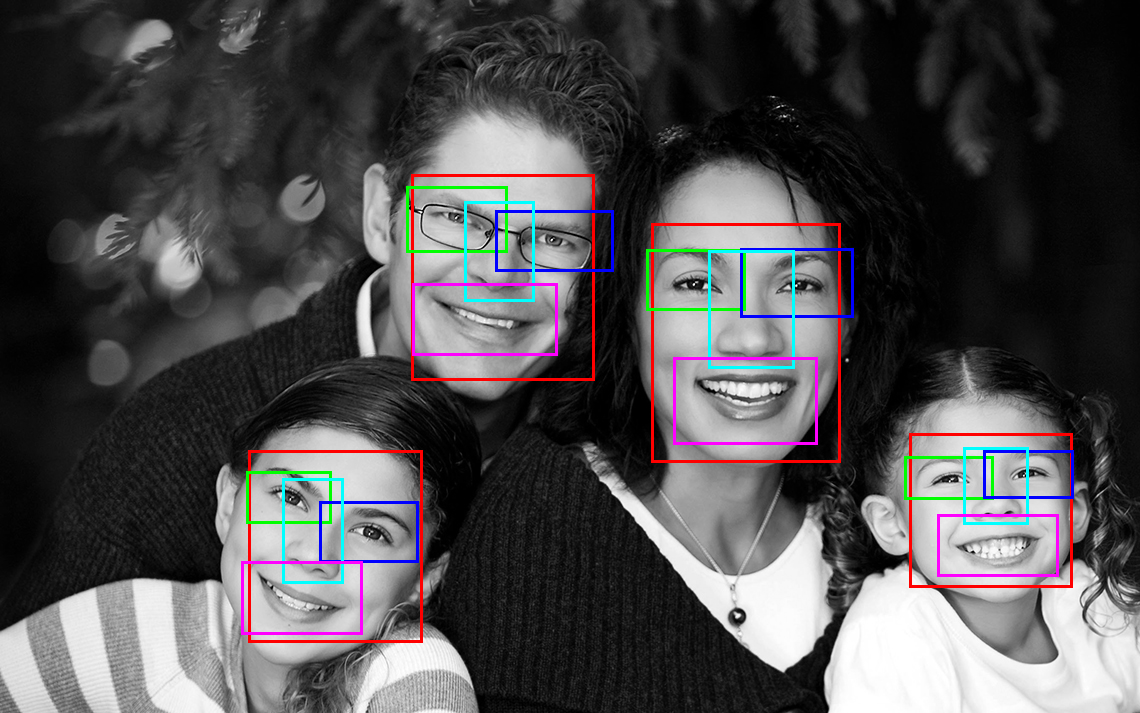} &
    \includegraphics[height=4cm]{results/hogFace/detections/portraits2/groundTruth/ex22.png} \\
    \multicolumn{2}{c}{Ground Truth} \\ \\
    \includegraphics[height=4cm]{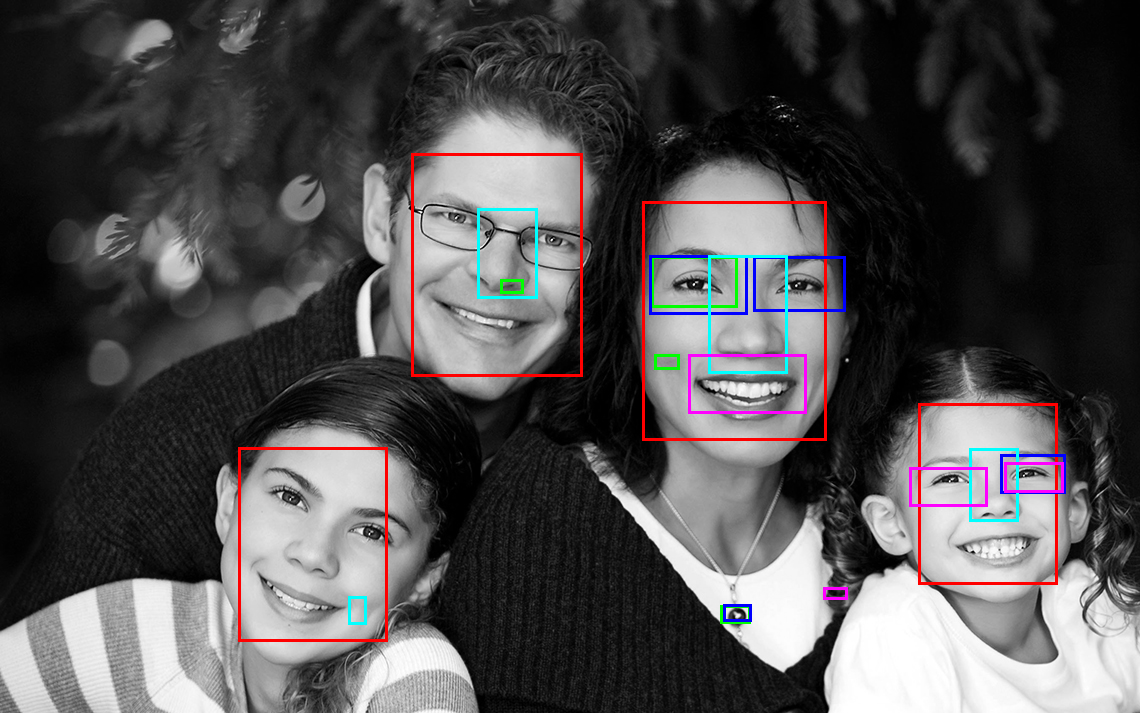} &
    \includegraphics[height=4cm]{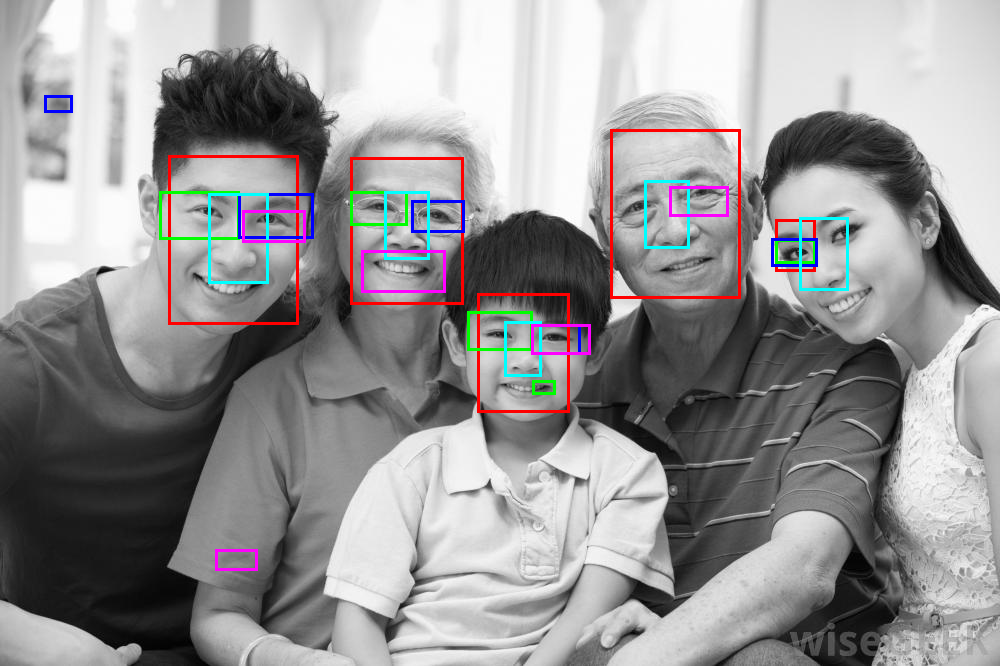} \\
    \multicolumn{2}{c}{No-Context} \\ \\
    \includegraphics[height=4cm]{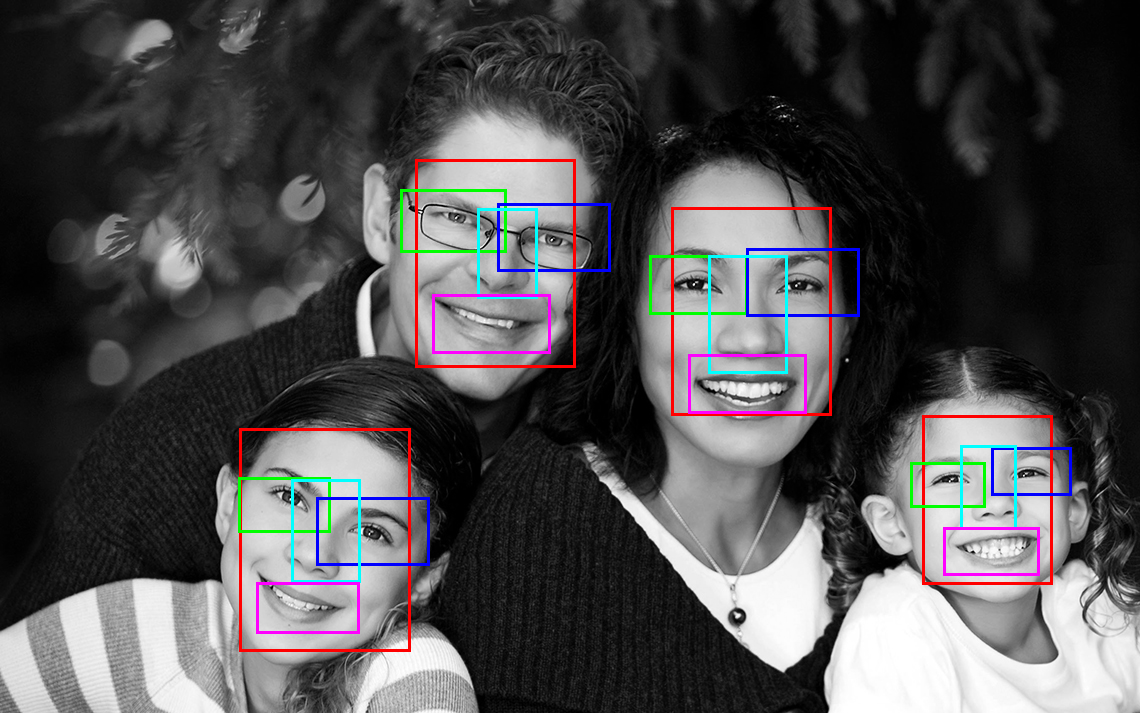} &
    \includegraphics[height=4cm]{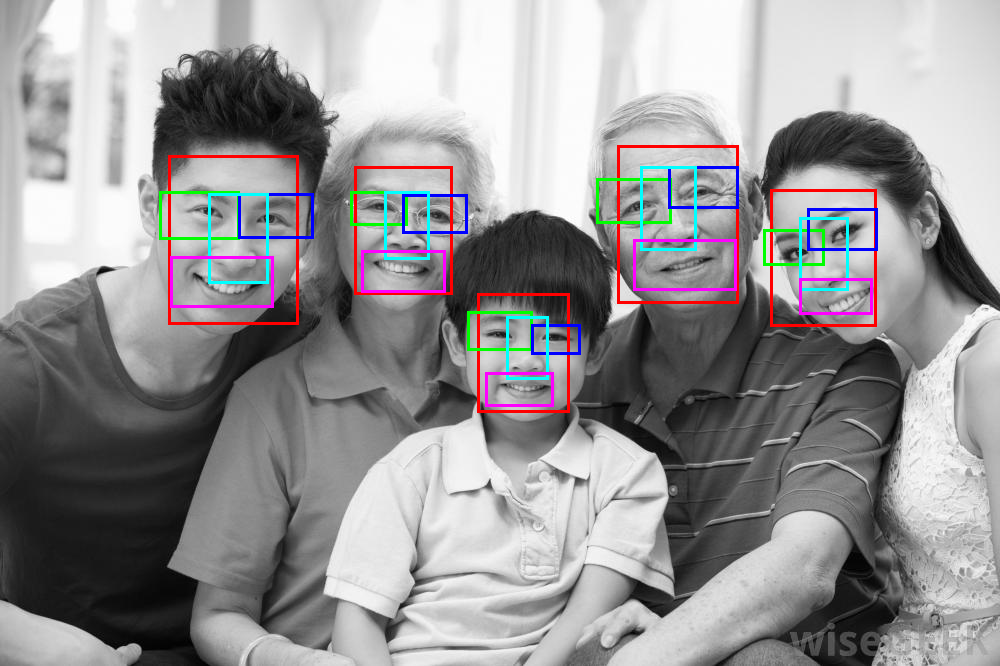} \\
    \multicolumn{2}{c}{PSG face model}
  \end{tabular}
  \caption{Top $K$ detections of each object type on two images from the
    Portraits dataset. The objects are $\face$ (red), $\lefteye$ (green),
    $\righteye$ (blue), $\nose$ (cyan), and $\mouth$ (magenta).}
  \label{fig:exp_detect_portraits_1}
\end{figure}

\begin{figure}
  \centering
  \begin{tabular}{cc}
    \includegraphics[height=4cm]{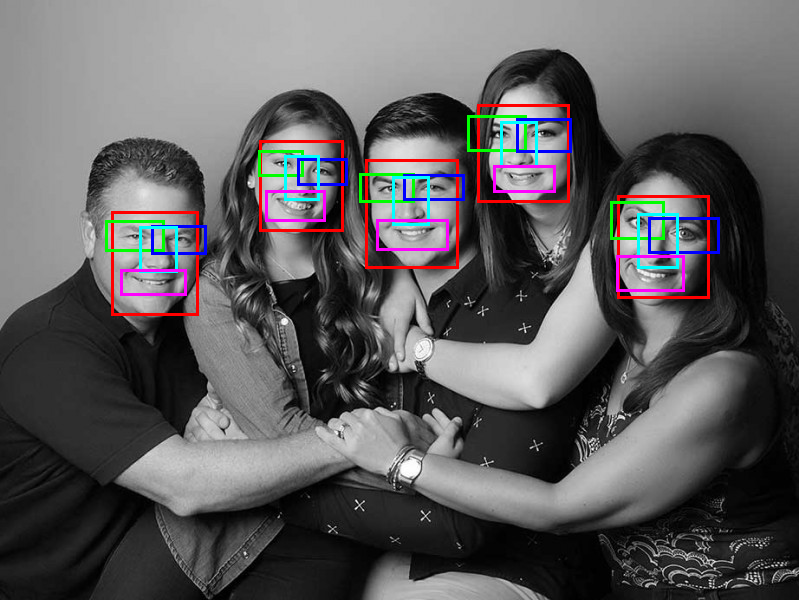} &
    \includegraphics[height=4cm]{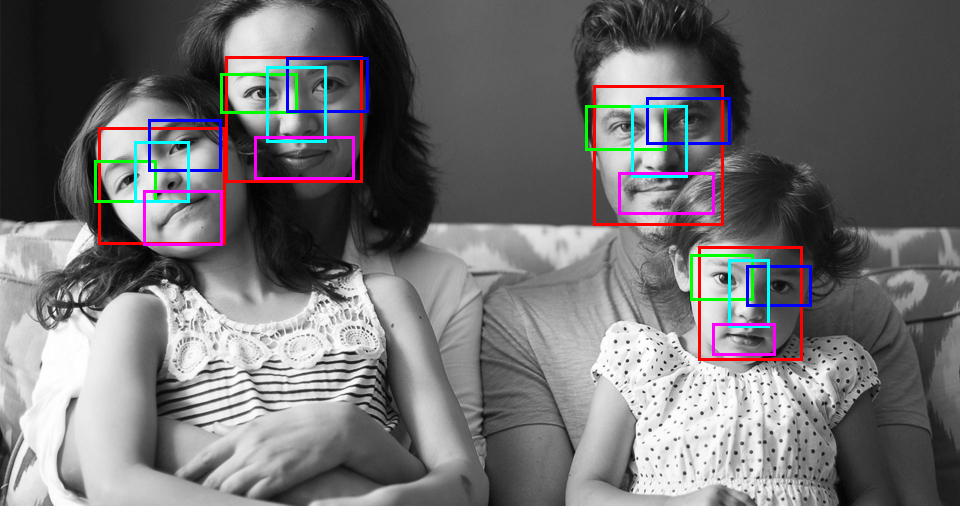} \\ 
    \multicolumn{2}{c}{Ground Truth} \\ \\
    \includegraphics[height=4cm]{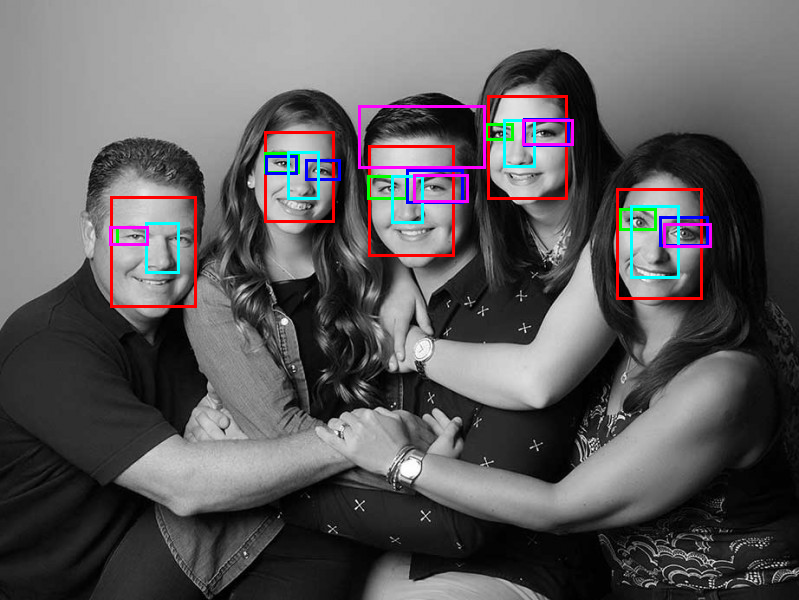} &
    \includegraphics[height=4cm]{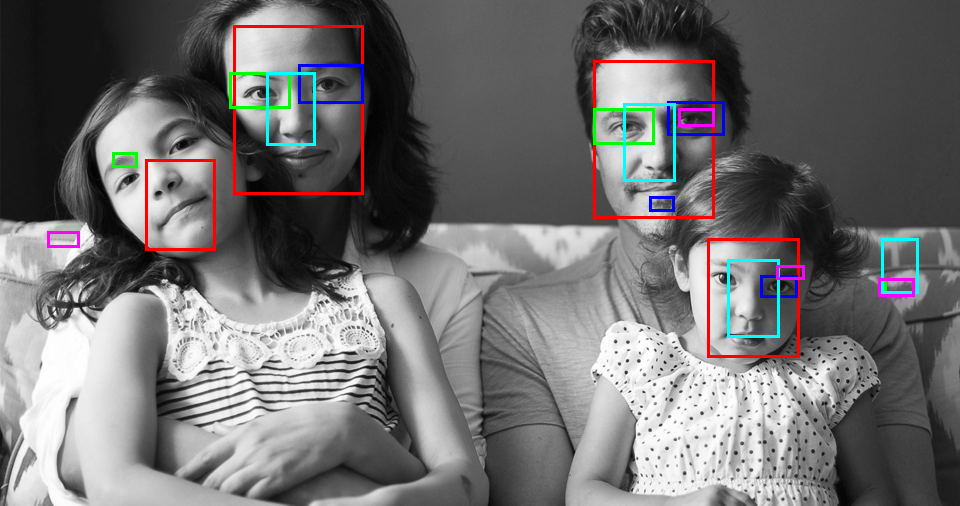} \\ 
    \multicolumn{2}{c}{No-Context} \\ \\
    \includegraphics[height=4cm]{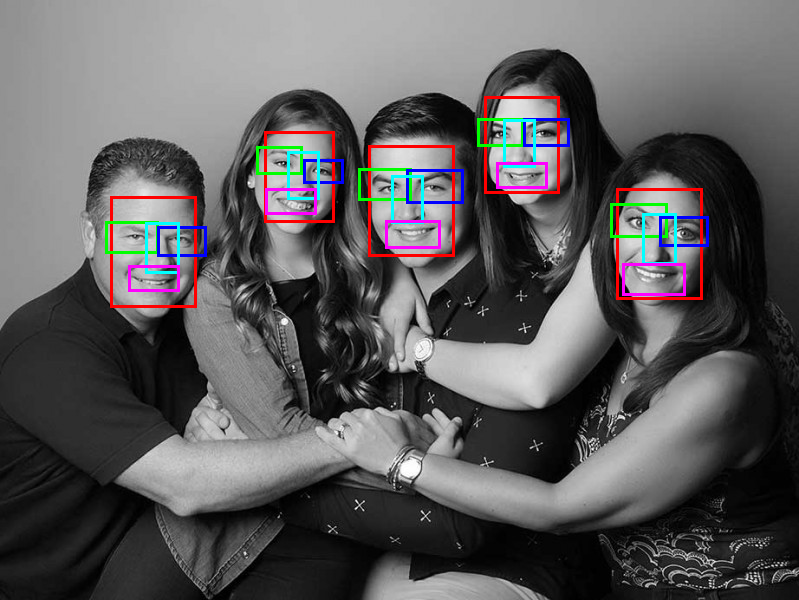} &
    \includegraphics[height=4cm]{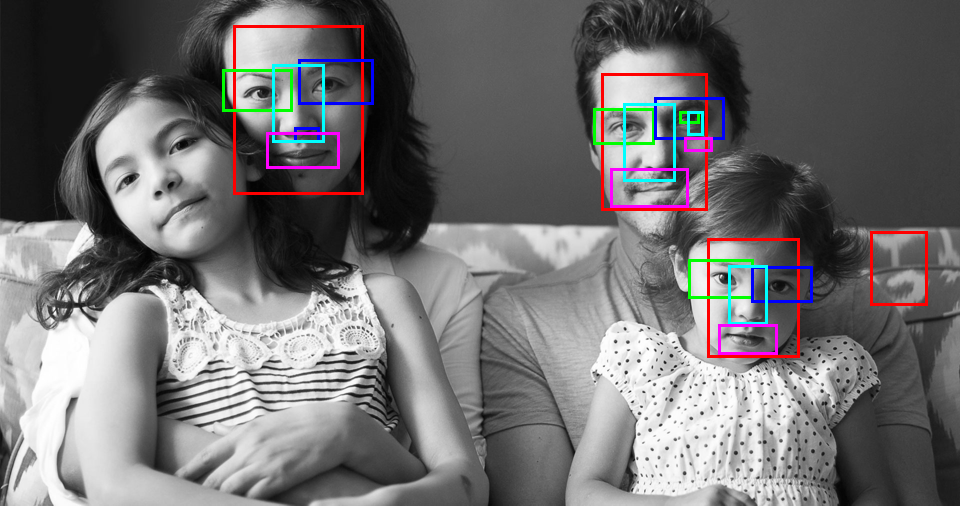} \\
    \multicolumn{2}{c}{PSG face model}
  \end{tabular}
  \caption{Top $K$ detections of each object type on two images from the
    Portraits dataset. The objects are $\face$ (red), $\lefteye$ (green),
    $\righteye$ (blue), $\nose$ (cyan), and $\mouth$ (magenta).  The
    example on the right shows a difficult case where a face is
    significantly rotated, and the pattern on the couch resembles a
    face.}
  \label{fig:exp_detect_portraits_2}
\end{figure}

To study face detection in images with multiple faces we use the
Portraits dataset described above.  Figures
\ref{fig:exp_detect_portraits_1} and \ref{fig:exp_detect_portraits_2}
show several results obtained using the PSG face model and the
No-Context approach.  In each image we show the top $K$ poses for
each object type, after performing non-maximum suppression, where $K$
is the number of faces in the image.  As in the LFW dataset we see
several mistakes in the No-Context results, while the results obtained
with the PSG face model are almost perfect.  

For a quantitative evaluation of the PSG face model when the number of
faces in the image is unknown we generate a set of detections for
each symbol by thresholding the estimated conditional probabilities
$\hat{p}(X(A,\omega)=1 \,|\, H)$ after non-maximum suppression.

For the No-Context model we generate detections by performing
non-maximum suppression and thresholding the likelihood ratios
$p_{A,1}(H_A(\omega,I))/p_{A,0}(H_A(\omega,I))$.

By considering different detection thresholds we obtain a
precision-recall curve for each object type with each
method.\footnote{We use the PASCAL VOC criteria to evaluate
  the results.  A detection is considered correct if it has an
  intersection-over-union ratio of at least 0.5 with a ground-truth
  bounding box, otherwise the detection is considered a false
  positive.  If multiple detections overlap with a single ground-truth
  bounding box, only one is considered correct, and the others are
  considered false positives.}  Table~\ref{table:exp_portraits_auc}
summarizes the area under the precision-recall curves.

For comparison, Table~\ref{table:exp_portraits_auc} also summarizes
the results obtained using a pictorial structure model described below
and a PSG model with no face template.  The performance of the PSG
model without a face template is still quite good, including for
detecting the face itself.  This demonstrates how we can detect
objects defined solely by their relationship to other objects.

\begin{table}
  \centering
\begin{tabular}{|c|c|c|c|c|c|c|}
\hline
\textbf{Model} & $\face$ & $\lefteye$ & $\righteye$ & $\nose$ & $\mouth$ & \textbf{Average} \\ \hline
No-Context & 0.95 & 0.50 & 0.48 & 0.90 & 0.32 & 0.63 \\ \hline
PSG face model & 0.97 & 0.81 & 0.81 & 0.96 & 0.80& 0.87 \\ \hline 
PSG with no face template & 0.93 & 0.78 & 0.80 & 0.95 & 0.76 & 0.84 \\ \hline
Pictorial Structure & 0.97 & 0.78 & 0.69 & 0.96 & 0.73 & 0.82 \\ \hline
\end{tabular}
\vspace{.2cm}
\caption{Area under the precision-recall curves in the Portraits dataset.}
\label{table:exp_portraits_auc}
\end{table}

\subsubsection{Comparison to a Pictorial Structure model}
\label{sec:pstruct}

The PSG face model is closely related to a pictorial structure model
(see \cite{FH05}).  We obtain a pictorial structure model if we include
an additional constraint that a scene has exactly one face and one set of
corresponding face parts.  With this constraint the posterior, $p(S
\,|\, H)$, can be represented by a tree-structured graphical model.
The graphical model has one variable for each face part connected to a
variable for the face.  The value of a variable specifies the pose of
the corresponding object.  For this tree-structure model we can
compute exact posterior marginals using dynamic programming, following
the approach in \cite{FH05}.

Tables \ref{table:exp_lfw_dist} and \ref{table:exp_portraits_auc}
summarize results obtained using exact inference with a pictorial
structure model for comparison with the PSG face model.  To detect
multiple objects of a single type in an image we threshold the
probability of each pose after non-maximum suppression.

In the LFW dataset, where there is a single face in each image, the
pictorial structure model and the PSG face model perform similarly.
On the other hand, the PSG face model outperforms the pictorial
structure model when detecting faces in the Portraits dataset.  In
this case we see an advantage to modeling scenes with a variable
number of objects.

\section{Summary}

We described a general framework for knowledge representation and
for scene understanding using probabilistic scene grammars and loopy belief
propagation.

Probabilistic scene grammars can generate complex
structures and capture regularities of scenes in a variety of settings.  
The approach we develop for inference involves the construction of
large graphical models that represent distributions
over regular scenes.  We use loopy belief propagation
for inference and derive techniques for implementing the
sum-product algorithm in this setting.  The resulting method is
efficient, robust, and broadly applicable.

We also considered the
problem of learning model parameters using maximum-likelihood
estimation, both from fully observed and partially observed secenes.

The main contribution of our work is a mathematical and algorithmic
foundation for probabilistic modeling of scenes and for scene
understanding with grammar models.  We demonstrate the 
feasibility of the approach with several examples.  The experimental
results show how the same framework and computational engine 
can be used to address different
problems and lead to practical methods for inference.

The design of sophisticated grammars for different applications and
the development of methods to learn the structure of scene
grammars are important topics for future research.

\subsection*{Acknowledgments}

We thank Stuart Geman, Basilis Gidas, Caroline Klivans and Jackson
Loper for many helpful discussions about this research.  We also thank
the anonymous reviewers for many helpful comments that helped improve
the presentation of this material.  This material is based upon work
supported by the National Science Foundation under Grant No. 1447413.

\bibliography{paper}

\begin{thebibliography}{10}

\bibitem{ASU86}
Alfred~V. Aho, Ravi Sethi, and Jeffrey~D. Ullman.
\newblock {\em Compilers: Principles, Tools, and Techniques}.
\newblock Addison-Wesley, 1986.

\bibitem{A02}
Yali Amit.
\newblock {\em \mbox{2D} Object Detection and Recognition}.
\newblock MIT Press, 2002.

\bibitem{AMFM11}
Pablo Arbelaez, Michael Maire, Charless Fowlkes, and Jitendra Malik.
\newblock Contour detection and hierarchical image segmentation.
\newblock {\em IEEE Transactions on Pattern Analysis and Machine Intelligence},
  33(5):898--916, May 2011.

\bibitem{B86}
Julian Besag.
\newblock On the statistical analysis of dirty pictures.
\newblock {\em Journal of the Royal Statistical Society. Series B
  (Methodological)}, pages 259--302, 1986.

\bibitem{BGP97}
Elie Bienenstock, Stuart Geman, and Daniel Potter.
\newblock Compositionality, \mbox{MDL} priors, and object recognition.
\newblock In {\em Advances in Neural Information Processing Systems}, pages
  838--844, 1997.

\bibitem{BWP98}
Michael Burl, Markus Weber, and Pietro Perona.
\newblock A probabilistic approach to object recognition using local photometry
  and global geometry.
\newblock In {\em European Conference on Computer Vision}, pages 628--641,
  1998.

\bibitem{CJZBG11}
Lo-Bin Chang, Ya~Jin, Wei Zhang, Eran Borenstein, and Stuart Geman.
\newblock Context, computation, and optimal \mbox{ROC} performance in
  hierarchical models.
\newblock {\em International Journal of Computer Vision}, 93(2):117--140, 2011.

\bibitem{CJ93}
Rama Chellapa and Anil Jain.
\newblock {\em Markov Random Fields: Theory and Application}.
\newblock Academic Press, 1993.

\bibitem{C56}
Noam Chomsky.
\newblock Three models for the description of language.
\newblock {\em IRE Transactions on information theory}, 2(3):113--124, 1956.

\bibitem{CLRS}
Thomas Cormen, Charles Leiserson, Ronald Rivest, and Clifford Stein.
\newblock {\em Introduction to algorithms second edition}.
\newblock The MIT Press, 2001.

\bibitem{DT05}
Navneet Dalal and Bill Triggs.
\newblock Histograms of oriented gradients for human detection.
\newblock In {\em IEEE Conference on Computer Vision and Pattern Recognition},
  pages 886--893, 2005.

\bibitem{DLR77}
A.~P. Dempster, N.~M. Laird, and D.~B. Rubin.
\newblock Maximum likelihood from incomplete data via the em algorithm.
\newblock {\em Journal of the Royal Statistical Society, Series B},
  39(1):1--38, 1977.

\bibitem{D06}
Frank Drewes.
\newblock {\em Grammatical Picture Generation}.
\newblock Springer, 2006.

\bibitem{D98}
Richard Durbin, Sean~R. Eddy, Anders Krogh, and Graeme Mitchison.
\newblock {\em Biological sequence analysis: probabilistic models of proteins
  and nucleic acids}.
\newblock Cambridge university press, 1998.

\bibitem{pascal-voc-2012}
M.~Everingham, L.~Van~Gool, C.~K.~I. Williams, J.~Winn, and A.~Zisserman.
\newblock The {PASCAL} {V}isual {O}bject {C}lasses {C}hallenge 2012 {(VOC2012)}
  {R}esults, 2012.

\bibitem{voc-release4}
Pedro~F. Felzenszwalb, Ross~B. Girshick, and David McAllester.
\newblock Discriminatively trained deformable part models, release 4, 2010.

\bibitem{FGMR10}
Pedro~F. Felzenszwalb, Ross~B. Girshick, David McAllester, and Deva Ramanan.
\newblock Object detection with discriminatively trained part-based models.
\newblock {\em IEEE transactions on pattern analysis and machine intelligence},
  32(9):1627--1645, 2010.

\bibitem{FH05}
Pedro~F. Felzenszwalb and Daniel~P. Huttenlocher.
\newblock Pictorial structures for object recognition.
\newblock {\em International Journal of Computer Vision}, 61(1):55--79, 2005.

\bibitem{FM10}
Pedro~F. Felzenszwalb and David McAllester.
\newblock Object detection grammars.
\newblock {\em Univerity of Chicago Computer Science Technical Report 2010-02},
  2010.

\bibitem{FOP}
Pedro~F. Felzenszwalb and John~G. Oberlin.
\newblock Multiscale fields of patterns.
\newblock In {\em Advances in Neural Information Processing Systems}, pages
  82--90, 2014.

\bibitem{FE73}
Martin~A. Fischler and Robert~A. Elschlager.
\newblock The representation and matching of pictorial structures.
\newblock {\em IEEE Transactions on computers}, (1):67--92, 1973.

\bibitem{Fu74}
King~Sun Fu.
\newblock {\em Syntactic methods in pattern recognition}.
\newblock Elsevier, 1974.

\bibitem{GJ96}
Donald Geman and Bruno Jedynak.
\newblock An active testing model for tracking roads in satellite images.
\newblock {\em IEEE Transactions on Pattern Analysis and Machine Intelligence},
  18(1):1--14, 1996.

\bibitem{GG84}
Stuart Geman and Donald Geman.
\newblock Stochastic relaxation, gibbs distributions, and the bayesian
  restoration of images.
\newblock {\em IEEE Transactions on Pattern Analysis and Machine Intelligence},
  6:721--741, 1984.

\bibitem{GPZ02}
Stuart Geman, Daniel~F. Potter, and Zhiyi Chi.
\newblock Composition systems.
\newblock {\em Quarterly of Applied Mathematics}, 60(4):707--736, 2002.

\bibitem{GPT}
Ulf Grenander.
\newblock {\em General Pattern Theory}.
\newblock Oxford University Press, 1993.

\bibitem{H05}
Matthew~T. Harrison.
\newblock {\em Discovering compositional structures}.
\newblock PhD thesis, Brown University, 2005.

\bibitem{HZW04}
Tom Heskes, Onno Zoeter, and Wim Wiegerinck.
\newblock Approximate expectation maximization.
\newblock In {\em Advances in Neural Information Processing Systems 16}, pages
  353--360, 2004.

\bibitem{LFW}
Gary~B. Huang, Manu Ramesh, Tamara Berg, and Erik Learned-Miller.
\newblock Labeled faces in the wild: A database for studying face recognition
  in unconstrained environments.
\newblock Technical Report 07-49, University of Massachusetts, Amherst, October
  2007.

\bibitem{JG06}
Ya~Jin and Stuart Geman.
\newblock Context and hierarchy in a probabilistic image model.
\newblock In {\em IEEE Conference on Computer Vision and Pattern Recognition},
  volume~2, pages 2145--2152, 2006.

\bibitem{K05}
Dan Klein.
\newblock {\em The Unsupervised Learning of Natural Language Structure}.
\newblock PhD thesis, Stanford University, 2005.

\bibitem{KFL01}
Frank~R. Kschischang, Brendan~J. Frey, and Hans-Andrea Loeliger.
\newblock Factor graphs and the sum-product algorithm.
\newblock {\em IEEE Transactions on Information Theory}, 47(2):498--519, 2001.

\bibitem{Picture}
Tejas~D Kulkarni, Pushmeet Kohli, Joshua~B Tenenbaum, and Vikash Mansinghka.
\newblock Picture: A probabilistic programming language for scene perception.
\newblock In {\em IEEE Conference on Computer Vision and Pattern Recognition},
  pages 4390--4399, 2015.

\bibitem{Manning99}
Christopher~D. Manning and Hinrich Sch{\"u}tze.
\newblock {\em Foundations of statistical natural language processing}.
\newblock MIT Press, 1999.

\bibitem{M94b}
David Mumford.
\newblock The bayesian rationale for energy functionals.
\newblock {\em Geometry-driven diffusion in Computer Vision}, pages 141--153,
  1994.

\bibitem{M94}
David Mumford.
\newblock Elastica and computer vision.
\newblock In {\em Algebraic geometry and its applications}, pages 491--506.
  Springer, 1994.

\bibitem{MWJ99}
Kevin~P. Murphy, Yair Weiss, and Michael~I. Jordan.
\newblock Loopy belief propagation for approximate inference: An empirical
  study.
\newblock In {\em Uncertainty in Artificial Intelligence}, pages 467--475,
  1999.

\bibitem{Palmer99}
Stephen~E. Palmer.
\newblock {\em Vision science: Photons to phenomenology}.
\newblock MIT press, 1999.

\bibitem{Pearl88}
Judea Pearl.
\newblock {\em Probabilistic reasoning in intelligent systems: networks of
  plausible inference}.
\newblock Morgan Kaufmann, 1988.

\bibitem{PL91}
Przemyslaw Prusinkiewicz and Aristid Lindenmayer.
\newblock {\em The algorithmic beauty of plants (The Virtual Laboratory)}.
\newblock Springer, 1991.

\bibitem{R79}
Azriel Rosenfeld.
\newblock {\em Picture Languages (Formal Models for Picture Recognition)}.
\newblock Academic Press, 1979.

\bibitem{SU88}
A.~Shashua and S.~Ullman.
\newblock Structural saliency: The detection of globally salient structures
  using a locally connected network.
\newblock {\em MIT AI Lab Memo No. 1061}, 1988.

\bibitem{S94}
Andreas Stolcke.
\newblock {\em Bayesian Learning of Probabilistic Language Models}.
\newblock PhD thesis, University of California at Berkeley, 1994.

\bibitem{Tarlow12}
Daniel Tarlow, Kevin Swersky, Richard~S. Zemel, Ryan~Prescott Adams, and
  Brendan~J. Frey.
\newblock Fast exact inference for recursive cardinality models.
\newblock In {\em Uncertainty in Artificial Intelligence}, 2012.

\bibitem{Edward}
Dustin Tran, Matthew~D. Hoffman, Rif~A. Saurous, Eugene Brevdo, Kevin Murphy,
  and David~M. Blei.
\newblock Deep probabilistic programming.
\newblock In {\em International Conference on Learning Representations}, 2017.

\bibitem{TCYZ05}
Zhuowen Tu, Xiangrong Chen, Alan~L. Yuille, and Song-Chun Zhu.
\newblock Image parsing: Unifying segmentation, detection, and recognition.
\newblock {\em International Journal of Computer Vision}, 63(2):113--140, 2005.

\bibitem{WJ08}
Martin~J. Wainwright and Michael~I. Jordan.
\newblock Graphical models, exponential families, and variational inference.
\newblock {\em Foundations and Trends in Machine Learning}, 1(1–2):1--305,
  2008.

\bibitem{W00}
Yair Weiss.
\newblock Correctness of local probability propagation in graphical models with
  loops.
\newblock {\em Neural computation}, 12(1):1--41, 2000.

\bibitem{WJ97}
Lance~R. Williams and David~W. Jacobs.
\newblock Stochastic completion fields: A neural model of illusory contour
  shape and salience.
\newblock {\em Neural computation}, 9(4):837--858, 1997.

\bibitem{YFW01}
Jonathan~S. Yedidia, William~T. Freeman, and Yair Weiss.
\newblock Understanding belief propagation and its generalizations.
\newblock In {\em Exploring artificial intelligence in the new millennium},
  pages 236--239. Morgan Kaufmann, 2001.

\bibitem{ZZ11}
Yibiao Zhao and Song-Chun Zhu.
\newblock Image parsing with stochastic scene grammar.
\newblock In {\em Advances in Neural Information Processing Systems}, pages
  73--81, 2011.

\bibitem{ZCY09}
Long Zhu, Yuanhao Chen, and Alan Yuille.
\newblock Unsupervised learning of probabilistic grammar-markov models for
  object categories.
\newblock {\em IEEE Transactions on Pattern Analysis and Machine Intelligence},
  31(1):114--128, 2009.

\bibitem{ZM07}
Song-Chun Zhu and David Mumford.
\newblock A stochastic grammar of images.
\newblock {\em Foundations and Trends in Computer Graphics and Vision},
  2(4):259--362, 2007.

\end{thebibliography}
\bibliographystyle{plain}

\appendix
\section{Derivation of message passing formulas}
\label{app:msg}

Below we derive the expressions that are used in the efficient
implementation of loopy belief propagation.  We assume all messages
are non-zero and are normalized to sum to one.

Note that if $\mu_1(x_1),\ldots,\mu_n(x_n)$ are $n$ vectors that sum
to one and $x=(x_1,\ldots,x_n)$, then
$$\sum_x \prod_i \mu_i(x_i) = \prod_i \sum_{x_i} \mu_i(x_i) = 1.$$

\leakyortoz*
\begin{proof}
  By definition,
\begin{eqnarray}
\mu_{f \rightarrow Z}(\tau) & = & \kappa \sum_{y} \Psi^L_\epsilon(y,\tau) \prod_i \mu_{Y_i \rightarrow f}(y_i).
\end{eqnarray}
Consider the case $\tau = 0$ and note that $\Psi^L_\epsilon(y,0)=0$ if $y \neq (0,\ldots,0)$.
\begin{eqnarray}
\mu_{f \rightarrow Z}(0) & = & \kappa \sum_y \Psi^L_\epsilon(y,0) \prod_i \mu_{Y_i \rightarrow f}(y_i), \\
& = & \kappa (1-\epsilon) \prod_i \mu_{Y_i \rightarrow f}(0). 
\end{eqnarray}
Now consider the case $\tau = 1$.
\begin{eqnarray}
  \mu_{f \rightarrow Z}(1) & = & \kappa \sum_y \Psi^L_\epsilon(y,1) \prod_i \mu_{Y_i \rightarrow f}(y_i), \\
  & = & \kappa \sum_y (1-\Psi^L_\epsilon(y,0)) \prod_i \mu_{Y_i \rightarrow f}(y_i), \\
  & = & \kappa \sum_y \prod_i \mu_{Y_i \rightarrow f}(y_i) - \kappa \sum_y \Psi^L_\epsilon(y,0) \prod_i \mu_{Y_i \rightarrow f}(y_i), \\
  & = & \kappa - \mu_{f \rightarrow Z}(0).
\end{eqnarray}
Setting $\kappa=1$ ensures $\sum_{\tau} \mu_{f \rightarrow Z}(\tau) = 1$.
\end{proof}

\leakyortoy*
\begin{proof}
  By definition,
\begin{eqnarray}
\mu_{f \rightarrow Y_i}(\tau) & = & \kappa \sum_{\substack{y\\y_i=\tau}} \sum_z \Psi^L_\epsilon(y,z) \mu_{Z \rightarrow f}(z) \prod_{j \neq i} \mu_{Y_j \rightarrow f}(y_j).
\end{eqnarray}
Consider the case $\tau = 0$.
\begin{eqnarray}
  \mu_{f \rightarrow Y_i}(0)
  & = & \kappa \sum_{\substack{y \\y_i=0,\\c(y)=0}} \sum_z \Psi^L_\epsilon(y,z) \mu_{Z \rightarrow f}(z) \prod_{j \neq i} \mu_{Y_j \rightarrow f}(y_j) \nonumber \\
  & & +\; \kappa \sum_{\substack{y\\y_i=0\\c(y)>0}} \sum_z \Psi^L_\epsilon(y,z) \mu_{Z \rightarrow f}(z) \prod_{j \neq i} \mu_{Y_j \rightarrow f}(y_j), \\
  & = & \kappa ((1-\epsilon) \mu_{Z \rightarrow f}(0)  + \epsilon \mu_{Z \rightarrow f}(1)) \prod_{j \neq i} \mu_{Y_j \rightarrow f}(0) \nonumber \\
  & & +\; \kappa \sum_{\substack{y\\y_i=0\\c(y)>0}} \mu_{Z \rightarrow f}(1) \prod_{j \neq i} \mu_{Y_j \rightarrow f}(y_j), \\
  & = & \kappa ((1-\epsilon) \mu_{Z \rightarrow f}(0)  + \epsilon \mu_{Z \rightarrow f}(1)) \prod_{j \neq i} \mu_{Y_j \rightarrow f}(0) \nonumber \\
  & & +\; \kappa \mu_{Z \rightarrow f}(1) ((\sum_{\substack{y\\y_i=0}} \prod_{j \neq i} \mu_{Y_j \rightarrow f}(y_j)) - \prod_{j \neq i} \mu_{Y_j \rightarrow f}(0)), \\
  & = & \kappa ((1-\epsilon) \mu_{Z \rightarrow f}(0)  + \epsilon \mu_{Z \rightarrow f}(1)) \prod_{j \neq i} \mu_{Y_j \rightarrow f}(0) \nonumber \\
  & & +\; \kappa \mu_{Z \rightarrow f}(1) (1-\prod_{j \neq i} \mu_{Y_j \rightarrow f}(0)), \\
  & = & \kappa (\mu_{Z \rightarrow f}(1) + (1-\epsilon) (\mu_{Z \rightarrow f}(0)  - \mu_{Z \rightarrow f}(1)) \prod_{j \neq i} \mu_{Y_j \rightarrow f}(0)).
\end{eqnarray}
Now consider the case $\tau = 1$:
\begin{eqnarray}
  \mu_{f \rightarrow Y_i}(1)
  & = & \kappa \sum_{\substack{y\\y_i=1}} \sum_z \Psi^L_\epsilon(y,z) \mu_{Z \rightarrow f}(z) \prod_{j \neq i}\mu_{Y_j \rightarrow f}(y_j), \\
  & = & \kappa \sum_{\substack{y\\y_i=1}} \mu_{Z \rightarrow f}(1) \prod_{j \neq i}\mu_{Y_j \rightarrow f}(y_j), \\
  & = & \kappa \mu_{Z \rightarrow f}(1).
\end{eqnarray}
\end{proof}

\selectiontoy*
\begin{proof}
  By definition,
\begin{eqnarray}
\mu_{f \rightarrow Y}(\tau) & = & \kappa \sum_{z} \Psi^S_\theta(\tau,z) \prod_i \mu_{Z_i \rightarrow f}(z_i).
\end{eqnarray}
Consider the case $\tau = 0$ and note that $\Psi^S_\theta(0,z)=0$ if $z \neq (0,\ldots,0)$.
\begin{eqnarray}
\mu_{f \rightarrow Y}(0) & = & \kappa \prod_i \mu_{Z_i \rightarrow f}(0).
\end{eqnarray}
Now consider the case $\tau = 1$ and note that $\Psi^S_\theta(1,z)=0$ if $c(z) \neq 1$.
\begin{eqnarray}
  \mu_{f \rightarrow Y}(1)
  & = & \kappa \sum_{\substack{z\\c(z)=1}} \Psi^S_\theta(1,z) \prod_i \mu_{Z_i \rightarrow f}(z_i), \\
  & = & \kappa \sum_{i} \theta_i \mu_{Z_i \rightarrow f}(1) \prod_{j \neq i} \mu_{Z_j \rightarrow f}(0), \\
  & = & \kappa (\sum_{i} \theta_i \frac{\mu_{Z_i \rightarrow f}(1)}{\mu_{Z_i \rightarrow f}(0)}) (\prod_{i} \mu_{Z_j \rightarrow f}(0)).
\end{eqnarray}
\end{proof}

\selectiontoz*
\begin{proof}
  By definition,
\begin{eqnarray}
\mu_{f \rightarrow Z_i}(\tau) & = & \kappa \sum_{\substack{z\\z_i=\tau}} \sum_y \Psi^S_\theta(y,z) \mu_{Y \rightarrow f}(y) \prod_{j \neq i} \mu_{Z_j \rightarrow f}(z_j).
\end{eqnarray}
Consider the case $\tau = 0$.
\begin{eqnarray}
  \mu_{f \rightarrow Z_i}(0) & = & \kappa \sum_{\substack{z\\z_i=0\\c(z)=0}} \sum_y \Psi^S_\theta(y,z) \mu_{Y \rightarrow f}(y) \prod_{j \neq i} \mu_{Z_j \rightarrow f}(z_j) \nonumber \\
  & & +\; \kappa \sum_{\substack{z\\z_i=0\\c(z)=1}} \sum_y \Psi^S_\theta(y,z) \mu_{Y \rightarrow f}(y) \prod_{j \neq i} \mu_{Z_j \rightarrow f}(z_j), \\
  & = & \kappa \mu_{Y \rightarrow f}(0) \prod_{j \neq i} \mu_{Z_j \rightarrow f}(0) \nonumber \\
  & & +\; \kappa \sum_{\substack{z\\z_i=0\\c(z)=1}} \Psi^S_\theta(1,z) \mu_{Y \rightarrow f}(1) \prod_{j \neq i} \mu_{Z_j \rightarrow f}(z_j), \\
  & = & \kappa \mu_{Y \rightarrow f}(0) \prod_{j \neq i} \mu_{Z_j \rightarrow f}(0) \nonumber \\
  & & +\; \kappa \mu_{Y \rightarrow f}(1) (\sum_{j \neq i} \theta_j \frac{\mu_{Z_j \rightarrow f}(1)}{\mu_{Z_j \rightarrow f}(0)}) (\prod_{j \neq i} \mu_{Z_j \rightarrow f}(0)).
\end{eqnarray}
Now consider the case $\tau = 1$.  Recall that $\Psi^S_\theta(y,z) = 0$ if $c(z)>1$.
\begin{eqnarray}
  \mu_{f \rightarrow Z_i}(1) & = & \kappa \sum_{\substack{z\\z_i=1\\c(z)=1}} \sum_y \Psi^S_\theta(y,z) \mu_{Y \rightarrow f}(y) \prod_{j \neq i} \mu_{Z_j \rightarrow f}(z_j), \\
  & = & \kappa \theta_i \mu_{Y \rightarrow f}(1) \prod_{j \neq i} \mu_{Z_j \rightarrow f}(0).
\end{eqnarray}
\end{proof}

\end{document}